%% file: arxiv.tex
\documentclass[11pt]{article}

\input{math_commands}

\usepackage[a4paper,top=3cm,bottom=3cm,left=3cm,right=3cm,marginparwidth=1.75cm]{geometry}
\usepackage{amsmath,amssymb,amsthm,amsfonts,bbm,nicefrac,float}
\usepackage{algorithm,algorithmicx,algpseudocode,graphicx}
\usepackage{babel,url,booktabs,microtype,lipsum,natbib,doi,physics,color,xcolor,soul,multirow,caption,rotating,blindtext}
\usepackage{xr,caption,hyperref,enumerate,array}
\usepackage{authblk}
\newtheorem{thm}{Theorem}[section]

\newtheorem{lem}{Lemma}[section]
\newtheorem{cor}{Corollary}[section]
\theoremstyle{definition}

\newtheorem{asm}{Assumption}[section]
\theoremstyle{remark}
 
\numberwithin{equation}{section}

\renewcommand{\hat}{\widehat}
\newcommand{\method}{ACERL }

\title{Contrastive Network Representation Learning}
\makeatletter
\patchcmd{\AB@output}{\par\vspace{1em}}{}{}{}
\renewcommand\AB@affilsepx{\quad}  
\makeatother
\author[1]{Zihan Dong}
\author[2]{Xin Zhou}
\author[1]{Ryumei Nakada}
\author[2]{Lexin Li}
\author[1]{Linjun Zhang}

\affil[1]{Rutgers University}
\affil[2]{University of California at Berkeley}

\begin{document}
\maketitle

\begin{abstract}
Network representation learning seeks to embed networks into a low-dimensional space while preserving the structural and semantic properties, thereby facilitating downstream tasks such as classification, trait prediction, edge identification, and community detection. Motivated by challenges in brain connectivity data analysis that is characterized by subject-specific, high-dimensional, and sparse networks that lack node or edge covariates, we propose a novel contrastive learning-based statistical approach for network edge embedding, which we name as Adaptive Contrastive Edge Representation Learning (ACERL). It builds on two key components: contrastive learning of augmented network pairs, and a data-driven adaptive random masking mechanism. We establish the non-asymptotic error bounds, and show that our method achieves the minimax optimal convergence rate for edge representation learning. We further demonstrate the applicability of the learned representation in multiple downstream tasks, including network classification, important edge detection, and community detection, and establish the corresponding theoretical guarantees. We validate our method through both synthetic data and real brain connectivities studies, and show its competitive performance compared to the baseline method of sparse principal components analysis.
\end{abstract}

\section{Introduction}

Representation learning has emerged as a central problem in statistics and machine learning, aiming to discover meaningful low-dimensional latent features or embeddings that capture the underlying structure of observed data \citep{XuQu2025}. Within this paradigm, network representation learning seeks to embed network data into a low-dimensional space while preserving its  essential key topological and semantic properties. Such learned representations often enable more effective graph-based data analysis, supporting downstream tasks including graph  classification, trait prediction, edge selection, and community detection. See \citet{cai2018survey, zhang2020survey} for systematic reviews on network representation learning. 

Our motivation stems from brain connectivity analysis, a vital field in neuroscience that studies how different brain regions interact and communicate with each other. It provides crucial insights into brain function and structure, as well as the mechanisms underlying numerous neurological and psychiatric disorders. It relies on neuroimaging modalities, such as functional magnetic resonance imaging (fMRI) and diffusion magnetic resonance imaging (dMRI). There has been a rapid advancement in brain connectivity analysis through the use of graph theory tools \citep{wang2021network}. Central to this development is the notion that brain connectivity can be represented as a graph, where nodes correspond to brain regions, and edges capture functional, structural, or directional interactions between regions. It also exhibits some unique features. Each individual subject has a connectivity network derived from a neuroimaging modality, and the subject-to-subject variation is typically large. The underlying network is often sparse, containing only a small subset of nonzero edges, yet high-dimensional, with hundreds of nodes and tens of thousands of edges. Moreover, unlike gene network or social network, brain connectivity network usually contains little node or edge attribute information \citep{fornito2013connectivity}. 

Numerous approaches have been developed for network representation learning. Depending on the target of the embedding, these methods can be broadly categorized into node embedding, which learns a vector representation for each node in the network; edge embedding, which captures relationships between pairs of nodes; and whole-graph or subgraph embedding, which generates representations for an entire graph or its subsets. The last category of methods, e.g., graph2vec \citep{narayanan2017graph2vec}, often build upon the first category, by employing node embedding as the first step. For node embedding, there have been many proposals. Notable examples include deepwalk \citep{perozzi2014deepwalk}, node2vec \citep{grover2016node2vec}, high-order proximity preserved embedding \citep{ou2016hope}, graph convolutional networks \citep{kipf2016semi}, and graphSAGE \citep{hamilton2017inductive}, among many others. In contrast, for edge embedding, there have been much fewer solutions. Notable examples include edge2vec \citep{wang2020edge2vec}, dual hypergraph transformation \citep{jo2021edge}, attrE2vec \citep{bielak2022attrE2vec}, and edge-wise bipartite graph representation learning \citep{wang2024effectiveedgewiserepresentationlearning}. 

There is a line of network representation solutions that employ contrastive learning techniques \citep{you2020graph, zhu2020grace, zhu2021gca, xia2022simgrace}. Contrastive learning is a self-supervised approach that learns meaningful representations by teaching models to distinguish between similar (positive) and dissimilar (negative) pairs of data \citep{chen2020simple}. The core idea is to bring representations of similar inputs closer in the embedding space, while pushing apart dissimilar ones, implicitly regularizing the learned representations \citep{ji2023power}. This technique is particularly useful for network representation learning, as it does not require labeled data, and it encourages robust representations that generalize well on downstream tasks. There has also been some recent work studying statistical properties of contrastive learning  \citep{gui2023unraveling, gui2025multi, wang2023self, wang2023linear, lin2025statistical} 

Despite their success, however, several challenges remain in applying current node or edge embedding methods to brain connectivity analysis. In particular, most existing solutions require node attributes, and are not effective in handling heterogeneous and sparse networks. Moreover, there is very limited literature providing theoretical guarantees when networks are heterogeneous and sparse.

In this article, we propose a new contrastive learning-based statistical model and accompanying methodology, which we name as Adaptive Contrastive Edge Representation Learning (ACERL), for network edge embedding. Our proposal offers several advantages and novel contributions. First, \method constructs network embedding by contrasting positive and negative pairs, and unlike most existing embedding solutions, does \emph{not} require additional node or edge attribute information. It can effectively capture meaningful structural information, and is naturally suited to handle data heterogeneity by learning representation through data-driven augmentation. It further incorporates iterative truncation to induce sparsity in the learned network representation. Second, we introduce an adaptive random masking mechanism, which dynamically adjusts the augmentation process based on the signal-to-noise ratio to enhance feature extraction. This mechanism improves robustness in the presence of heterogeneous noise, and enables effective handling of sparsity by iteratively refining informative features. We note that the principles of this technique extend beyond heterogeneous networks, offering a novel and principled approach to feature extraction from heterogeneous data in general. Third, we establish sharp performance guarantees, and show that \method achieves the minimax optimal rate of convergence, which is significantly faster than most existing contrastive learning methods in the literature. To the best of our knowledge, this is the first minimax optimality result for contrastive network embedding in sparse and heterogeneous settings. Finally, we demonstrate the applicability of the learned representation in multiple downstream tasks, including network classification, important edge detection, and community detection, and derive the corresponding theoretical guarantees. We validate \method  through both synthetic data and real brain connectivity studies, and show its competitive performance compared to the baseline solution.

We adopt the following notation throughout this article. For a set $\gI$, let $\lvert \gI \rvert$ denote its size, and $\gI^c$ its complement. Let $a \vee b = \max\{a, b\}$, and $a \wedge b = \min\{a, b\}$. We use $c$ and $C$ to denote absolute positive constants that may vary from place to place. For two positive sequences $\{a_n\}, \{b_n\}$, write \(a_n \lesssim b_n\) if \(a_n \leq Cb_n\) for all \(n\), \(a_n \gtrsim b_n\) if \(b_n \lesssim a_n\), and \(a_n \asymp b_n\) if \(a_n \lesssim b_n\) and \(b_n \lesssim a_n\). Write $a_n = o(b_n)$ if $\lim_{n} a_n / b_n = 0$, $a_n = O(b_n)$ if $a_n \leq Cb_n$, and $a_n = \Omega(b_n)$ if $a_n \geq Cb_n$ for all $n$ and some positive $C$. For a matrix $A$, let $\|A\|_{\text{sp}}$ and $\|A\|_F$ be its spectral norm and Frobenius norm. Let $\| A \|_{2, 0}$ be the $(2,0)$-matrix-norm defined as the number of non-zero rows of $A$, and $\| A \|_{2, \infty}$ the $(2,\infty)$-matrix-norm defined as the maximum $\ell_2$ norm among all rows of $A$. Let $A_{\gI\gJ}$ be its $|\gI| \times |\gJ|$ submatrix formed by $a_{ij}$ with $(i, j) \in \gI \times \gJ$ for two sets $\gI, \gJ$. For two matrices $A, B$ of the same dimension, let $\langle A, B \rangle = \operatorname{tr}(A^\top B)$ denote the trace inner product, and $\dist(A, B) = \min_{O} \|A O - B\|_F$, where $O$ is an orthogonal rotation matrix.  

The rest of the article is organized as follows. Section \ref{sec:model} introduces the model and the estimation procedure. Section \ref{sec:theory} presents the theoretical analysis, and Section \ref{sec:downstream} studies several downstream tasks. Section \ref{sec:comparePCA} compares to a key alternative method, sparse principal components analysis (sPCA). Section \ref{sec:simulations} presents the simulations, and Section \ref{sec:realdata} illustrates with two real-world brain connectivity studies. The Supplementary Appendix collects all  proofs and additional results.

\section{Model and Estimation}
\label{sec:model}

\subsection{Model Set-up}

Consider a graph $\gG = \gV \times \gE$ with the node set $\gV$ and the edge set $\gE$. Let $v = |\gV|$ denote the total number of nodes, and $d = |\gE|$ the total number of edges. For an undirected acyclic graph, $d = v(v - 1)/2$, and for a directed acyclic graph, $d = v^2$. To simplify the presentation, we focus on the undirected graph in this article, whereas all results can be naturally extended to the directed graph.  

For an individual subject $i \in [n]$, let $x_i = (x_{i1}, \ldots, x_{id})^\top \in \R^d$ denote the vectorized connectivity strength of all edges. Suppose each edge $e \in \gE$ has a corresponding latent edge embedding representation $q_e \in \R^{r}$, and each subject $i$ has a low-dimensional latent subject embedding representation $z_i \in \R^r$, where $r$ is the dimension of the latent feature space. We consider the following generative model,  
\vspace{-0.01in}
\begin{equation} \label{model: conti}
x_i = Q^* z_i + \xi_i, 
\end{equation}
where $\xi_i \in \R^d$ is an independent zero mean sub-Gaussian noise. Here we model the connectivity strength, represented by the $e$th entry $x_{i,e}$ of vector $x_i$, through the inner product $\langle q_e, z_i \rangle$, which, intuitively, measures the alignment between the latent edge embedding $q_e$ and the subject embedding $z_i$. For identifiability, we assume $\Cov(z_i) = I_r$. We also assume $\Cov(\xi_i) = \Sigma^* = \diag(\sigma_1^*, \ldots, \sigma_d^*)$, and throughout this article, we focus on the \emph{heteroscedastic} case, where $\sigma_1^*, \ldots, \sigma_d^*$ can be different. Let $Q^* = (q_1, \ldots, q_d) \in \R^{d \times r}$ denote the \emph{edge embedding matrix} that collects all latent edge representation $q_e$'s, and $Z = (z_1, \dots, z_n) \in \R^{r \times n}$ denote the \emph{subject embedding matrix} that collects all latent subject representation $z_i$'s. Due to rotational invariance, $Q^*$ and $Z$ are only identifiable up to an orthogonal transformation. Moreover, in numerous applications, the graph is inherently sparse. Correspondingly, we introduce a sparsity constraint on $Q^*$, such that $\| Q^* \|_{2, 0} \leq s^*$, where $s^*$ denotes the true sparsity level. 

Following model \eqref{model: conti}, we define $q_{Q^*}(x) = Q^{*\top}x : \mathbb{R}^d \to \mathbb{R}^r$, which is a linear representation function that maps the network data $x_i$ to its latent representation. Following this formulation, learning the edge embedding matrix $Q^*$ is the same as learning the linear representation $q_{Q^*}$ that is parameterized by $Q^*$. Toward that end, we employ contrastive learning, leveraging its capabilities of learning latent representation without label information and handling heterogeneous noise.

Before turning to estimation, we make a few remarks about our model \eqref{model: conti}. First, the edge embedding $q_e$ is crucial for downstream tasks such as edge detection and community detection, whereas the subject embedding $z_i$ is crucial for downstream tasks such as patient classification. For an edge $e$, the connectivity strength $x_{i,e} = \langle q_e, z_i\rangle$ becomes stronger if the subject embedding $z_i$ aligns with the edge embedding $q_e$. Second, in this model, we essentially vectorize all the edges. We note that this is a common practice in edge embedding \citep{wang2024effectiveedgewiserepresentationlearning, jo2021edgerepresentationlearninghypergraphs, goyal2018embedding}. More importantly, there are typically no inherent structural constraints on the adjacency matrices for most networks. As such, representing edges in a vectorized form is mathematically equivalent to using their adjacency matrices. Such formulation is not only simple, but also provides a flexible framework that can be naturally extended to incorporate additional structural constraints if there are any. For instance, for the downstream task of community detection that we study in Section \ref{sec: Community Detection}, the corresponding adjacency matrix satisfies a block-wise structure up to a permutation. In such a case, we extend our model to explicitly reflect the underlying community structure among the nodes, allowing for a more structured and refined analysis. Finally, we remark that, in this article, we primarily focus on linear latent representation learning. Despite its simplicity, it is widely adopted in the literature of both network embedding \citep{roweis2000nonlinear, cui2018survey} and principal components-based analysis \citep{cai2013sparse, gao2023sparsegcathresholdedgradient}, thanks to its interpretability and computational efficiency. We believe it serves as a solid starting point, whereas its theoretical properties already require careful characterization.

\subsection{Construction of augmented views with adaptive masking}

Contrastive learning begins with applying a data augmentation technique, for instance, masking, to generate two correlated views, $h_1(x_i)$ and $h_2(x_i)$, of a given data instance $x_i$, then constructing positive pairs from these views and negative pairs from unrelated instances. However, current contrastive learning methods use a fixed masking strategy, which struggles in the presence of heteroscedastic noise, and can actually introduce significant bias into the learned representations. To overcome this hurdle, we propose an \emph{adaptive} random masking mechanism to construct augmented views of the raw data. Specifically, we utilize two augmentation functions, 
\vspace{-0.01in}
\begin{equation*}
h_1(x_i) = A x_i, \quad\quad h_2(x_i) = (I - A) x_i, 
\end{equation*}
where $A = \diag (a_1, \ldots, a_d) \in \mathbb{R}^{d \times d}$ is the diagonal masking matrix, and the diagonal elements $\{a_e\}_{e=1}^d$ are independent random variables sampled from the following distribution with the masking parameter $p =(p_1, \cdots, p_d)^{\top} \in [0, 1]^d$, 
\begin{align} \label{eq: masking distribution}
a_e = \begin{cases} 
0 & \text{with probability } \frac{1 - p_e}{2}\\
\frac{1}{2} & \text{with probability } p_e \\
1 & \text{with probability } \frac{1 - p_e}{2}
\end{cases}.
\end{align}

Again, the key here is an adaptive adjustment of the masking probability $p_e$. All existing contrastive learning methods set a \emph{fixed} masking probability of $p_e = 0$ for all edges $e \in \gE$. In contrast, we propose to adjust $p_e$ \emph{adaptively} based on the signal-to-noise ratio, which may vary significantly in the heterogeneous noise setting. We describe the detailed adaptive adjustment procedure in Section \ref{sec:estimation}, and provide its rigorous theoretical guarantee in Section \ref{sec:theory}. At the high level, the masking probability $p_e$ directly controls the degree of information shared between the two views. That is, when $a_e = 0$ or $a_e = 1$, the corresponding feature is completely masked in one of the views, and when $a_e = 1/2$, the feature information is shared between the two views. As such, when the estimated signal-to-noise ratio is large, we increase $p_e$, making $a_e$ more likely equal $1/2$, which in turn helps retain more shared information. Conversely, when the estimated signal-to-noise ratio is low, we decrease $p_e$, making $a_e$ more likely equal to 0 or 1, which decreases information sharing between the views. This way, we are able to control the trade-off between complete masking for denoising in highly noisy edges and the preservation of shared feature information across views in less noisy edges. Such an adaptive strategy thus allows our method to effectively handle heterogeneous noise across the network. To the best of our knowledge, our proposal is among the first contrastive learning approach with adaptive masking.

\subsection{Contrastive loss function}

After obtaining the augmented views, we construct the positive pair and the negative pairs for $h_j(x_i)$, $i \in [n], j \in [2]$, as,
\begin{equation*}
\gB_{h_j(x_i)}^{pos} = \big\{ h_w(x_{i}): w \in [2] \text{\textbackslash} \{j\} \big\}, \quad
\gB_{h_j(x_i)}^{neg} = \big\{ h_w(x_{i'}): w \in [2] \text{\textbackslash} \{j\}, i' \in [n] \text{\textbackslash} \{i\} \big\}.
\end{equation*}
That is, we form the positive pair by taking the other view of the same sample $x_i$, because intuitively two augmented views of the same sample contain more similar information. Meanwhile, we form the negative pair by taking the counterpart view of other samples $x_{i'} \in [n] \text{\textbackslash} \{i\}$, because the training samples are usually independent from each other and thus contain different information. 

We then estimate the edge embedding matrix $Q^*$ by minimizing the distance between the positive pairs while maximizing the distance between the negative pairs. That is, we minimize the following loss function with respect to $Q \in \R^{d \times r}$,
\begin{equation}
\label{loss: contrastive loss general form}
\mathcal{L}(Q) =  \frac{1}{n} \sum_{i=1}^{n} \sum_{j=1}^2 \ell\qty(h_j(x_i), \gB_{h_j(x_i)}^{pos}, \gB_{h_j(x_i)}^{neg}; q_Q ) + R(Q),
\end{equation}
where $\ell$ is a contrastive loss function, and $R(Q)$ is a regularization term. For $\ell$, we employ the triplet contrastive loss, which is widely used in contrastive learning \citep[see, e.g.,][]{Hadsell2006, He_2018_CVPR}, and is defined as,
\vspace{-0.01in}
\begin{multline} \label{loss: triplet contrastive loss}
\ell(h_j(x_i), \gB_{h_j(x_i)}^{pos}, \gB_{h_j(x_i)}^{neg}, q_Q) = \\
- \sum_{x \in \gB_{h_j(x_i)}^{pos}} \frac{\langle q_Q(h_j(x_i)), q_Q(x) \rangle}{|\gB_{h_j(x_i)}^{pos}|} + \sum_{x \in \gB_{h_j(x_i)}^{neg}} \frac{\langle q_Q(h_j(x_i)), q_Q(x) \rangle}{|\gB_{h_j(x_i)}^{neg}|}.
\end{multline}

\subsection{Two-level iterative estimation}
\label{sec:estimation}

Next, we develop an estimation procedure to iteratively minimize the contrastive loss while updating the masking parameter, aiming to preserve essential structural information while at the same time to mitigate noise. Our algorithm consists of two levels of iterations: during the \emph{inner} iteration, we estimate the embedding matrix $Q^*$ based on the current masking parameter $p_e$, and during the \emph{outer} iteration, we update $p_e$ based on the current estimate of $Q^*$. 

More specifically, during the inner iteration, given the input network data $\gD = \{ x_i \}_{i=1}^n \subset \R^d$ and the current masking parameter $p$, we employ the contrastive loss function \eqref{loss: contrastive loss general form} to estimate $Q^*$. We set the regularization term $R(Q) = \|Q Q^\top\|_F^2 / 8$ to prevent $Q$ from growing arbitrarily large and to ensure numerical stability. We set the regularization parameter to $1/8$ to simplify the method and theory development, while we allow a more flexible choice of this parameter. We provide more discussion in Appendix \ref{sec:proof of prop distance V and Vtk}. The loss function in \eqref{loss: contrastive loss general form} then becomes, 
\vspace{-0.01in}
\begin{align} \label{loss: cl conti triplet}
\begin{split}
\mL(Q; \mathcal{D}, A) = - \frac{1}{n} \sum_i \langle Q^\top A x_i, & \; Q^\top (I - A) x_i \rangle \; + \\
& \frac{1}{n^2} \sum_{i, j} \langle Q^\top A x_i, Q^\top (I - A) x_j \rangle + \frac{1}{8} \|Q Q^\top\|_F^2.
\end{split}
\end{align}
To minimize \eqref{loss: cl conti triplet}, we employ gradient decent. To further enforce sparsity, we apply the hard thresholding function $HT(Q, s)$ with the working sparsity level $s \in \mathbb{N}$, by keeping the $s$ rows of $Q$ with the largest $\ell_2$ norms and setting all other rows to zeros. We run the inner iterations for $T$ times.

During the outer iteration, we update the masking parameter $p = (p_1, \dots, p_d)^\top$ based on the current estimate of $Q^*$, by setting
\begin{equation} \label{eq: masking p at step k}
p^{(k)}_e = \frac{ \| \hat{q}^{(k-1)}_e\|_2}{\sqrt{\hat\Var(x_e})} \wedge 1,
\end{equation}
where $\hat{q}^{(k-1)}_e$ is the estimate of $q_e$ at the $(k-1)$th iteration, and the sample variance $\hat\Var(x_e) = {\sum_i^n x_{i,e}^2} / {n} - ({\sum _i^n x_{i,e}} / {n})^2$. Since $x_{ie} = q_e^\top z_i + \xi_{ie}$, we note that $\| \hat{q}^{(k-1)}_e\|_2$ can be interpreted as the estimated signal strength of edge $e$ at the $k$th iteration. When the estimated signal is strong, we increase $p_e$ to retain more shared information between the augmented views. When the estimated signal is weak, we decrease $p_e$ to encourage stronger masking augmentation, which in effect facilitates denoising. We run the outer iterations for $K$ times. 

\begin{algorithm}[t!]
\caption{\method.}
\label{alg: edge embeddings conti sparse}
\begin{algorithmic}[1]
\State \textbf{Input:}
Network edge data $\mathcal{D} = \{x_i\}_{i=1}^n \subset \R^d$, initial edge embedding matrix $\hat{Q}^{(0)}$; initial masking parameter $p$, dimension of the representation space $r$, working sparsity level $s$, step size $\eta > 0$, number of inner iterations $T$, number of outer iterations $K$.
\For{$k \in [K]$}
\State Set $\hat{Q}_{(0)}^{(k)} = \hat{Q}^{(k-1)}$.
\For{$t \in [T]$}
\State Generate $A_{(t)}^{(k)} = \diag\left( a_{(t),1}^{(k)}, \dots, a_{(t),d}^{(k)} \right)$ following \eqref{eq: masking distribution}.
\State Update the edge embedding matrix estimate as
\begin{eqnarray*} 
\hat{Q}_{(t)}^{(k)} & \leftarrow & \widetilde{Q}_{(t-1)}^{(k)} - \eta \partial_Q \mL\left( \widetilde{Q}_{(t-1)}^{(k)}; \mathcal{D}, A_{(t)}^{(k)} \right),\\
\widetilde{Q}_{(t)}^{(k)} & \leftarrow & HT\left( \hat{Q}_{(t)}^{(k)}, s \right).  
\end{eqnarray*}
\EndFor
\State Set $\hat{Q}^{(k)} = \widetilde{Q}_{(T)}^{(k)}$.
\State Update the masking parameter $p^{(k+1)}_e = \frac{ \| \hat{q}^{(k)}_e\|_2}{\sqrt{\hat\Var(x_e)}} \wedge 1$ following \eqref{eq: masking p at step k}.
\EndFor
\State \textbf{Output:} Edge embedding matrix estimator $\hat{Q} = \hat Q^{(K)}$.
\end{algorithmic}
\end{algorithm}

We also briefly discuss the initial value and tuning parameters. For the initial edge embedding estimator $\hat{Q}^{(0)}$, let $M = n^{-1} X X^\top - n^{-2} X 1_n 1_n^\top X^\top \in \R^{d \times d}$, $X = (x_1, \ldots, x_n) \in \R^{d \times n}$, $1_n = (1, \ldots, 1)^\top \in \R^{n}$, $D(M)$ be the diagonal matrix with the same diagonal elements of $M$, and $\Delta(M) = M - D(M)$. We then apply the Fantope projection method \citep{vu2013fantope} to $\Delta(M)$ and apply PCA to obtain an initial estimator for $\hat{Q}^{(0)}$. For the latent dimension $r$ and the sparsity level $s$, we recommend some graph-based empirical approach, and give more details in Section \ref{sec:simulations}. For the number of inner and outer iterations $T$ and $K$, we suggest setting them to be of the order of $\log n$. We set the step size $\eta$ between 0.1 and 1 , and our numerical experiments show that the results are not overly sensitive within this range. 

We summarize the above estimation procedure in Algorithm \ref{alg: edge embeddings conti sparse}.

\section{Theoretical Guarantees}
\label{sec:theory}

We establish the theoretical guarantee for our proposed method, in particular, the minimax optimal convergence rate of the estimator from Algorithm \ref{alg: edge embeddings conti sparse}. We first obtain the upper bound for the estimator at the $k$th outer iteration, then the upper bound of the estimated edge embedding matrix $Q^*$. 

We note that, since our goal is to learn $Q^*$ up to an orthogonal transformation, we present our results in the form of $Q^* Q^{*^\top}$. Consider the spectral decomposition 
\begin{equation*}
Q^* Q^{*^\top} = U^* \Lambda^* U^{*^\top}, \quad \text{where} \;\;  \Lambda^* = \diag(\lambda_1^{*2}, \dots, \lambda_r^{*2}) \in \R^{r \times r}, 
\end{equation*}
with the diagonal elements in a descending order, and $U^* \in \R^{d \times r}$ is orthonormal. Moreover, since the random masking is generated at each iteration of the gradient descent, generating multiple masking matrices $A$'s can be viewed as averaging over different samples. Consequently, the gradient updates can be interpreted as stochastic gradient descent on the following expected loss \eqref{loss: cl conti triplet transpose expectation} over the randomness of $A$. \begin{equation} \label{loss: cl conti triplet transpose expectation}
\gL(Q) = \mathbb{E}_A[\mathcal{L}(Q; \mathcal{D}, A)].
\end{equation}

We begin with a set of regularity conditions. 

\begin{asm}\label{asm: Sigma}
Suppose $\sigma_{(1)}^{*2} \asymp 1$, where $\sigma_{(j)}^{*2}$ denotes the $j$th largest of $\sigma_1^{*2}, \ldots, \sigma_d^{*2}$. 
\end{asm}

\begin{asm}\label{asm: Lambda}
Suppose $\lambda_1^{2*} \asymp 1$, and $\kappa_{\Lambda^*} = {\lambda_1^{*2}} / {\lambda_r^{*2}} < c_1$ for some constant $c_1$.
\end{asm}

\begin{asm}\label{asm: Q_sp incoherent}
Suppose the incoherent constant of $U^*$, i.e., $I(U^*) = \|U^*\|_{2,\infty} ^2$, satisfies that $I(U^*) \leq {c_2} / {r}$ for some sufficiently small constant $c_2 > 0$.
\end{asm}

\begin{asm}\label{asm: initializer}
Suppose the initial edge embedding matrix estimator $\hat Q^{(0)}$ satisfies that $\big\| \hat{Q}^{(0)}\hat{Q}^{(0)^\top} - Q^*Q^{*^\top} \big\|_F \leq {\lambda_r^{*3}}/{(20 \lambda_1^*)}$.
\end{asm}

We make some remarks on these conditions. Assumption~\ref{asm: Sigma} allows heterogeneous noise levels. Assumption~\ref{asm: Lambda} ensures that the signal strengths across coordinates are of the same order. Additionally, they assume that the signals and noise variances are of a constant order. These conditions are commonly imposed in the literature \citep[see, e.g.,][]{zhang2022heteroskedastic, gao2023sparsegcathresholdedgradient, johnstone2009consistency, vu2013minimax}. Assumption~\ref{asm: Q_sp incoherent} intuitively quantifies the extent to which the signals are evenly distributed across coordinates. It implies that the signal coordinates are correlated, and they can be distinguished from random noise. Similar conditions have also been considered in the literature of low-rank matrix analysis \citep[see, e.g.,][]{candes2008exactmatrixcompletionconvex, zhang2022heteroskedastic}. Assumption~\ref{asm: initializer} is about the initial estimator $\hat{Q}^{(0)}$, which is obtained by applying the Fantope projection method in our case. By \citet[][Theorem 3.3]{vu2013fantope} and Assumption \ref{asm: Q_sp incoherent}, we have that $\|\hat{Q}^{(0)}\hat{Q}^{(0)\top} - Q^*Q^{*\top}\|_F \lesssim s^*/\sqrt{r}$. Consequently, our initial estimator $\hat{Q}^{(0)}$ satisfies Assumption \ref{asm: initializer}.

We now present our main theoretical results in two parts, first for the estimator at a fixed $k$th outer iteration, then for the estimator when $k$ increases to $K$. 

\begin{thm} \label{prop: distance V and Vtk}
For a given $k = 1, \ldots, K$, define the masking-adjusted covariance matrix estimation error as,
\begin{equation} \label{eq: W^{(k)}-1 defination}
W^{(k-1)} = \sup_{\gI \subset [d] \text{ with } |\gI| = 2s + s^*}\norm{\qty(\Delta (M) + P^{(k)^2} D(M) - Q^* Q^{*^\top})_{\gI\gI}}_{\text{sp}}
\end{equation}
where $P^{(k)} = \diag\left( p^{(k)}_1, \ldots, p^{(k)}_d \right)$, and $M, \Delta(M)$ and $D(M)$ are as defined in Section \ref{sec:estimation}. Suppose $W^{(k-1)} \leq {c_3\lambda_r^{*3}}/{(\lambda^*_1\sqrt{r})}$ for some sufficiently small positive constant $c_3$, and $\big\| \widetilde{Q}^{(k)}_{(0)}\widetilde{Q}^{(k)^\top}_{(0)} - Q^*Q^{*^\top} \big\|_F \leq {\lambda_r^{*3}}/{(20 \lambda_1^*)}$. Set the step size $\eta \leq {6}/{(43\lambda_1^{*2})}$, and the sparsity level $s \geq {36 s^*}/{(\lambda_r^{*4}  \eta^2)}$. Then, for all \(t \geq 1\), we have
\begin{equation} \label{eq: distance of the output of the inner iteration}
\dist\qty( \widetilde{Q}^{(k)}_{(t+1)}, Q^* ) \leq \xi^t \dist\qty( \widetilde{Q}^{(k)}_{(0)}, Q^* ) + \frac{3}{\lambda_r^*}\sqrt{\frac{sr}{s^*}} W^{(k-1)}. 
\end{equation}
where $\xi = 1 - {\lambda_r^{*2} \eta}/{2}$.
\end{thm}

We first briefly remark that, as $\dist\qty( \widetilde{Q}^{(k)}_{(t+1)}, Q^* )$ gradually decays, the conditions for $W^{(k-1)}$ and $\widetilde{Q}^{(k)}_{(0)}$ in this theorem hold for all $k$ once they hold for $k = 0$. Further details can be found in the proof.
Theorem \ref{prop: distance V and Vtk} provides an upper bound of the distance between $Q^*$ and $\tilde{Q}^{(k)}_t$ for a fixed outer iteration $k$ while the number of inner iterations $t$ increases. This bound consists of two terms, the first is an optimization error term that decays geometrically as $t$ increases, and the second is a statistical error term that accounts for the irreducible error due to finite samples and does not diminish as $t$ increases. As such, when $t$ increases, the first term is to be dominated by the second term, which relies on $W^{(k-1)}$. Note that $W^{(k-1)}$ depends only on the masking parameter $p$, which varies with the outer iterations, but not the inner iterations. In the next theorem, we establish the estimation error of our edge embedding estimator by characterizing the evolution of $W^{(k-1)}$ when the outer iteration $k$ increases to $K$. 

\begin{thm} \label{thm: edge embeddings sparse}
Suppose Assumptions \ref{asm: Sigma} to \ref{asm: initializer} hold, and $s^* \log d \ll n$. Set the step size $\eta \leq {6}/{(43\lambda_1^{*2})}$, and the sparsity level $s \geq {36 s^*}/{(\lambda_r^{*4}  \eta^2)}$. For inner iterations, we run $T^{(k)} = \Omega\qty(\log n \wedge \log \qty(2^k r))$ times, and for outer iterations, we run $K = \Omega(\log n)$ times. Then, with probability at least $1 - ce^{-c s \log d}$, we have
\vspace{-0.01in}
\begin{equation} \label{eq: edge embedding error bound sparse}
\norm{\hat{Q}^{(K)} \hat{Q}^{(K)^\top} - Q^* Q^{*^\top}}_F \lesssim \frac{s}{s^*}\sqrt{\frac{s^* r \log d}{n}}.
\end{equation}
\end{thm}

Theorem \ref{thm: edge embeddings sparse} shows a controlled deviation of our final estimated edge embedding matrix $\tilde{Q}^{(K)}$ after $K$ outer iterations from the ground truth  $Q^*$. In particular, the factor ${s}/{s^*}$ accounts for the selection of the sparsity level, and since $s$ can be chosen as the same order of $s^*$, the error remains $\sqrt{{s^* r \log d} / {n}}$ up to a constant.

We also remark that Theorem \ref{thm: edge embeddings sparse} essentially shows \method  achieves the minimax optimal rate of convergence. This is because, when the noise is homogeneous, i.e., when $\sigma_1^* = \ldots = \sigma_d^*$ in model \eqref{model: conti}, our problem reduces to sPCA. Since the homogeneous noise setting is a special case of the heterogeneous one, the results established for sPCA under the homogeneous setting \citep{cai2013sparse, vu2013minimax} provide a direct minimax lower bound for \method, which is $\sqrt{{s^*(r+\log d)}/{n}}$. This implies that, when $r$ is finite, \method  achieves the minimax optimal rate. We further study and compare with sPCA under the heterogeneous setting in Section \ref{sec:comparePCA}. 

Finally, we remark on the advantage of adaptive masking. As shown in Theorem \ref{thm: edge embeddings sparse}, \method removes the bias term, an issue that is often further amplified in a sparse setting. \cite{ji2023power} studied standard contrastive learning in the classical non-sparse setting with fixed masking, and established the estimation error rate of $\sqrt{{rd}/{n}} + (r^{3/2} \log d) / d$. We can easily modify our proposed method to incorporate the non-sparse setting. That is, we still consider model \eqref{model: conti}, but impose no sparsity constraint. Accordingly, for estimation, we modify Algorithm \ref{alg: edge embeddings conti sparse} slightly by dropping the hard thresholding step, $\widetilde{Q}_{(t)}^{(k)} \leftarrow HT\left( \hat{Q}_{(t)}^{(k)}, s \right)$. Then, we can establish the corresponding estimation error rate of the non-sparse estimator. 
 
\begin{asm}\label{asm: Q incoherent}
Suppose the incoherent constant of $U^*$ satisfies that $I(U^*) \leq c_4$ for some sufficiently small constant $c_4 > 0$.
\end{asm}

\begin{cor}
\label{cor: edge embeddings}
Suppose Assumptions \ref{asm: Sigma}, \ref{asm: Lambda} and \ref{asm: Q incoherent} hold, and that $r < d \ll n$. We run the outer iterations $K = \Omega\qty(\log n)$ times. Then, with probability at least $1 - ce^{-c\sqrt{n/d} \wedge d}$, we have
\vspace{-0.01in}
\begin{align} \label{eq: edge embedding error bound}
\norm{\hat Q^{(K)} \hat Q^{(K)^\top}  - Q^* Q^{*^\top}}_{\text{sp}} \lesssim \sqrt{\frac{d}{n}}.
\end{align}
\end{cor}

Compared to \citet{ji2023power}, our method offers two main advantages. First, our condition is milder. That is, our Assumption \ref{asm: Q incoherent}, which is the counterpart of Assumption \ref{asm: Q_sp incoherent} in the non-sparse setting, is weaker than the condition in \citet{ji2023power}, which required the incoherence constant to satisfy $I(U^*) = O((r \log d)/d)$. Second, our algorithm achieves a sharper error rate. That is, Corollary \ref{cor: edge embeddings} establishes a Frobenius norm error of $\sqrt{rd/n}$, which improves upon the rate $\sqrt{rd/n} + (r^{3/2}\log d)/d$ obtained in \citet{ji2023power}. The improvement comes from masking, where fixed masking introduces a bias term, and our adaptive masking successfully removes it.

\section{Downstream Tasks}
\label{sec:downstream}

We next investigate how network edge embedding can facilitate downstream tasks, in particular, subject classification, edge selection, and node community detection. For each task, we discuss the underlying model, our solution based on edge embedding, the evaluation criterion, and the theoretical quantification of task performance.

\subsection{Subject classification task}
\label{sec: Classification}

For the task of subject classification, we consider a binary classification model setting, where for the subject $i \in [n]$,
\vspace{-0.1in}
\begin{equation} \label{eqn:binary-model}
y_i \mid z_i \sim \text{Bernoulli}(F(\langle z_i, w^* \rangle)),
\vspace{-0.05in}
\end{equation}
in which $y_i$ is the binary label, $z_i$ is the latent subject embedding that connects with the observed feature vector $x_i$ through model \eqref{model: conti}, $w^* \in \mathbb{R}^r$ is a unit-norm coefficient vector, and $F : \mathbb{R} \to [0, 1]$ is a monotonically increasing function satisfying $1 - F(u) = F(-u)$ for all $u \in \mathbb{R}$. This formulation covers many standard classification models, including the logistic model and the probit model. Our goal is to classify a new subject, based on the feature information $x_0$,  into one of the two classes. 

Given the training data $\{x_i, y_i\}_{i=1}^{n}$, we first obtain the estimated edge embedding matrix $\hat{Q}$ following Algorithm \ref{alg: edge embeddings conti sparse}. Note that this estimation step only uses $\{x_i\}_{i=1}^{n}$, and thus is unsupervised. We then estimate the corresponding patient embedding as,  
\begin{equation*}
\hat{z}_i = (\hat{Q}^\top \hat{Q} )^{-1} \hat{Q}^\top x_i, \;\; i \in [n].
\end{equation*}
Given $\{\hat{z}_i, y_i\}_{i=1}^{n}$, we fit a classification model and obtain the estimated coefficient $\hat{w}$. Finally, for a new testing data point $x_0$, we obtain the classifier, 
\begin{equation*}
\delta_{\hat{Q}, \hat{w}}(x_0) = \mathbbm{1}\left\{F\left( \hat{w}^\top \hat z_0 \right) \geq \frac{1}{2}\right\} = \mathbbm{1}\left\{F\left( \hat{w}^\top (\hat{Q}^\top \hat{Q} )^{-1} \hat{Q}^\top x_0 \right) \geq \frac{1}{2}\right\},
\end{equation*}
where $\mathbbm{1}(\cdot)$ is the indicator function. 

We evaluate the classifier using the usual 0-1 loss, $\mathcal{L}_c(\delta) = \mathbbm{1}\left\{y_0 \neq \delta(x_0)\right\}$, for a classifier $\delta$. We then establish the theoretical bound for the excess risk of classification, 
\begin{equation*}
\inf_{w \in \mathbb{R}^r} \mathbb{E}_0\qty[\mathcal{L}_c\qty(\delta_{\hat{Q}, w})],
\end{equation*}
where $\mathbb{E}_0[\cdot]$ is the expectation with respect to $(y_0, x_0, z_0)$. Since our goal is to evaluate the performance of classification task based on the learned edge embedding $\hat{Q}$, rather than the specific procedure to train the classifier, we take the infimum over $w \in \R^r$ to select the best classifier within the family $\left\{ \delta_{\hat{Q}, w} \mid w \in \R^r \right\}$ in our evaluation.

\begin{thm} \label{thm: classification risk}
Suppose the eigenvalues of the noise covariance matrix $\Sigma^*$ are all of a constant order. Suppose $z_0$ and $\xi_0$ are  Gaussian random variables. In addition, for the edge embedding estimation error, $E(\hat Q) = \| \hat{Q} \hat{Q}^\top - Q^* Q^{*^\top} \|_F$, suppose 
\begin{equation} \label{eq: assumption bound for R}
E(\hat Q) \leq {\sigma_{(d)}^2} \wedge \frac{1}{2},
\end{equation}
where $\sigma_{(d)}$ is the smallest eigenvalue of $\Sigma$. Then, the excess risk for classification is bounded as:
\begin{equation*}
\inf_{w \in \mathbb{R}^r} \mathbb{E}_0\qty[\mathcal{L}_c\qty(\delta_{\hat{Q}, w})] - \inf_{w \in \mathbb{R}^r} \mathbb{E}_0\qty[\mathcal{L}_c\qty(\delta_{Q^*, w})] \lesssim E(\hat Q).
\end{equation*}
\end{thm}

We first note that the condition on $E(\hat Q)$ in \eqref{eq: assumption bound for R} is satisfied by Theorem \ref{thm: edge embeddings sparse}. Meanwhile, the Gaussian variable assumption is made for simplicity and is commonly used when evaluating the excess risk \citep{ji2023power, xu2019numbervariablesuseprincipal}. Besides, this assumption can be relaxed. Then Theorem \ref{thm: classification risk} shows that the excess classification error is controlled by the estimation error of the learned edge embedding. For the downstream subject classification task, the classification error when using our edge embedding estimator would not be worse than the oracle case when using the true edge embedding by more than $(s / s^*) \sqrt{(s^* r \log d)/n}$.

\subsection{Edge selection task}
\label{sec: detect important edges}

For the task of edge selection, we adopt a simple yet effective strategy; i.e., we deem the edges with larger $\ell_2$-norms of their embeddings more important. Correspondingly, after we obtain the estimated edge embedding matrix $\hat{Q}$ following Algorithm \ref{alg: edge embeddings conti sparse}, we compute the norm $\|\hat q_e\|_2$ for each edge $e$, and select the top $s$ edges with the largest norm values. 

We evaluate the selection performance based on the estimated support, $\hat{\mathcal{C}} = \{ e : \|\hat{q}_e\|_2 \text{ ranks among the } s^* \text{ largest values in } \{\|\hat{q}_j\|_2\}_{j=1}^{d} \}$, and compare it to the true support, ${\mathcal{C}} = \{ e : \|q_e\|_2 \text{ ranks among the } s^* \text{ largest values in } \{\|{q}_j\|_2\}_{j=1}^{d} \}$. We next establish the theoretical guarantee of the support recovery. 
\begin{thm} \label{thm: detect important edges continuous}
Suppose the assumptions of Theorem~\ref{thm: edge embeddings sparse} hold. Suppose  
\begin{equation*}
\|q_e\|_2 \gtrsim \frac{s}{s^*} \sqrt{\frac{s^* r \log d}{n}}, \;\; \text{if} \;\; \|q_e\|_2 \neq 0.
\end{equation*}  
Then, with probability at least $1 - ce^{-c s \log d}$, we have $\hat{\mathcal{C}} = \mathcal{C}$. 
\end{thm}

We first note that the signal gap condition is quite weak. When the first $r$ eigenvalues of $\Lambda^*$ are at a constant level, we have $\|q_e\| \approx r/d$, which is much larger than $(s/s^*) \sqrt{(s^* r \log d)/n}$. Then Theorem \ref{thm: detect important edges continuous} shows that, by leveraging the estimation error bound in Theorem~\ref{thm: edge embeddings sparse}, all important edges can be reliably identified by ranking the norms of the estimated edge embeddings.

\subsection{Node community detection task}
\label{sec: Community Detection}

For the task of community detection, we consider a stochastic block model setting \citep{holland1983stochastic, Lei_2015}. Recall that, under our model \eqref{model: conti}, the norm of the edge embedding represents the connectivity strength between two nodes. For community detection, we introduce an additional structure to \eqref{model: conti}, by assuming that, if two edges that connect the node pairs from the same pair of communities, then the norms of their embeddings would be the same. This in turn induces a stochastic block model in the embedding space, and enables community detection based on the edge embedding matrix $\hat{Q}$. More specifically, suppose the node set $\mathcal{V}$ with $v = |\mathcal{V}|$ nodes is partitioned into $G$ disjoint communities $\mathcal{C}_1, \ldots, \mathcal{C}_G$, where each node belongs to exactly one community. We consider the following model,
\begin{equation}
\label{model: community}
x_i = Q^* z_i + \xi_i, \quad \|q_e\|^2 = S_{u_e u_e'}, \quad S = \Theta B \Theta^\top \in \R^{v \times v},
\end{equation}
where $e$ denotes an edge connecting the nodes $u_e$ and $u_e'$, $\Theta \in \R^{v \times G}$ is the membership matrix that encodes the community to which the each node belongs, with each row containing exactly one element of one and all remaining elements zero, $B \in \R^{G \times G}$ is the inter-community connection strength matrix, with its $(g,g')$th entry $b_{gg'}$ representing the connection strength between community $\mathcal{C}_g$ and $\mathcal{C}_{g'}$, and $\text{rank}(B) = G$, and $S = \Theta B \Theta^\top \in \R^{v \times v}$ is the corresponding similarity matrix, which captures the overall connection strengths between all pairs of nodes. Our goal is to recover the community membership matrix $\Theta$.

Given the observed network data $\{x_i\}_{i=1}^{n}$, we first obtain the estimated edge embedding matrix $\hat{Q}$ following Algorithm \ref{alg: edge embeddings conti sparse}. We then estimate the similarity matrix $\hat{S}$, by setting $\hat{S}_{u_e u_e'} = \hat{S}_{u_e' u_e}= \| \hat{q}_e \|_2$, and setting the diagonal entries to zero. We then conduct spectral clustering for community detection, by applying the $(1 + \epsilon)$-approximate $k$-means clustering \citep{Kumar2004, Lei_2015} to the normalized and Laplacian transformed similarity matrix $\hat{S}$. We denote the final estimated community membership matrix as $\hat{\Theta}$. We give more details of this community detection procedure based on edge embedding in Appendix \ref{sec:appendix-community}.

We evaluate the community detection accuracy using two criteria. The first is the overall relative error that measures the overall percentage of mis-clustered nodes, 
\vspace{-0.01in}
\begin{equation*}
\mathcal{L}(\hat{\Theta}, \Theta) = v^{-1} \min_{J} \| \hat{\Theta} J - \Theta \|_0, 
\end{equation*}
over all $G \times G$ permutation matrices $J$. The second is the worst-case community-wise misclassification error, 
\begin{equation*}
\widetilde{\mathcal{L}}(\hat{\Theta}, \Theta) = \min_{J} \max_{g} v_g^{-1} \| (\hat{\Theta} J)_{G_{g}} - \Theta_{G_{g}} \|_0,
\end{equation*}
over all $G \times G$ permutation matrices $J$ and over $g \in [G]$, $G_g = G_g(\Theta) = \{1 \leq i \leq v : \Theta_{ig} = 1\}$, $v_g = |G_g| = \sum_{i=1}^v \Theta_{ig}$ is the size of community $g$, $g\in[G]$. Note that  $0 \leq \mathcal{L}(\hat{\Theta}, \Theta) \leq \widetilde{\mathcal{L}}(\hat{\Theta}, \Theta) \leq 2$. As such, $\widetilde{\mathcal{L}}$ is a stronger criterion than $\mathcal{L}$, in the sense that $\widetilde{\mathcal{L}}$ requires the estimator to perform well among all communities, while $\mathcal{L}$ allows some small communities to have large relative errors. We next establish the theoretical bound for these two criteria. 

\begin{thm} \label{thm: community detection error}
Let $\rho_G = \max_{g \in [G]} B_{gg}$, $\tau_v = \min_{j \in [v]} D_{jj} / v$,where $D$ is the diagonal degree matrix with $D_{vv} = \sum_{v'} S_{v,v'}$, and $\lambda_G$ be the smallest absolute nonzero eigenvalue of the normalized Laplacian matrix $L = D^{-\frac{1}{2}} S D^{-\frac{1}{2}}$. Suppose $128 (2 + \epsilon) \rho_G^2 G^2 / (\tau_v^2 v^2 \lambda_G^2)$ $\leq 1$. In addition, for the edge embedding estimation error, suppose 
\begin{equation} \label{eq: edge embedding condition for cluster}
E(\hat Q) = o\qty(\tau_v \sqrt{v} \wedge \rho_G),
\end{equation}
Then, the community detection errors are bounded as:
\begin{align*}
\widetilde{\mathcal{L}}(\hat{\Theta}, \Theta) \leq 128 (2 + \epsilon) \frac{\rho_G^2 G^2}{\tau_v^2 v^2 \lambda_G^2}, \quad\quad
\mathcal{L}(\hat{\Theta}, \Theta)  \leq 128 (2 + \epsilon) \frac{\rho_G^2 G^2 v_{\max}'}{\tau_v^2 v^3 \lambda_G^2},
\end{align*}
where $v_{\max}'$ is the second largest community size.
\end{thm}

We first note that the condition on $E(\hat Q)$ in \eqref{eq: edge embedding condition for cluster} is satisfied by Theorem \ref{thm: edge embeddings sparse}. Moreover, when the community sizes are balanced, i.e., $v_{\max} / v_{\min} = O(1)$, where $v_{\min} = \min_{1 \leq g \leq G} v_g$ and $v_{\max} = \max_{1 \leq g \leq G} v_g$, and $\tau_v = \Omega(\log v / v)$, we have that $128 (2 + \epsilon) \rho_G^2 G^2 / (\tau_v^2 v^2 \lambda_G^2) \leq O(G^2 / \log^2 v)$. Consequently, when the number of communities satisfies that $G = o(\log v)$, we have that the condition $128 (2 + \epsilon) \rho_G^2 G^2 / (\tau_v^2 v^2 \lambda_G^2) \leq 1$ holds. Then Theorem \ref{thm: community detection error} shows the community detection error remains small, as long as the edge embedding error is small.

\section{Comparison with Sparse PCA}
\label{sec:comparePCA}

Our proposed method is closely related to another popular unsupervised dimension reduction method, i.e., sparse principal components analysis. There have been several methods studying the theoretical properties of sPCA in detail \citep{cai2013sparse, vu2013minimax, vu2013fantope, gao2023sparsegcathresholdedgradient}. However, they all focus on the homogeneous noise setting. In contrast, \method  considers the heterogeneous noise setting. In this section, we study sPCA under heterogeneous noise, by establishing the lower bound for its embedding estimation as well as downstream classification, and comparing with \method. To the best of our knowledge, this is the first work to theoretically study sparse PCA under heterogeneous noise. 

We begin with a brief review of sPCA, which seeks the latent embedding through the top $r$ eigenvectors $U_x \in \R^{d \times r}$ of the covariance matrix $\Sigma_x = \E[X X^\top] \in \R^{d \times d}$, where $X = (x_1, \ldots, x_n) \in \R^{d \times n}$, while imposing additional sparsity that $\|U_x\|_{2, 0} \leq s^*$. Actually, sPCA is closely related to our latent model \eqref{model: conti}, under which $\Sigma_x = Q^* Q^{*^\top} + \Sigma^* = U^*\Lambda^* U^{*\top} + \Sigma^*$, $U^* \in \R^{d \times r}, \Lambda^* \in \R^{r \times r}$. When the noise is homogeneous, i.e., $\sigma_1^* = \ldots = \sigma_d^* = \sigma^*$,  $\Sigma_x = U^* (\Lambda^* + \sigma^* I_d) U^{*\top}$. Therefore, sPCA recover the target $U^*$ through the top eigenvectors $\hat{U}_x$ of the sample covariance matrix $\hat{\Sigma}_x$. 

Next, we study the estimation performance of sPCA under the heterogeneous noise setting. We establish the lower bound for the sPCA estimator, and demonstrate that heteroskedastic noise leads to a fundamental deterioration in its statistical performance. We first note that the existing sPCA methods \citep{cai2013sparse, vu2013minimax, vu2013fantope, gao2023sparsegcathresholdedgradient} all satisfy the following assumption. 

\begin{asm} \label{asm: bound for spca}
Suppose $\norm{\hat{U}_{x} \hat{U}_{x}^\top - U_x U_x^\top}_F \leq c_5$, with probability at least $1 - c e^{-c s \log d}$, for some constants $c_5 > 0$. 
\end{asm}

\begin{thm} \label{thm: lower bound for sPCA}
Suppose Assumptions \ref{asm: Q_sp incoherent} and \ref{asm: bound for spca} hold. Then, 
\begin{equation*}
\sup_{(Q^*, \Sigma^*) \in \gM_s} \P \qty(\norm{\hat{U}_{x}\hat{U}_{x}^\top - U^*U^{*^\top}}_F \gtrsim \sqrt{r}) \geq 1- c e^{-c s \log d},  
\end{equation*}
over the parameter space $\gM_s$ of  $(Q^*, \Sigma^*)$, such that $\| Q^* \|_{2,0} \leq s^*$ and $Q^*$ satisfies Assumptions \ref{asm: Lambda}, \ref{asm: Q_sp incoherent}, and $\Sigma^*$ satisfies Assumption \ref{asm: Sigma}.
\end{thm}

Theorem \ref{thm: lower bound for sPCA} shows that, under heterogeneous noise, the worst-case scenario of sPCA incurs an estimation error of at least $O(\sqrt{r})$. In contrast, as shown in Theorem~\ref{thm: edge embeddings sparse}, our estimator achieves an error rate of $(s/s^*) \sqrt{(s^* r \log d)/n}$ that is much smaller. Moreover, under heterogeneous  noise, the top $r$ eigenvalues of the sample covariance matrix are distorted due to the heterogeneous structure of the noise. As a result, sPCA is likely to fail in recovering the true signal subspace. 

Next, we investigate the impact of sPCA on the downstream task of subject classification. Given the sPCA estimator $\hat{U}_{x}$ and a new data $x_0$, its low-dimensional embedding is computed as $\hat{z}_0 = \hat{\Lambda}_r^{-\frac{1}{2}} \hat{U}_x^\top x_0$, where $\hat{\Lambda}_r = \diag(\hat\sigma_1, \ldots, \hat\sigma_r)$.  Following Section \ref{sec: Classification}, the corresponding classifier is $\delta'_{\hat{U}_x, \hat{w}}(x_0) = \mathbbm{1}\left\{F(\hat{w}^\top \hat{\Lambda}_r^{-\frac{1}{2}} \hat{U}_x^\top x_0) \geq 1/2 \right\}$. Similar to Theorem~\ref{thm: classification risk}, we take the infimum over $w \in \R^r$ to select the best classifier within the family $\{\delta_{\hat{Q}, w} \mid w \in \R^r\}$ in our evaluation. 

\begin{thm}
\label{thm: classification risk lower bound}
Suppose $\sigma_{(1)}^{*2} \asymp 1$ and $\kappa_\Sigma = {\sigma_{(1)}^{*2}} / {\sigma_{(d)}^{*2}} \asymp 1$. Suppose $z_0$ and $\xi_0$ are Gaussian random variables. In addition, suppose 
\vspace{-0.01in}
\begin{equation}
\label{eq: RU assumption bound}
E_{U^*}^2(\hat{U}_x) \geq r - \frac{\sigma_{(1)}^2}{\kappa_\Sigma \qty(1 + \sigma_{(1)}^2)} + \frac{c_6}{\kappa_\Sigma},
\end{equation}
for some constant $c_6> 0$, where $E_{U^*}^2(\hat{U}_x) = \big\| \hat{U}_x \hat{U}_x^\top - U^* U^{*^\top} \big\|_F$. Then, the excess risk for classification is bounded as
\begin{equation*}
\inf_{w \in \mathbb{R}^r} \mathbb{E}_0\qty[\mathcal{L}_c\qty(\delta'_{\hat{U}_x, w})] - \inf_{w \in \mathbb{R}^r} \mathbb{E}_0\qty[\mathcal{L}_c\qty(\delta_{Q^*, w})] \geq c_P,
\end{equation*}
where $c_P$ is a positive constant that depends only on $c_6$, $\sigma_{(1)}^{*2}$, and $\kappa_\Sigma$.
\end{thm}

We first note that the condition in on $E_{U^*}^2(\hat{U}_x)$ in \eqref{eq: RU assumption bound} is satisfied by applying Theorem~\ref{thm: lower bound for sPCA}. Then this theorem shows that the classification based on sPCA suffers a poor performance due to heterogeneous errors. In contrast, as shown in Theorem~\ref{thm: classification risk}, our method leads to a much better classification performance, demonstrating its advantage in the presence of heterogeneous errors.

\section{Simulation Studies}
\label{sec:simulations}

We conduct intensive simulations to investigate the empirical performance of our \method method through the three downstream tasks. We also compare \method with sPCA, following  the implementation of \citet{seghouane2019sparse}. We exclude graph neural networks (GNNs) from comparison, as our preliminary study indicates that GNNs perform poorly when node feature information is unavailable. 

We simulate the model following \eqref{model: conti}, 
\vspace{-0.1in}
\begin{equation*}
x_{i,e} = 1.25 c_e q_e^\top z_i   + \sigma_\xi \xi_{i,e}, 
\vspace{-0.05in}
\end{equation*}
where for every $i \in [n]$ and $e \in [d]$, we draw the entry of $z_i$ and $q_e$ independently from a standard normal distribution, and the noise $\xi_{i,e}$ independently from a normal distribution with mean zero and standard deviation $e/d$. We set the number of nodes $v = \{45, 55, 64\}$, such that the number of edges $d = \{990, 1485, 2016\}$ and the number of edges are close to $\{1000, 1500, 2000\}$, respectively. We set the dimension of the latent space $r =\{10, 20\}$. We use the binary vector $c = (c_1, \ldots, c_d)$ to represent important edges, set the true sparsity level $s^* = 50$, and randomly set $s^*$ entries of $c$ as 1 and the rest as 0. We use the constant $\sigma_\xi$ to control the magnitude of the noise level, and set $\sigma_\xi = \{0, 2, 4, 6\}$ to represent no, small, medium and large noise, respectively. In addition, for the subject classification task, we assign each subject is to a binary class based on whether the first entry or the second entry in the subject embedding is larger. We set the sample size $n = \{250, 500, 750\}$, and use the first 60\% data for training and the remaining 40\% for testing. We train \method  or sPCA based on the training data, and report the performance of each task based on the testing data. We replicate all experiments 50 times. For the subject classification task, we employ a support vector machine classifier. For the edge detection task, we choose a working sparsity level $s = 150$. For the community detection task, we consider a smaller non-sparse network, by setting the number of nodes $v = \{15, 18, 21\}$, such that the number of edges $d = \{105, 153, 210\}$, and setting all edges important. We generate $x_{i,e} = 5 \sqrt{c_{u_e} c_{u_e'}} f(u_e, u_e') q_e^\top z_i   + \sigma_\xi \xi_{i,e}$, where $c_{u_e}, c_{u_e'}$ are uniformly independently drawn from $\text{Uniform}(0.1,1.1)$,  and $f(c_{u_e},c_{u_e'})=10^{-|c_{u_e}-c_{u_e'}|}$. 

Table \ref{tab:sim_12} reports the classification accuracy for the subjection classification task. Table \ref{tab:sim_11} reports the selection accuracy for the edge selection task, and Table \ref{tab:sim_21} reports the rand index of node clustering for the community detection task. In all these tables, we see that the performance improves with a larger sample size, a smaller number of edges, or a smaller noise level, which agrees with our theory. Furthermore, \method achieves a better performance than sPCA across all tasks and settings.

\begin{table}[t!]
	\centering
	\caption{Subject classification task: average classification accuracy (in percentage) and standard error (in parenthesis), across different noise levels.}
	\label{tab:sim_12}
	\resizebox{\textwidth}{!}{
		\begin{tabular}{c|@{\hspace{0.3em}}c|@{\hspace{0.3em}}c|@{\hspace{0.3em}}c@{\hspace{0.3em}}|c@{\hspace{0.3em}}c@{\hspace{0.3em}}c@{\hspace{0.3em}}c@{\hspace{0.3em}}|c@{\hspace{0.3em}}c@{\hspace{0.3em}}c@{\hspace{0.3em}}c}
			\hline
			& & &  & \multicolumn{4}{c|}{$r=10$} & \multicolumn{4}{c}{$r=20$} \\ \hline
			$n$ & $v$ & $d$ & $\sigma_\xi$ & 0 &  2 &  4 &  6 &  0 &  2 &  4 &  6 \\
			\hline
			\multirow{6}{*}{$250$} & \multirow{2}{*}{$45$} & \multirow{2}{*}{$990$} & \method & 100.0(0.0) & 100.0(0.0) & 98.8(1.4) & 78.1(5.6)
			& 100.0(0.0) & 100.0(0.0) & 100.0(0.3) & 90.9(3.6)\\
			&&& sPCA & 94.8(2.5) & 92.2(3.5) & 82.2(5.1) & 72.0(5.5)
			& 93.7(3.0) & 87.2(3.9) & 78.4(5.8) & 68.5(6.6)\\
			\cline{2-12}
			& \multirow{2}{*}{$45$} & \multirow{2}{*}{$1485$} & \method & 100.0(0.0) & 100.0(0.0) & 97.9(2.0) & 72.8(6.7)
			& 100.0(0.0) & 100.0(0.0) & 99.9(0.4) & 90.2(3.0)\\
			&&& sPCA & 96.0(2.2) & 91.3(3.2) & 78.7(5.7) & 67.9(5.4)
			& 92.8(2.8) & 87.2(3.6) & 74.8(5.8) & 67.7(5.2)\\
			\cline{2-12}
			& \multirow{2}{*}{$45$} & \multirow{2}{*}{$2016$} & \method & 100.0(0.0) & 100.0(0.0) & 98.1(1.4) & 67.4(7.3)
			& 100.0(0.0) & 100.0(0.0) & 99.8(0.6) & 87.9(4.1)\\
			&&& sPCA & 96.1(1.8) & 90.5(3.1) & 77.3(5.5) & 66.1(7.0)
			& 92.5(2.9) & 87.0(4.5) & 73.7(5.3) & 66.1(5.8)\\
			\hline
			\multirow{6}{*}{$500$} & \multirow{2}{*}{$45$} & \multirow{2}{*}{$990$} & \method & 100.0(0.0) & 100.0(0.0) & 100.0(0.0) & 93.6(3.0)
			& 100.0(0.0) & 100.0(0.0) & 100.0(0.0) & 96.9(2.2)\\
			&&& sPCA & 97.3(1.4) & 93.6(1.9) & 86.2(3.4) & 75.0(4.8)
			& 96.3(1.5) & 91.2(2.4) & 81.1(3.7) & 72.1(4.6)\\
			\cline{2-12}
			& \multirow{2}{*}{$45$} & \multirow{2}{*}{$1485$} & \method & 100.0(0.0) & 100.0(0.0) & 99.9(0.4) & 90.9(3.6)
			& 100.0(0.0) & 100.0(0.0) & 100.0(0.0) & 97.8(1.5)\\
			&&& sPCA & 97.3(1.2) & 93.5(1.8) & 84.0(4.0) & 71.3(5.2)
			& 96.4(1.8) & 90.6(3.1) & 79.1(4.1) & 70.9(5.1)\\
			\cline{2-12}
			& \multirow{2}{*}{$45$} & \multirow{2}{*}{$2016$} & \method & 100.0(0.0) & 100.0(0.0) & 99.8(0.8) & 89.6(3.9)
			& 100.0(0.0) & 100.0(0.0) & 100.0(0.0) & 97.4(2.1)\\
			&&& sPCA & 97.2(1.4) & 93.3(2.3) & 83.2(3.7) & 70.4(5.4)
			& 96.1(1.5) & 90.1(2.4) & 80.0(4.6) & 70.6(4.8)\\
			\hline
			\multirow{6}{*}{$750$} & \multirow{2}{*}{$45$} & \multirow{2}{*}{$990$} & \method & 100.0(0.0) & 100.0(0.0) & 100.0(0.0) & 96.6(2.1)
			& 100.0(0.0) & 100.0(0.0) & 100.0(0.0) & 99.3(0.9)\\
			&&& sPCA & 97.8(1.0) & 94.7(1.5) & 86.5(3.4) & 77.3(4.8)
			& 97.2(1.0) & 92.4(2.1) & 83.0(4.2) & 73.7(3.8)\\
			\cline{2-12}
			& \multirow{2}{*}{$45$} & \multirow{2}{*}{$1485$} & \method & 100.0(0.0) & 100.0(0.0) & 100.0(0.0) & 96.5(2.0)
			& 100.0(0.0) & 100.0(0.0) & 100.0(0.0) & 99.5(1.0)\\
			&&& sPCA & 98.0(0.9) & 93.8(1.7) & 85.2(3.9) & 75.4(4.8)
			& 96.8(1.2) & 91.8(1.9) & 81.6(4.9) & 72.8(4.5)\\
			\cline{2-12}
			& \multirow{2}{*}{$45$} & \multirow{2}{*}{$2016$} & \method & 100.0(0.0) & 100.0(0.0) & 100.0(0.0) & 94.9(2.7)
			& 100.0(0.0) & 100.0(0.0) & 100.0(0.0) & 99.2(1.3)\\
			&&& sPCA & 97.6(0.9) & 93.8(1.5) & 84.5(3.8) & 72.1(5.3)
			& 96.9(0.9) & 91.4(1.9) & 80.5(4.1) & 73.0(5.0)\\
			\hline
	\end{tabular}}
\end{table}

\begin{table}[t!]
	\centering
	\caption{Edge selection task: the average selection accuracy (in percentage) and the standard error (in parenthesis) of important edges under the working sparsity level $s=150$ when the true sparsity level $s^* = 50$.}
	\label{tab:sim_11}
	\resizebox{\textwidth}{!}{
		\begin{tabular}{c|@{\hspace{0.3em}}c|@{\hspace{0.3em}}c|@{\hspace{0.3em}}c@{\hspace{0.3em}}|c@{\hspace{0.3em}}c@{\hspace{0.3em}}c@{\hspace{0.3em}}c@{\hspace{0.3em}}|c@{\hspace{0.3em}}c@{\hspace{0.3em}}c@{\hspace{0.3em}}c}
			\hline
			& & &  & \multicolumn{4}{c|}{$r=10$} & \multicolumn{4}{c}{$r=20$} \\ \hline
			$n$ & $v$ & $d$ & $\sigma_\xi$ & 0 &  2 &  4 &  6 &  0 &  2 &  4 &  6 \\
			\hline
			\multirow{6}{*}{$250$} & \multirow{2}{*}{$45$} & \multirow{2}{*}{$990$} & \method & 100.0(0.0) & 100.0(0.0) & 98.8(1.4) & 78.1(5.6)
			& 100.0(0.0) & 100.0(0.0) & 100.0(0.3) & 90.9(3.6)\\
			&&& sPCA & 100.0(0.0) & 100.0(0.0) & 98.0(2.0) & 69.2(6.5)
			& 100.0(0.0) & 100.0(0.0) & 99.8(0.6) & 85.4(4.8)\\
			\cline{2-12}
			& \multirow{2}{*}{$45$} & \multirow{2}{*}{$1485$} & \method & 100.0(0.0) & 100.0(0.0) & 97.9(2.0) & 72.8(6.7)
			& 100.0(0.0) & 100.0(0.0) & 99.9(0.4) & 90.2(3.0)\\
			&&& sPCA & 100.0(0.0) & 100.0(0.0) & 96.8(2.8) & 64.6(6.3)
			& 100.0(0.0) & 100.0(0.0) & 99.8(0.6) & 84.6(4.3)\\
			\cline{2-12}
			& \multirow{2}{*}{$45$} & \multirow{2}{*}{$2016$} & \method & 100.0(0.0) & 100.0(0.0) & 98.1(1.4) & 67.4(7.3)
			& 100.0(0.0) & 100.0(0.0) & 99.8(0.6) & 87.9(4.1)\\
			&&& sPCA & 100.0(0.0) & 100.0(0.0) & 96.5(2.2) & 59.0(7.3)
			& 100.0(0.0) & 100.0(0.0) & 99.5(0.9) & 81.8(4.5)\\
			\hline
			\multirow{6}{*}{$500$} & \multirow{2}{*}{$45$} & \multirow{2}{*}{$990$} & \method & 100.0(0.0) & 100.0(0.0) & 100.0(0.0) & 93.6(3.0)
			& 100.0(0.0) & 100.0(0.0) & 100.0(0.0) & 96.9(2.2)\\
			&&& sPCA & 100.0(0.0) & 100.0(0.0) & 99.7(0.7) & 87.1(5.2)
			& 100.0(0.0) & 100.0(0.0) & 100.0(0.0) & 94.4(2.9)\\
			\cline{2-12}
			& \multirow{2}{*}{$45$} & \multirow{2}{*}{$1485$} & \method & 100.0(0.0) & 100.0(0.0) & 99.9(0.4) & 90.9(3.6)
			& 100.0(0.0) & 100.0(0.0) & 100.0(0.0) & 97.8(1.5)\\
			&&& sPCA & 100.0(0.0) & 100.0(0.0) & 99.7(0.7) & 83.5(4.9)
			& 100.0(0.0) & 100.0(0.0) & 100.0(0.0) & 94.6(2.3)\\
			\cline{2-12}
			& \multirow{2}{*}{$45$} & \multirow{2}{*}{$2016$} & \method & 100.0(0.0) & 100.0(0.0) & 99.8(0.8) & 89.6(3.9)
			& 100.0(0.0) & 100.0(0.0) & 100.0(0.0) & 97.4(2.1)\\
			&&& sPCA & 100.0(0.0) & 100.0(0.0) & 99.5(1.0) & 81.0(5.9)
			& 100.0(0.0) & 100.0(0.0) & 100.0(0.0) & 94.0(2.9)\\
			\hline
			\multirow{6}{*}{$750$} & \multirow{2}{*}{$45$} & \multirow{2}{*}{$990$} & \method & 100.0(0.0) & 100.0(0.0) & 100.0(0.0) & 96.6(2.1)
			& 100.0(0.0) & 100.0(0.0) & 100.0(0.0) & 99.3(0.9)\\
			&&& sPCA & 100.0(0.0) & 100.0(0.0) & 100.0(0.3) & 92.4(3.7)
			& 100.0(0.0) & 100.0(0.0) & 100.0(0.0) & 97.3(1.8)\\
			\cline{2-12}
			& \multirow{2}{*}{$45$} & \multirow{2}{*}{$1485$} & \method & 100.0(0.0) & 100.0(0.0) & 100.0(0.0) & 96.5(2.0)
			& 100.0(0.0) & 100.0(0.0) & 100.0(0.0) & 99.5(1.0)\\
			&&& sPCA & 100.0(0.0) & 100.0(0.0) & 100.0(0.3) & 91.6(3.2)
			& 100.0(0.0) & 100.0(0.0) & 100.0(0.0) & 97.5(2.4)\\
			\cline{2-12}
			& \multirow{2}{*}{$45$} & \multirow{2}{*}{$2016$} & \method & 100.0(0.0) & 100.0(0.0) & 100.0(0.0) & 94.9(2.7)
			& 100.0(0.0) & 100.0(0.0) & 100.0(0.0) & 99.2(1.3)\\
			&&& sPCA & 100.0(0.0) & 100.0(0.0) & 99.9(0.4) & 89.4(4.2)
			& 100.0(0.0) & 100.0(0.0) & 100.0(0.0) & 97.4(1.8)\\
			\hline
	\end{tabular}}
\end{table}

\begin{table}[t!]
\centering
\caption{Community detection task: the average rand index of node clustering (in percentage) and the standard error (in parenthesis). }
\label{tab:sim_21}
\resizebox{\textwidth}{!}{
\begin{tabular}{c|@{\hspace{0.3em}}c|@{\hspace{0.3em}}c|@{\hspace{0.3em}}c@{\hspace{0.3em}}|c@{\hspace{0.3em}}c@{\hspace{0.3em}}c@{\hspace{0.3em}}c@{\hspace{0.3em}}|c@{\hspace{0.3em}}c@{\hspace{0.3em}}c@{\hspace{0.3em}}c}
\hline
& & &  & \multicolumn{4}{c|}{$r=10$} & \multicolumn{4}{c}{$r=20$} \\ \hline
$n$ & $v$ & $d$ & $\sigma_\xi$ & 0 &  2 &  4 &  6 &  0 &  2 &  4 &  6 \\
\hline
\multirow{6}{*}{250} & \multirow{2}{*}{15}& \multirow{2}{*}{105}
& \method & 82.7(14) & 82.3(14) & 64.2(22) & 57.1(8.5) & 82.3(14) & 76.0(14) & 45.7(8.1) & 52.0(10) \\
& && sPCA    & 55.4(11) & 46.1(9.0)  & 40.4(10) & 41.9(5.1) & 44.8(4.7)  & 42.1(9.7)  & 40.4(6.3) & 41.3(8.6) \\
\cline{2-12}

 & \multirow{2}{*}{18}& \multirow{2}{*}{153}
& \method & 88.4(14) & 82.6(14) & 83.1(14) & 91.4(14) & 82.4(14) & 85.0(19) & 73.5(16) & 51.4(11) \\
& && sPCA    & 50.1(7.4)  & 45.8(9.6)  & 39.9(7.8)  & 44.2(12) & 42.4(6.7)  & 34.0(0.0)  & 46.0(10) & 40.0(9.5) \\
\cline{2-12}

 & \multirow{2}{*}{21}& \multirow{2}{*}{210}
& \method & 88.4(9.8) & 96.9(6.3) & 93.0(11) & 88.8(14) & 88.7(23) & 91.1(18) & 71.7(20) & 52.4(12) \\
& && sPCA    & 45.0(6.7) & 42.1(10) & 44.6(12) & 42.3(8.8)  & 51.4(6.8)  & 39.2(6.8)  & 37.4(7.5)  & 38.4(7.2) \\
\hline
\multirow{6}{*}{500} & \multirow{2}{*}{15}& \multirow{2}{*}{105}
& \method & 82.3(14) & 88.2(14) & 63.6(8.4) & 64.0(13) & 82.3(14) & 77.1(12) & 62.5(18) & 60.6(8.4) \\
& && sPCA    & 54.3(8.1)  & 47.0(5.5)  & 44.6(10) & 46.5(7.3)  & 49.0(4.4)  & 48.2(7.4)  & 51.2(1.1)  & 56.0(3.5) \\
\cline{2-12}

 & \multirow{2}{*}{18}& \multirow{2}{*}{153}
& \method & 94.6(11) & 94.2(12) & 88.9(14) & 92.5(15) & 90.6(12) & 88.4(14) & 70.2(18) & 63.5(21) \\
& && sPCA    & 46.9(11) & 53.6(1.4)  & 44.2(8.4)  & 49.5(7.9)  & 46.8(8.3)  & 43.9(8.1)  & 52.3(9.7)  & 50.2(4.0) \\
\cline{2-12}

 & \multirow{2}{*}{21}& \multirow{2}{*}{210}
& \method & 89.5(13) & 100(0.0) & 76.7(22) & 81.8(15) & 100(0.0) & 96.9(6.3) & 88.6(14) & 71.3(20) \\
& && sPCA    & 45.1(6.6)  & 47.4(9.8)  & 48.2(8.2)  & 46.5(12) & 54.3(1.7)  & 56.2(3.7)  & 47.1(9.4)  & 43.1(8.3) \\
\hline
\multirow{6}{*}{750} & \multirow{2}{*}{15}& \multirow{2}{*}{105}
& \method & 82.7(14) & 78.5(19) & 62.9(20) & 63.4(13) & 83.0(14) & 70.9(0.8) & 79.8(17) & 66.1(16) \\
& && sPCA    & 51.0(1.7)  & 51.2(1.3)  & 45.9(11) & 46.9(12) & 49.0(4.4)  & 55.4(2.6)  & 53.3(3.5)  & 55.0(14) \\
\cline{2-12}

 & \multirow{2}{*}{18}& \multirow{2}{*}{153}
& \method & 88.8(14) & 82.7(14) & 82.9(14) & 87.3(13) & 89.5(13) & 88.8(14) & 68.4(15) & 79.7(19) \\
& && sPCA    & 51.8(7.2)  & 47.6(6.8)  & 56.2(4.0)  & 52.5(2.3)  & 51.5(3.0)  & 56.7(6.6)  & 42.6(8.7)  & 51.6(2.9) \\
\cline{2-12}

 & \multirow{2}{*}{21}& \multirow{2}{*}{210}
& \method & 84.1(13) & 100(0.0) & 90.7(11) & 89.5(13) & 94.4(11) & 95.3(9.3) & 89.1(13) & 73.7(14) \\
& && sPCA    & 48.4(7.5)  & 51.5(4.3)  & 50.7(6.5)  & 46.0(10) & 54.3(1.7)  & 48.5(6.6)  & 43.3(10) & 42.5(9.0) \\
\hline
\end{tabular}}
\end{table}

\begin{figure}[t!]
\centering
\begin{tabular}{cc}
\includegraphics[width=0.4\linewidth,height=3.8cm]{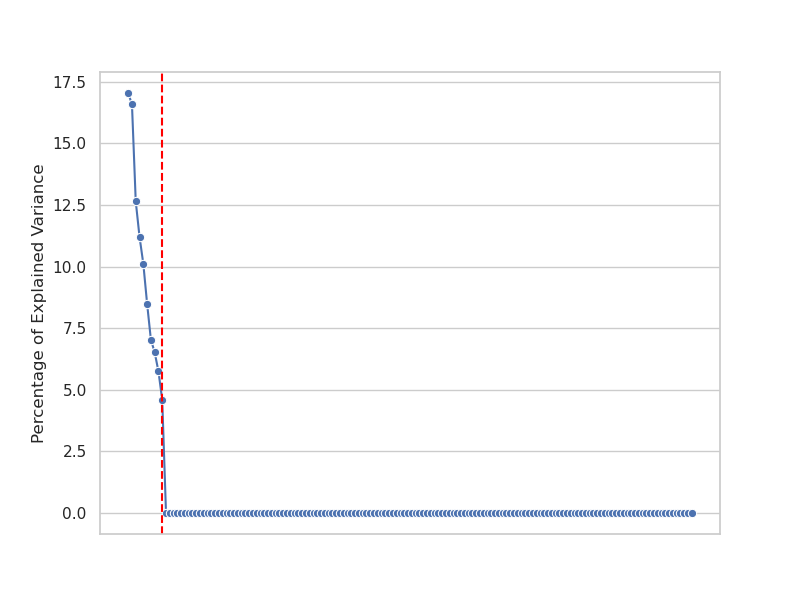} &
\includegraphics[width=0.4\linewidth,height=3.8cm]{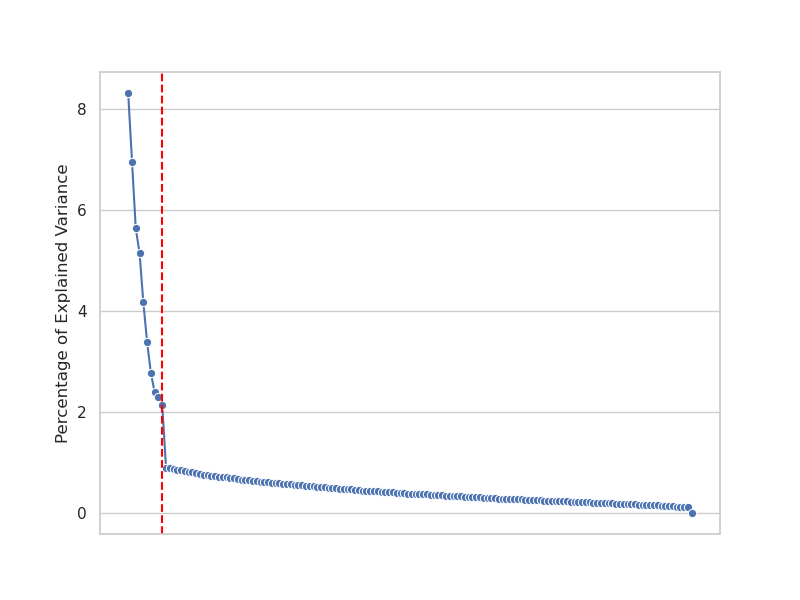} \\
$\sigma_\xi = 0$ & $\sigma_\xi = 2$ \\
\includegraphics[width=0.4\linewidth,height=3.8cm]{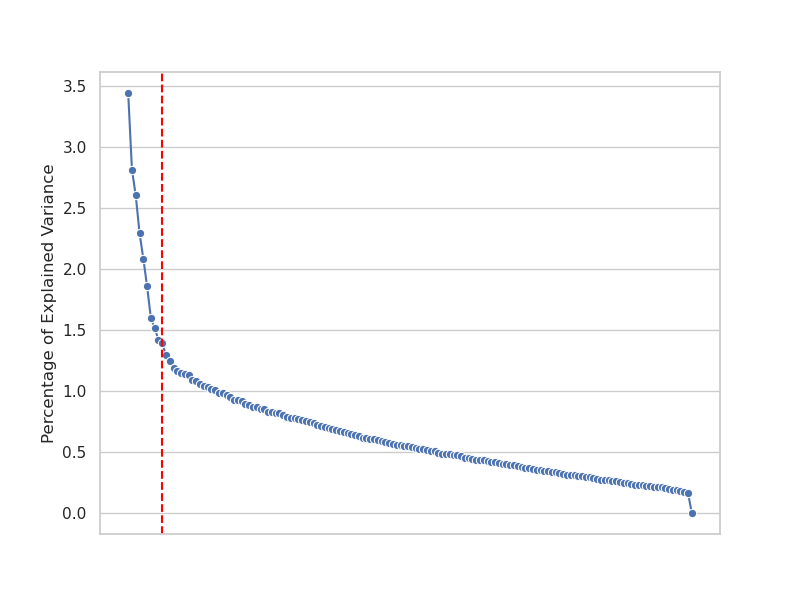} &
\includegraphics[width=0.4\linewidth,height=3.8cm]{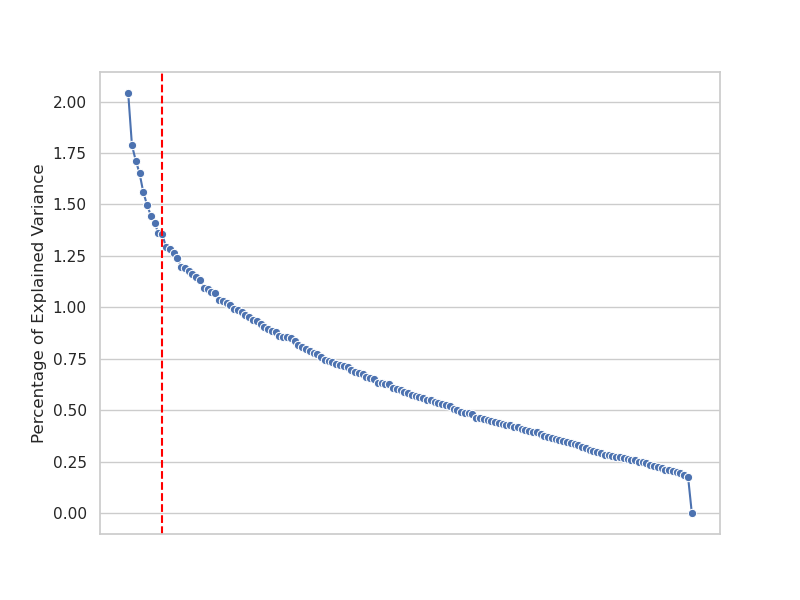} \\
$\sigma_\xi = 4$ & $\sigma_\xi = 6$ \\
\end{tabular}
\caption{Percentage of explained variance for each component or $n=250$, $d=990$, and $r=10$ under different noise levels. The red line corresponds to the component at the true value of $r = 10$.}
\label{fig:exp_var_PCA}
\end{figure}

\begin{figure}[t!]
\centering
\begin{tabular}{cc}
\includegraphics[width=0.4\linewidth,height=3.8cm]{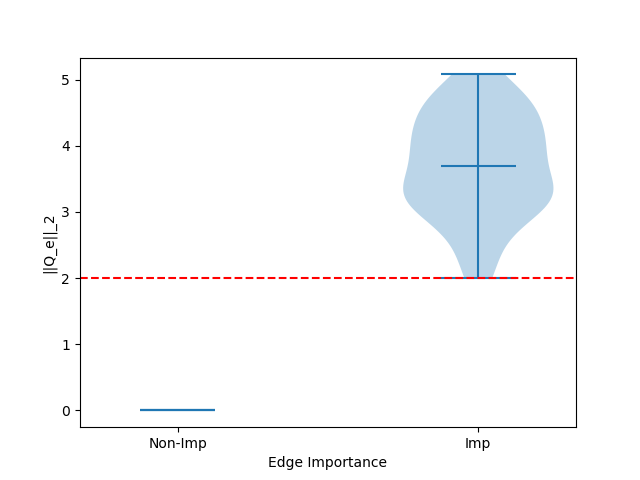} & 
\includegraphics[width=0.4\linewidth,height=3.8cm]{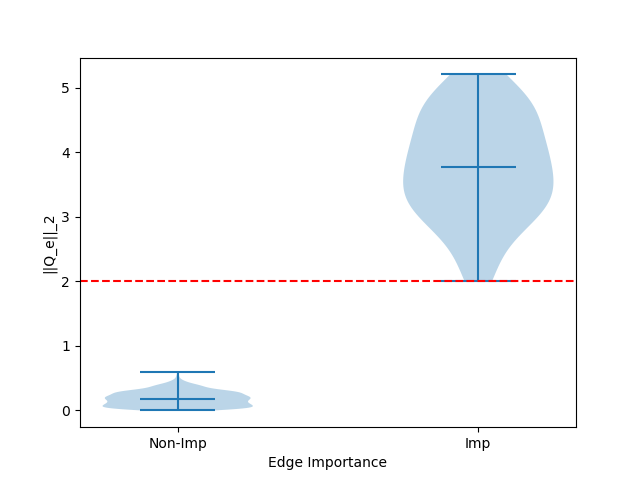} \\
$\sigma_\xi = 0$ & $\sigma_\xi = 2$ \\
\includegraphics[width=0.4\linewidth,height=3.8cm]{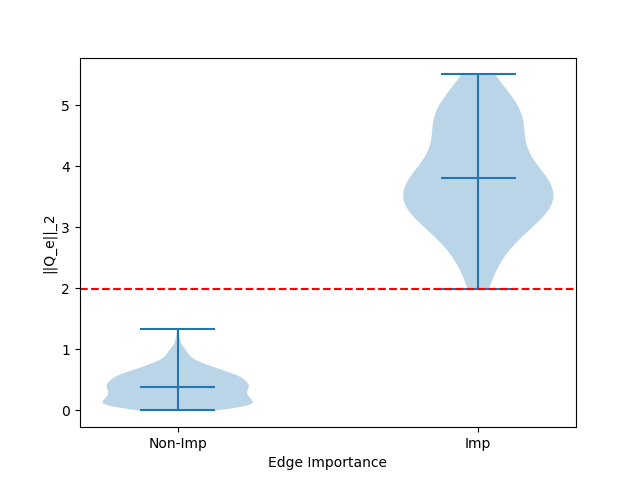} &
\includegraphics[width=0.4\linewidth,height=3.8cm]{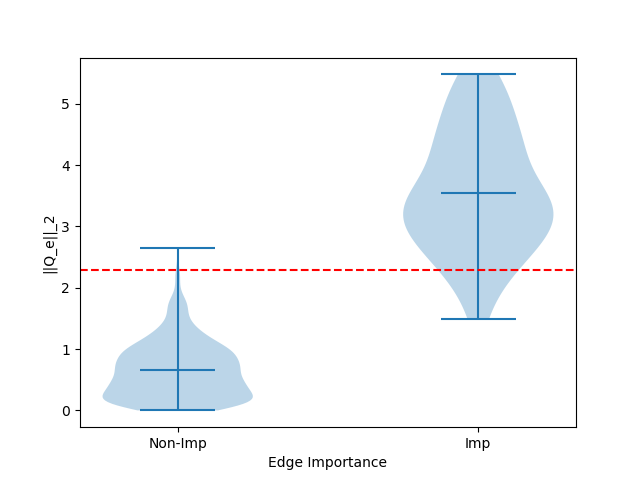} \\
$\sigma_\xi = 4$ & $\sigma_\xi = 6$ \\
\end{tabular}
\caption{Violin plot for the $\ell_2$ norms of the edge embeddings for \method. The red line corresponds to the true sparsity level.}
\label{fig:Id_ImpEdge}
\end{figure}

In the simulations so far, we fix the latent space dimension $r$ at the true values, and fix the sparsity level $s$ at a predefined value. Next, we discuss how to choose $r$ and $s$ given the data. For $r$, we recommend to adopt a similar approach as PCA and use the explained variance percentage. Figure \ref{fig:exp_var_PCA} shows the percentage of explained variance when  $n=250$, $d=990$, and $r=10$ under different noise levels, with the red line corresponding to the component at the true value of $r=10$. We observe a relatively clear gap around the true value of $r$. For $s$, we recommend to examine the empirical distribution of $\|\hat{Q}_e\|$ to help determine the threshold value. Figure \ref{fig:Id_ImpEdge} shows the violin plot when $n=500$, $d=990$, and $r=10$ under different noise levels, with the horizontal line corresponding to the true sparsity level.

\section{Brain Connectivity Analysis}
\label{sec:realdata}

Brain connectivity analysis examines the structural and functional relationships between different regions of the brain, providing insight into how neural circuits are organized and interact. It plays a crucial role in neuroscience by helping researchers understand the mechanisms underlying normal brain development, cognitive functions, and aging. Moreover, alterations in connectivity patterns are often associated with neurological and psychiatric disorders such as Alzheimer's disease, autism, and schizophrenia. Connectivity analysis models the brain as a network, where regions are represented as nodes and their interactions as edges, and network representation learning helps to uncover hidden brain patterns, identify potential biomarkers, and predict disease states \citep{fornito2013connectivity}. Next, we illustrate our \method method with two brain connectivity analysis examples.

\subsection{Autism Brain Imaging Data Exchange}

The first example is the Autism Brain Imaging Data Exchange (ABIDE), which studies autism spectrum disorder, a prevalent neurodevelopmental disorder with symptoms including social difficulties, communication deficits, stereotyped behaviors and cognitive delays \citep{DiMartino2014}. The dataset, after preprocessing, consists of 786 subjects, among whom 430 are diagnosed with autism, and 356 are healthy controls. Each subject received a resting-state functional magnetic resonance imaging (fMRI) scan, which is summarized as a $116 \times 116$ functional connectivity network, with nodes corresponding to 116 brain regions-of-interest (ROIs) based on the standard Automated Anatomical Labelling brain atlas, and edges measuring the mutual information of the fMRI signals between all pairs of ROIs. 

We applied \method  and sPCA to this data, and carried out two downstream tasks: classifying the subjects into the two classes, and identifying potentially important brain regions. We transformed the mutual information measure by its square. We randomly split the data, using 60\% for training and the remaining 40\% for testing. We repeated the split 50 times, and report the results based on 50 replications. We employed a support vector machine classifier. We chose $r=10$ as the dimension of the latent space, by the percentage of explained variance. 

\begin{figure}[t!]
\centering
\begin{tabular}{cc}
\includegraphics[width=0.5\linewidth,height=5.5cm]{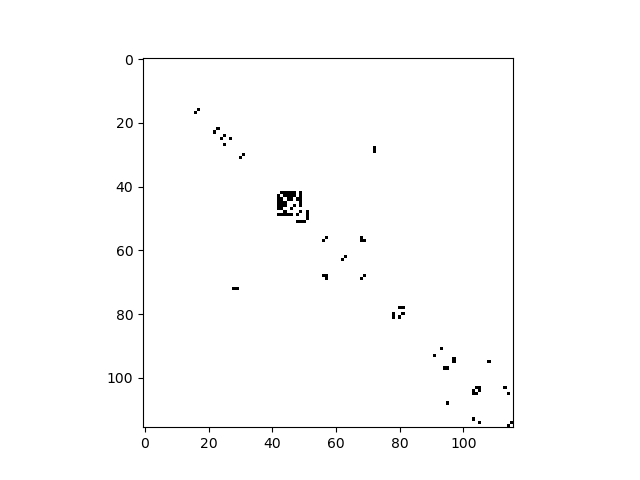} & 
\includegraphics[width=0.5\linewidth,height=5.5cm]{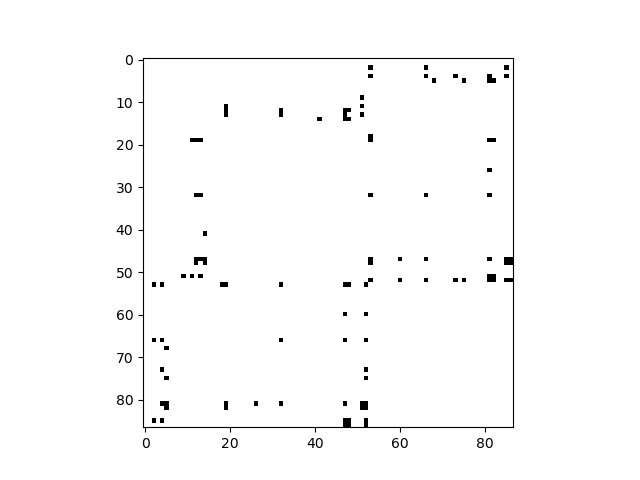} \\
ABIDE & HCP \\
\end{tabular}
\caption{Sparse graph for ABIDE and HCP data at the sparsity level $s=100$.}
\label{fig:real-data}
\end{figure}

Table \ref{tab:real-data}(a) reports the misclassification error rate on the testing data, where we see that \method  outperforms sPCA. Figure \ref{fig:real-data}(a) reports the sparse edge adjacency matrix when setting the sparsity level at $s=100$. There are four nodes with noticeably larger degrees, corresponding to four brain regions: right superior occipital gyrus, left calcarine sulcus, right calcarine sulcus, and left cuneus. These regions align well with the ASD literature. Notably, they are all part of the occipital lobe, which is primarily responsible for interpreting and processing visual information. Visual perception difference is a well-documented feature of ASD, often reflected in how individuals perceive faces, objects, and motion. The calcarine sulcus and cuneus play key roles in these visual functions and are closely linked to the altered visual processing commonly observed in ASD, such as heightened sensitivity to visual details alongside difficulties with integrating global visual information \citep{edgar2015resting,eilam2016neuroanatomical,zoltowski2021cortical}. 

\begin{table}[t!]
\centering
\caption{Trait classification accuracy (misclassification error rate) for the ABIDE study, and trait prediction accuracy (mean squared error) for the HCP study.}
\label{tab:real-data}
\resizebox{\textwidth}{!}{
\begin{tabular}{|c|c|cccc|} \hline
& (a) ABIDE & \multicolumn{4}{|c|}{(b) HCP} \\ \hline
& & Trait (i) & Trait (ii) & Trait (iii) & Trait (iv) \\ \hline
\method  & 39.9 (0.9) & 0.0201(0.0009) & 0.0207(0.0009) & 0.1149(0.0133) & 0.2102(0.0065) \\ \hline
sPCA & 41.5 (1.8) & 0.0211(0.0015) & 0.0223(0.0011) & 0.1211(0.0092) & 0.2187(0.0039) \\
\hline
\end{tabular}
}
\end{table}

\subsection{Human Connectome Project}

The second example is the Human Connectome Project (HCP), which aims to map the structural and functional connectivity networks of human brain and to understand their associations with behavioral traits \citep{VanEssen2013}. The dataset we analyzed consists of 924 subjects. Each subject received a diffusion tensor imaging (DIT) scan, which is summarized as a $97 \times 97$ structural connectivity network, with nodes corresponding to 68 cortical surface ROIs and 19 subcortical ROIs following the Desikan-Killiany brain atlas, and edges measuring the numbers of fibers connecting all pairs of ROIs \citep{liu2021graph}. In addition, for each subject, four trait scores were recorded, including (i) the picture vocabulary test score, (ii) the oral reading recognition test score, (iii) the crystallized composite score, and (iv) the cognition total composite score. 

We applied \method  and sPCA to this data, and carried out two downstream tasks: predicting the subject's trait scores, and identifying potentially important brain regions. We transformed the number of connecting fibers by $(1+a)^{-1}$ and the trait score by $\log(a+1)$. We randomly split the data, using 60\% for training and the remaining 40\% for testing. We repeated the split 50 times, and report the results based on 50 replications. We employed a support vector regression for subject trait prediction. We chose $r=19$ as the dimension of the latent space, by the percentage of explained variance. 

Table \ref{tab:real-data}(b) reports the mean squared error (MSE) on the testing data, where we see that \method  again outperforms sPCA, especially for the first two traits. We did not compare with the method of \citet{liu2021graph} because they used additional information other than structural connectivity, and used different training-test sample split. Figure \ref{fig:real-data}(b) reports the sparse edge adjacency matrix when setting the sparsity level at $s=100$. There are four nodes with noticeably larger degrees, corresponding to four brain regions: left superior temporal cortex, left insula, right banks of superior temporal sulcus, and right superior temporal cortex. These regions are closely related to reading, speech, comprehension, and cognition. For instance, superior temporal cortex is heavily involved in auditory processing and speech perception, and superior temporal sulcus is important for voice encoding \citep{rupp2022neural}. Meanwhile, the insula plays a key role in emotion regulation and cognitive control \citep{craig2009you}.

\bibliographystyle{apalike}
\bibliography{ref-contrast}

\newpage
\appendix

\clearpage
\begin{center}
  {\Large \bfseries Supplementary Appendix}
\end{center}
\vspace{1em}

\baselineskip=22pt

\begin{abstract}

\smallskip
\noindent
In this Appendix, Section \ref{sec:auxiliary lemmas} collects a number of auxiliary lemmas from the literature that facilitate our proofs. Sections \ref{sec:app proof estimation}, \ref{sec:app proof of section downstream}, and \ref{sec:app proof of comparepca} provide the proofs for the main theoretical results in Sections \ref{sec:theory}, \ref{sec:downstream}, and \ref{sec:comparePCA}, respectively. 

\end{abstract}

\section{Auxiliary Lemmas}
\label{sec:auxiliary lemmas}

The auxiliary lemmas include the ones related to matrix algebra, concentration inequalities, community detection, and miscellaneous technical lemmas.

\subsection{Lemmas related to matrix algebra}
\label{sec:app lemma linear algebra}

\begin{lem}[The Eckart-Young-Mirsky Theorem in \citet{eckart1936approximation}]
\label{lem: eckart}
For the singular value decomposition $A = U\Sigma V^T$, the best rank-$k$ approximation of $ A $ with respect to the Frobenius norm is given by $\rank_k(A) = \sum_{i=1}^{k} \sigma_i u_i v_i^T$. That is, for any matrix $B$ of rank at most $k$, $\|A - \rank_k(A) \|_F \leq \|A - B\|_F$. 
\end{lem}

\begin{lem}[Theorem 2 in \citet{yu2014useful}]
\label{lem: yu}
Let $\Sigma, \hat{\Sigma} \in \mathbb{R}^{p \times p}$ be symmetric matrices, with eigenvalues $\lambda_1 \geq \dots \geq \lambda_p$ and $\hat{\lambda}_1 \geq \dots \geq \hat{\lambda}_p$ respectively. Suppose $1 \leq r \leq s \leq p$, and $\min(\lambda_{r-1} - \lambda_r, \lambda_s - \lambda_{s+1}) > 0$, where $\lambda_0 = \infty$ and $\lambda_{p+1} = -\infty$. Let $d = s - r + 1$. Let $V = (v_r, v_{r+1}, \dots, v_s) \in \mathbb{R}^{p \times d}$, and $\hat{V} = (\hat{v}_r, \hat{v}_{r+1}, \dots, \hat{v}_s) \in \mathbb{R}^{p \times d}$ have orthonormal columns satisfying $\Sigma v_j = \lambda_j v_j$ and $\hat{\Sigma} \hat{v}_j = \hat{\lambda}_j \hat{v}_j$, $j = r, r+1, \dots, s$. Then
\vspace{-0.01in}
\begin{align*}
\|\sin\Theta(\hat{V}, V)\|_F &\leq \frac{2 \min\left( d^{\frac{1}{2}}\|\hat{\Sigma} - \Sigma\|_{\text{sp}}, \|\hat{\Sigma} - \Sigma\|_F \right)}{\min(\lambda_{r-1} - \lambda_r^*, \lambda_s - \lambda_{s+1})}.
\end{align*}
Moreover, there exists an orthogonal matrix $\hat{O} \in \mathbb{R}^{d \times d}$, such that
\begin{align*}
\|\hat{V}\hat{O} - V\|_F &\leq \frac{2^{\frac{3}{2}} \min\left( d^{\frac{1}{2}}\|\hat{\Sigma} - \Sigma\|_{\text{sp}}, \|\hat{\Sigma} - \Sigma\|_F \right)}{\min(\lambda_{r-1} - \lambda_r^*, \lambda_s - \lambda_{s+1})}.
\end{align*}
\end{lem}

\begin{lem}[Lemma 1 in \citet{zhang2022heteroskedastic}]
\label{lem: zhang projection contraction}
Suppose $U \in \sO_{m, r}$. Recall that $I(U^*) = \max_i \| e_i^\top U \|_2^2$, and $P_U = UU^\top$ is the projection matrix. Then for any matrix $A \in \mathbb{R}^{m \times m}$, we have $\| D(P_U (D(A))) \|_{\text{sp}} \leq I(U^*) \| D(A) \|_{\text{sp}}$, and $\| D(P_U A) \|_{\text{sp}} \leq \sqrt{I(U^*)} \| A \|_{\text{sp}}$. 
\end{lem}

\begin{lem}[Lemma 4 in \citet{zhang2022heteroskedastic}]
\label{lem: zhang off diagonal bound}
If $M \in \mathbb{R}^{p \times p}$ is any square matrix and $\Delta(M)$ is the matrix $M$ with diagonal entries set to $0$, then $\|\Delta(M)\|_{\text{sp}} \leq 2 \|M\|_{\text{sp}}$. \end{lem}

\begin{lem}[Lemma 6 in \citet{zhang2022heteroskedastic}]
\label{lem: zhang noise porjection bound}
Suppose $M, E \in \mathbb{R}^{p_1 \times p_2}$, $\operatorname{rank}(M) = r$, and $\hat{U} = \operatorname{SVD}_r(M + E)$, i.e., the leading $r$ left singular vectors of $M + E$. If $\hat{U}_\perp$ is the orthogonal complement of $\hat{U}$, then $\left\| P_{\hat{U}_\perp} M \right\|_{\text{sp}} \leq 2 \| E \|_{\text{sp}}$, and $\left\| P_{\hat{U}_\perp} M \right\|_F \leq 2 \min \left\{ \sqrt{r} \| E \|, \| E \|_F \right\}$. 
\end{lem}

\begin{lem}[Weyl's inequality in \citet{Weyl1912}]
\label{lem: Weyl}
Let $A, B$ be two $p \times p$ Hermitian matrices and $\lambda_i(A)$ be the $i$th eigenvalue of $A$. Then, for $1 \leq l, j \leq p$, 
\begin{align*}
    \lambda_l(A + B) &\leq \lambda_j(A) + \lambda_k(B), \quad \text{for } l \geq j + k - 1, \\
    \lambda_j(A) + \lambda_l(B) &\leq \lambda_{j + l - p}(A + B), \quad \text{for } j + l \geq p.
\end{align*}
\end{lem}

\begin{lem}[Lemma 5.4 in \citet{tu2016lowranksolutionslinearmatrix}]
\label{lem: tu dist and UUtop equivalence}
For any \( U, X \in \mathbb{R}^{p \times r} \), we have
\[
\dist^2(U, X) \leq \frac{1}{2(\sqrt{2} - 1) \sigma_r^2(X)} \norm{UU^\top - XX^\top}_F^2.
\]
\end{lem}

\begin{lem}[Theorem 2 in \citet{ten1977orthogonal}]
\label{lem: ten symmetric of XH^TL}
The function \(g(H) = \operatorname{tr}(H^\top X^\top L)\), where \(H\) varies without restriction over the set of orthonormal matrices of order \(r \times r\), is maximized if and only if \(H^\top X^\top L\) is symmetric and positive semi-definite (SPSD). Let \(X^\top L = PDQ^{*^\top}\) be an Eckart-Young decomposition of \(X^\top L\). Letting \(H = PQ^{*^\top}\), then \(H^\top X^\top L = Q^* D Q^{*^\top}\). Besides, \(Q^* D Q^{*^\top}\) is SPSD, which is both necessary and sufficient for attaining the best least-squares fit.
\end{lem}

\begin{lem}[Theorem 3.3 in \citet{vu2013fantope}]
\label{lem: vu deterministic error bound}
If $\lambda \geq \|\hat{\Sigma} - \Sigma\|_{\infty, \infty}$ and $s \geq \|\hat{U} \hat{U}^\top - U_{\Sigma} U_{\Sigma}^\top\|_{2, 0}$, then
\begin{equation*}
\norm{\hat{U} \hat{U}^\top - U_{\Sigma} U_{\Sigma}^\top}_{\text{sp}} \leq \frac{4s\lambda}{\sigma_{(r)}^2 - \sigma_{(r+1)}^2}.
\end{equation*}
\end{lem}

\subsection{Lemmas related to concentration inequalities}
\label{sec:app lemma concentration inequality}

\begin{lem}[Lemma 2 in \citet{zhang2022heteroskedastic}]
\label{lem: zhang concentration bound}
Suppose $E \in \mathbb{R}^{p_1 \times p_2}$ has independent sub-Gaussian entries, $\mathrm{Var}(E_{ij}) = \sigma_{ij}^2$, $\sigma_{ij}^2 = \max_{i}\sum_{j}\sigma_{ij}^2 = \max_{j}\sum_{i}\sigma_{ij}^2 = \max_{ij} \sigma_{ij}^2$. Suppose
\[
\left\|\frac{E_{ij}}{\sigma_{ij}}\right\|_{\psi_2} := \max_{q \geq 1} q^{-\frac{1}{2}} \left(\mathbb{E} \left| \frac{E_{ij}}{\sigma_{ij}} \right|^q\right)^{1/q} \leq \kappa.
\]
Let $V \in \sO_{p_2,r}$ be a fixed orthogonal matrix. Then
\begin{align*}
\mathbb{P} \Big( \| EV \|_{\text{sp}} \geq 2 (\sigma_C + x) \Big) &\leq 2 \exp \left( 5r - \min \left\{ \frac{x^4}{\kappa^4 \sigma_{(1)}^2 \sigma_C^2}, \frac{x^2}{\kappa^2 \sigma_{(1)}^2} \right\} \right), \\
        \E \qty[\| EV \|_{\text{sp}}] &\lesssim \sigma_C + \kappa r^{-\frac{1}{4}}(\sigma_{(1)}\sigma_C)^{\frac{1}{4}} + \kappa r^{\frac{1}{2}}\sigma_{(1)}.
\end{align*}
\end{lem}

\begin{lem}[Theorem 5.39 in \citet{vershynin2011introductionnonasymptoticanalysisrandom}]
\label{lem: vershynin subgaussian row}
Let $A$ be an $N \times n$ matrix whose rows $A_i$ are independent sub-Gaussian isotropic random vectors in $\mathbb{R}^n$. Then for every $x \geq 0$, with probability at least $1 - 2 \exp(-c x^2)$,
\begin{equation*}
        \sqrt{N} - C \sqrt{n} - x \leq \sigma_{\min}(A) \leq \sigma_{\max}(A) \leq \sqrt{N} + C \sqrt{n} + x, 
\end{equation*}
where $C = C_K$, $c = c_K > 0$ depend only on the sub-Gaussian norm $K = \max_i \| A_i \|_{\psi_2}$ of the rows.
\end{lem}

\begin{lem}[Theorem 6.5 in \citet{wainwright2019high}]
\label{lem: wainwright concentration bound prob}
There are universal constants $\{c_j\}_{j=0}^3$, such that, for any row-wise $\sigma$-sub-Gaussian random matrix $X \in \mathbb{R}^{n \times d}$, the sample covariance $\hat{\Sigma} = \frac{1}{n} \sum_{i=1}^n x_i x_i^\top$ satisfies the bounds
\begin{equation*}
        \E \left[ e^{\lambda \| \hat{\Sigma} - \Sigma \|_{\text{sp}}} \right] \leq e^{c_0 \lambda^2 \sigma^4 / n + 4 d}, \quad \text{for all } |\lambda| < \frac{n}{64 e^2 \sigma^2}.
\end{equation*}
Consequently, 
\begin{equation*}
\P \qty( \frac{\| \hat{\Sigma} - \Sigma \|_{\text{sp}}}{\sigma^2} \geq c_1 \left( \sqrt{\frac{d}{n}} + \frac{d}{n} \right) + x ) \leq c_2 e^{-c_3 n \min \{ x, x^2 \}}, \quad \text{for all } x \geq 0.
\end{equation*}
\end{lem}

\begin{lem}[Theorem 6 in \citet{cai2022nonasymptotic}]
\label{lem: cai}
    Suppose $Z$ is a $p_1 \times p_2$ random matrix with independent mean-zero sub-Gaussian entries. If there exist $\sigma_1, \dots, \sigma_p$ such that $\left\| Z_{ij}/\sigma_{i} \right\|_{\psi_2} \leq C_k$ for constant $C_k > 0$, then
    \begin{align*}
        \E \qty[\|ZZ^T - \mathbb{E}ZZ^T\|_{\text{sp}} ] &\lesssim \sum_{i} \sigma_i^2 + \sqrt{p_2 \sum_{i} \sigma_i^2}  \cdot \max_i \sigma_i.
    \end{align*}
\end{lem}

\begin{lem}[Lemma 1 in \citet{jin2019shortnoteconcentrationinequalities}]
\label{lem: jin norm concentration bound}Suppose a random vector $X \in \sR^d$ is $(\sigma/\sqrt{d})$-sub-Gaussian, then we have
\begin{equation*}
    \P \qty(\|X - \E[X]\|_2 \geq x) \leq 2e^{-\frac{x^2}{16\sigma^2}}
\end{equation*}
\end{lem}

\subsection{Lemmas related to community detection}
\label{sec:app lemma community detection}

\begin{lem}[Lemma 5.3 in \citet{Lei_2015}]
\label{lem: lei approximate k means error bound}
For $\epsilon > 0$ and any two matrices $\hat{\Gamma}, \Gamma \in \mathbb{R}^{p \times G}$, such that $\Gamma = \Theta Y$, where $Y \in \mathbb{R}^{G \times G}$ and $\Theta\in \mathbb{M}_{v,G}$, the set of membership matrices. Let $(\hat{\Theta}, \hat{Y})$ be a $(1+\epsilon)$-approximate solution to the k-means problem, which means it satisfies $\|\hat{\Theta} \hat{Y} - \hat{\Gamma}\|_F^2 \leq (1 + \epsilon) \min_{\Theta \in \mathbb{M}_{v,G}, Y \in \mathbb{R}^{G \times G}} \|\Theta Y - \hat{\Gamma}\|_F^2$. Define $\tilde{\Gamma} = \hat{\Theta} \hat{Y}$. For any $\delta_g \leq \min_{l \neq g} \|Y_{l *} - Y_{g *}\|$, define $S_g = \{i \in H_g(\Theta): \| \tilde{\Gamma}_{i *} - \Gamma_{i *}\| \geq \delta_g / 2\}$, where for any matrix $A$, $A_{\mathcal{I} *}$ denotes the submatrix consisting of rows indexed by $\mathcal{I}$. Then, 
\begin{equation} \label{eq: miserror Sk}
\sum_{g=1}^G |S_G| \delta_g^2 \leq (16 + 8\epsilon)\| \hat{\Gamma} - \Gamma \|_F^2.
\end{equation}
Moreover, if
\begin{equation} \label{eq: hatgamma - gamm}
(16 + 8\epsilon)\| \hat{\Gamma} - \Gamma \|_F^2 / \delta_g^2 < p_g, \quad \text{for all } g,
\end{equation}
then there exists a $G \times G$ permutation matrix $J$, such that $\hat{\Theta}_{H *} = \Theta_{H *} J$, where $H = \bigcup_{g=1}^G (H_G \setminus S_g)$.
\end{lem}

\begin{lem}[Lemma 2.1 in \citet{Lei_2015}]
\label{lem: lei basic eigen sbm}
Suppose $B \in \R^{G \times G}$ is symmetric and of full rank, and $\Theta \in \R^{p \times G}$ is a membership matrix. Let $\Gamma D \Gamma^\top$ be the eigen-decomposition of $L = \Theta B \Theta^\top$. Then $\Gamma = \Theta Y$ where $Y \in \mathbb{R}^{G \times G}$ and
\begin{equation*}
\|Y_{g *} - Y_{l *}\| = \sqrt{p_g^{-1} + p_l^{-1}}, \quad \text{for all } 1 \leq g < l \leq G.
\end{equation*}
\end{lem}

\begin{lem}[Lemma 5.1 in \citet{Lei_2015}]
\label{lem: lei principal subspace perturbation}
Suppose $L \in \mathbb{R}^{p \times p}$ is a rank $G$ symmetric matrix with smallest nonzero singular value $\lambda_G$. Let $A$ be any symmetric matrix, and $\hat{\Gamma}, \Gamma \in \mathbb{R}^{p \times G}$ be the $G$ leading eigenvectors of $A$ and $L$, respectively. Then there exists a $G \times G$ orthogonal matrix $O$, such that
\begin{equation*}
\norm{\hat{\Gamma} - \Gamma O}_F \leq \frac{2\sqrt{2G}}{\lambda_G} \norm{A - L}_{\text{sp}}.
\end{equation*}
\end{lem}

\subsection{Miscellaneous lemmas}
\label{sec:app lemma others}

\begin{lem}[Modified from Lemma B.19 in \citet{ji2023power}]
\label{lem: ji classification risk}
Suppose $O \in \sO_{d, r}$, for any $\hat{O} \in \sO_{d, r}$, with $\| \sin \Theta(\hat{O} , O) \|_{\text{sp}} \leq \frac{\sigma_{(1)}^2}{\kappa_\Sigma} \wedge \frac{1}{2}$. Then, 
\begin{multline*}
\inf_{w \in \mathbb{R}^r} \mathbb{E}_0\qty[l_c\qty(\mathbbm{1}\left\{F(w^\top \hat{O}^\top x_0) \geq \frac{1}{2}\right\})]  - \inf_{w \in \mathbb{R}^r} \mathbb{E}_0\qty[l_c\qty(\mathbbm{1}\left\{F(w^\top O^\top x_0) \geq \frac{1}{2}\right\})] \lesssim \\
\left[ \left\{ \kappa_\Sigma \qty(1 + \frac{1}{\sigma_{(1)}^2}) \right\}^3 +  \frac{\kappa_\Sigma}{\sigma_{(1)}^2} \qty(1 + \sigma_{(1)}^2)^2 \right] \norm{\sin \Theta(\hat{O}, O^*)}_{\text{sp}}.
\end{multline*}
\end{lem}

\begin{lem}[Lemma B.20 in \citet{ji2023power}]
\label{lem: ji classification risk lower bound}
Suppose $Q^* \in \sO_{d, r}$, for any $\hat{O} \in \sO_{d, r}$, with ${\sigma_{(1)}^2}{(1 + \sigma_{(1)}^2)} - \kappa_\Sigma \qty(r - \norm{\sin \Theta(\hat{O}, O)}_F^2) \geq 0$. Then,
\begin{multline*}
     \inf_{w \in \mathbb{R}^r} \mathbb{E}_0\qty[l_c\qty(\mathbbm{1}\left\{F(w^\top \hat{O}^\top x_0) \geq \frac{1}{2}\right\})]  - \inf_{w \in \mathbb{R}^r} \mathbb{E}_0\qty[l_c\qty(\mathbbm{1}\left\{F(w^\top O^\top x_0) \geq \frac{1}{2}\right\})] \gtrsim \\
    \frac{\qty(\sigma_{(1)}^2 + 1)^{\frac{3}{2}}}{\qty(\sigma_{(1)}^2 + \kappa_\Sigma)^{\frac{3}{2}}\sigma_{(1)}^2} \left\{ \frac{\sigma_{(1)}^2}{1 + \sigma_{(1)}^2} - \kappa_\Sigma \qty(r - \norm{\sin \Theta(\hat{O}, O)}_F^2) \right\}.
\end{multline*}
\end{lem}

\begin{lem}[Lemma B.7 in \citet{gao2023sparsegcathresholdedgradient}]
\label{lem: gao gradient descent}
Define \(L^*\) as a global minimizer of the function \(f(L)\), and $H_X = \argmin_{H \in \sO_{r,r}} \|X H - L^*\|_F$. Define a function \(f(L)\) to be \(\beta\)-smooth at \(L\), if for all \(Z\), we have
\[
\operatorname{vec}(Z)^\top \nabla^2 f(L) \operatorname{vec}(Z) \leq \beta \|Z\|_F^2.
\]
Suppose that \(f\) is \(\beta\)-smooth within a ball \(\mathcal{B}(L^*) = \{L : \|L - L^*\|_F \leq R\}\) and that \(\nabla f(L)P = \nabla f(LP)\) for any orthonormal matrix \(P\). Assume that for any \(L \in \mathcal{B}(L^*)\) and any \(Z\), we have $\operatorname{vec}(ZH_Z - L^*)^\top \nabla^2 f(L) \operatorname{vec}(ZH_Z - L^*) \geq \alpha \|ZH_Z - L^*\|_F^2$. In addition, if \(\eta \leq \frac{1}{\beta}\), then using gradient descent with \(\dist(L_t, L^*) \leq R\), we have $\dist^2(L_{t+1}, L^*) \leq (1 - \alpha \eta) \dist^2(L_t, L^*)$. Moreover, with \(\dist(L_0, L^*) \leq R\), we have $\dist^2(L_t, L^*) \leq (1 - \alpha \eta)^t \dist^2(L_0, L^*)$. 
\end{lem}

\section{Proofs for Section \ref{sec:theory}}
\label{sec:app proof estimation}

We first present the proofs for the main theoretical results in Section \ref{sec:theory}. We introduce a number of supporting lemmas along the way, and defer the proofs of those lemmas to the end of this section.

\subsection{Proof of Theorem \ref{prop: distance V and Vtk}}
\label{sec:proof of prop distance V and Vtk}

\begin{proof}
As mentioned in the Section \ref{sec:estimation}, the coefficient $1/8$ in \eqref{loss: cl conti triplet} was chosen for presentation simplicity. We now present a general version of the loss in \Eqref{loss: cl conti triplet} by introducing a general parameter $\lambda$ to control the penalty strength. 
\begin{multline}
\label{loss: cl conti triplet V}
\mL(V; \mathcal{D}, A) = - \frac{1}{n} \sum_i \langle V^\top A x_i, V^\top (I - A) x_i \rangle + \\\frac{1}{n^2} \sum_{i, j} \langle V^\top A x_i, V^\top (I - A) x_j \rangle + \frac{\lambda}{8} \|V V^\top\|_F^2.
\end{multline}
We will analyze this loss throughout the proof, and obtain the formulation in \eqref{loss: cl conti triplet}, by setting $\lambda = 1$. Correspondingly, at the inner iteration $t$ of the outer iteration $k$, denote the parameter matrices as $\hat{V}_{(t)}^{(k)}, \tilde{V}_{(t)}^{(k)} \in \mathbb{R}^{d \times r}$. Define
\begin{equation} \label{eq: V defination}
V^* = \frac{Q^*}{\sqrt{\lambda}}.
\end{equation}
Since in this theorem we focus solely on the $k$th iteration, we simplify the notation by omitting the superscripts $(k), (k-1)$. Also, recall the expected loss, 
\begin{equation} \label{eq: defn of glv}
\gL(V) = \mathbb{E}_A[\mathcal{L}(V; \mathcal{D}, A)].
\end{equation}
Since we always consider the same dataset $\gD$, we abbreviate $\gL(V; \gD)$ as $\gL(V)$ for notational simplicity. 

First, we have the following lemma that helps simplifies  the loss function. 
\begin{lem} \label{lem: simplified loss of cl conti triplet transpose expectation}
If $A$ is generated by \eqref{eq: masking distribution} with the masking parameter $p =(p_1, \cdots, p_d)^{\top} \in [0, 1]^d$, while omitting the constant, the loss function \eqref{eq: defn of glv} can be written as,
\begin{equation} \label{loss: cl conti triplet transpose expectation simplified}
\mL(V) = \frac{\lambda}{2} \norm{V V^\top -\frac{1}{\lambda} \qty(\Delta \qty(M) + P^2 D\qty(M))}_F^2.
\end{equation}
where $M = \frac{1}{n} X X^\top - \frac{1}{n^2} X 1_n 1_n^\top X^\top, X \in \R^{d \times n} = (x_1, \cdots, x_n)$, and $P = \diag(p_1, \cdots, p_d)$.
\end{lem}

Without loss of generality, we write $Q^* = U^* \Lambda^{*^\frac{1}{2}}$. Define the row support of \(U^*\) as \(S\), and suppose \(\gI \supseteq S\). Define \(\hat{U}(\gI) \in \sR^{d \times r}\) as the solution of the optimization, 
\[
\max \, \langle N, F F^\top \rangle, 
\quad \text{ such that } F^\top F = I_r, \quad \supp(F) \subseteq \gI,
\]
where we rewrite $N = \Delta (M) + P^2 D(M)$.

We further define $\hat \Lambda_r(\gI)$ as the diagonal matrix formed by the first $r$ eigenvalues of $\qty(N)_{\gI\gI}$. Since \(\supp(\hat{U}(\gI)) \subseteq \gI\), we have that \(\hat{U}_{\gI *}(\gI)\) is the solution of
\[
\max \, \langle \qty(N)_{\gI\gI}, L L^\top \rangle 
\quad \text{such that } L^\top L = I_r,
\]
where, for any matrix $A$, $A_{\mathcal{I} *}$ denotes the submatrix consisting of rows indexed by $\mathcal{I}$. Moreover, \(\supp(U^*) = S \subseteq \gI\), then \(U_{\gI *}\) is the solution of
\[
\max \, \langle (Q^* Q^{*^\top})_{\gI\gI}, L L^\top \rangle 
\quad \text{such that } L^\top L = I_r.
\]

Moreover, define \(\hat{V}(\gI) = \hat{U}(\gI) \qty(\frac{\hat{\Lambda}_r(\gI)}{\lambda})^{\frac{1}{2}}\), and recall \(V^*\) as defined in \eqref{eq: V defination}. By Lemma \ref{lem: tu dist and UUtop equivalence}, we have,
\begin{equation}
\begin{aligned} \label{eq: distance between VhatSt and V}
    \dist(\hat{V}(\gI), V^*) &\leq \frac{1}{\sqrt{2\sqrt{2} - 2} \lambda_r^* \sqrt{\lambda}} \norm{\hat{U}(\gI) \hat{\Lambda}_r(\gI) \hat{U}(\gI)^\top - U^* \Lambda^* U^{*^\top}}_F \\
    &\leq \frac{\sqrt{2r}}{\sqrt{2\sqrt{2} - 2} \lambda_r^* \sqrt{\lambda}} \norm{\hat{U}(\gI) \hat{\Lambda}_r(\gI) \hat{U}(\gI)^\top - U^* \Lambda^* U^{*^\top}}_{\text{sp}} \\
    &\leq \frac{2 \sqrt{2r}}{\sqrt{2\sqrt{2} - 2} \lambda_r^* \sqrt{\lambda}} \norm{\qty(N - Q^* Q^{*^\top})_{\gI\gI}}_{\text{sp}}\\
    &<  \frac{9 \sqrt{r}}{2 \lambda_r^* \sqrt{\lambda}} W^{(k-1)}
\end{aligned}
\end{equation}

Next, we characterize the progress of the gradient descent step using the following lemma. Recall $W^{(k-1)}$ is defined as the masking-adjusted covariance matrix estimation error from \eqref{eq: W^{(k)}-1 defination}. We define the effective restricted set as $S_{(t)} = \supp(\tilde V_{(t)}) \cup \supp(\tilde V_{(t+1)}) \cup S$. 
\begin{lem} \label{lem: gradient descent}
Suppose \(W^{(k-1)} < {\lambda_r^{*2}} / {4}\), and the step size \(\eta\) is chosen such that \(\eta \leq \frac{1}{8 W^{(k-1)} + \frac{85}{12}\lambda_1^{*2}}\). If
\begin{equation} \label{eq: radius condition of Vt}
\dist\left( \hat{V}_{(t)}, \hat{V}(S_{(t)}) \right) \leq \frac{\lambda_r^{*2}}{12 \sqrt{\lambda (\lambda_1^{*2} +  W^{(k-1)})}},
\end{equation}
then after the gradient step, we have
\[
\dist^2\left( \hat{V}_{(t+1)}^o, \hat{V}(S_{(t)}) \right) \leq \left\{ 1 - \eta \qty(\frac{17}{12}\lambda_r^{*2} - 4R^{(k-1)}) \right\} \dist^2 \left( \hat{V}_{(t)}, \hat{V}(S_{(t)}) \right),
\]
where $\hat{V}_{(t+1)}^o \in \mathbb{R}^{d \times r}$ denotes a matrix that has the same entries as those in $\hat{V}_{(t+1)}$ on $S_{(t)} \times [r]$ and zeros elsewhere.
\end{lem}

Next, we present a lemma showing the property of truncation. 
\begin{lem}[Proposition B.6 in \citet{gao2023sparsegcathresholdedgradient}]
\label{lem: gao truncation}
Let \(V^*\) be as defined in \eqref{eq: V defination}. Suppose we perform hard thresholding by selecting the top $s$ elements of $\hat{V}_{(t+1)}$. Then we have
\[
\dist^2\left( V^*, \tilde{V}_{(t+1)} \right) \leq \left\{ 1 + 2 \sqrt{\frac{s^*}{s}} \left( 1 + \sqrt{\frac{s^*}{s}} \right) \right\} \dist^2\left( V^*, \hat{V}_{(t+1)} \right).
\]
\end{lem}

Now, we turn to the proof of this theorem. 

We first define the effective support in each step. Considering the situation in Lemma \ref{lem: gradient descent} and the effective restricted set \(S_{(t)}\), with the loss \eqref{loss: cl conti triplet transpose expectation simplified}, the gradient descent step restricted to $S_{(t)}$ can be viewed as
\[
\hat{V}_{(t+1),S_{(t)}*} = \tilde V_{(t),S_{(t)}*} - 2 \eta \qty(- N_{S_{(t)}S_{(t)}} \tilde V_{(t),S_{(t)}*} + \lambda \tilde V_{(t),S_{(t)}*}\tilde V_{(t),S_{(t)}*}^\top \tilde V_{(t),S_{(t)}*}).
\]

Note that applying hard thresholding on $\hat{V}_{(t+1)}^o$ is equivalent to applying hard thresholding on original $V_{(t+1)}$. This allows us to replace the intermediate update by $\hat{V}_{(t+1)}^o$ and still obtain the same output sequence $\hat{V}_{(t+1)}$. Thus, we will prove instead for the update using $N_{S_{(t)}S_{(t)}}$, then we can use Lemma \ref{lem: gradient descent}.

We achieve this by induction on $t$. Specifically, for $t = 1, 2, \ldots$, we will prove that $\tilde{V}_{(t)}$ satisfies the radius condition \eqref{eq: radius condition of Vt}, and that
\begin{equation}
\label{eq: induction for distance Vbart and V}
\dist\left( \tilde{V}_{(t)}, V^* \right) \leq \xi^{t-1} \dist(\tilde{V}_{(1)}, V^*) + \frac{9 \sqrt{r}}{(1 - \xi) \lambda_r^* \sqrt{\lambda}} W^{(k-1)}.
\end{equation}

We start with the base case of $t=1$. Since $\dist(\tilde{V}_{(1)}, V^*) \leq \dist(\tilde{V}_{(1)}, V^*)$, \eqref{eq: induction for distance Vbart and V} holds trivially. To check the radius condition \ref{eq: radius condition of Vt}, we have, by \eqref{eq: distance between VhatSt and V} and \(W^{(k-1)} \leq \frac{c_3\lambda_r^{*3}}{\lambda_1^*\sqrt{r}}\), 
\begin{align*}
\dist\left( \hat{V}(S_{(1)}), V^* \right) \leq \frac{9 \sqrt{r}}{2 \lambda_r^* \sqrt{\lambda}} W^{(k-1)} 
 < \frac{\lambda_r^{*2}}{24 \sqrt{\lambda (\lambda_1^{*2} + W^{(k-1)})}}.
\end{align*}
Moreover, by $\big\| \hat{Q}^{(0)}\hat{Q}^{(0)^\top} - Q^*Q^{*^\top} \big\|_F \leq {\lambda_r^{*3}}/{(20 \lambda_1^*)}$, and Lemma~\ref{lem: tu dist and UUtop equivalence}, we have
\begin{equation*}
\dist\left( \tilde{V}_{(1)}, \hat{V}(S_{(1)}) \right) \leq \dist\left( \tilde{V}_{(1)}, V^* \right) + \dist\left( \hat{V}(S_{(1)}), V^* \right) \leq \frac{\lambda_r^{*2}}{12 \sqrt{\lambda (\lambda_1^{*2} + W^{(k-1)})}}.
\end{equation*}

We then continue with the induction step. Suppose that $\tilde{V}_{(t)}$ satisfies the radius condition \eqref{eq: radius condition of Vt} and that the induction equation \eqref{eq: induction for distance Vbart and V} is satisfied at step $t$. We aim to show that \eqref{eq: radius condition of Vt} and \eqref{eq: induction for distance Vbart and V} hold for $\tilde V_{(t+1)}$. 

In the gradient step, Lemma \ref{lem: gradient descent} shows that under \eqref{eq: radius condition of Vt} on $\dist(\tilde{V}_{(t)}, \hat{V}(S_{(t)}))$, if we choose the step-size to be $\eta \leq \frac{1}{\beta}$, we have,
\begin{equation*}
\dist\left( \hat{V}_{(t+1)}^o, \hat{V}(S_{(t)}) \right) \leq \sqrt{1 - \eta \alpha} \dist\left( \tilde{V}_{(t)}, \hat{V}(S_{(t)}) \right),
\end{equation*}
where \(\alpha = \frac{4}{3}\lambda_r^{*2} \leq \frac{17}{12}\lambda_r^{*2} - 4 W^{(k-1)}\), \(\beta = \frac{43}{6}\lambda_1^{*2} \geq 8 W^{(k-1)} + \frac{85}{12}\lambda_1^{*2}\) given $W^{(k-1)} \leq \frac{c_3\lambda_r^{*3}}{\lambda_1^*\sqrt{r}}$.
By Lemma \ref{lem: gao truncation},
\begin{equation*}
\dist\left( \tilde{V}_{(t+1)}, V^* \right) \leq \sqrt{1 + 2 \sqrt{\frac{s^*}{s}} \qty(1 + \sqrt{\frac{s^*}{s}})} \dist\left( \hat{V}_{(t+1)}^o, V^* \right).
\end{equation*}
Therefore, 
\begin{align*}
\dist\left( \hat{V}_{(t+1)}^o, V^* \right) &\leq \dist\left( \hat{V}_{(t+1)}^o, \hat{V}(S_{(t)}) \right) + \dist\left( \hat{V}(S_{(t)}), V^* \right)\\
    &\leq \sqrt{1 - \eta \alpha} \dist\left( \tilde{V}_{(t)}, \hat{V}(S_{(t)}) \right) + \dist\left( \hat{V}(S_{(t)}), V^* \right)\\
    &\leq \sqrt{1 - \eta \alpha} \dist\left( \tilde{V}_{(t)}, V^* \right) + \qty(\sqrt{1 - \eta \alpha} + 1) \dist\left( \hat{V}(S_{(t)}), V^* \right).
\end{align*}
Combining with \eqref{eq: distance between VhatSt and V}, we have that,
\begin{multline*}
\dist\left( \tilde{V}_{(t+1)}, V^* \right) \leq \sqrt{\qty(1 + 2 \sqrt{\frac{s^*}{s}} \qty(1 + \sqrt{\frac{s^*}{s}}))\qty(1 - \eta \alpha)}\dist\left( \tilde{V}_{(t)}, V^* \right)\\
    + \sqrt{1 + 2 \sqrt{\frac{s^*}{s}} \qty(1 + \sqrt{\frac{s^*}{s}})}\qty(\sqrt{1 - \eta \alpha} + 1)\frac{9 \sqrt{r}}{2 \lambda_r^* \sqrt{\lambda}} W^{(k-1)}.
\end{multline*}
By taking \( s \geq \frac{64}{\alpha^2 \eta^2} s^*\), we ensure that
\begin{equation}
\label{eq: bound for xi}
    \xi \leq \sqrt{\qty( 1 + \frac{\alpha \eta}{4} + \frac{\alpha^2 \eta^2}{32}) \qty( 1 - \alpha \eta )} \leq \sqrt{\qty( 1 - \frac{3 \alpha \eta}{4} - \frac{7 \alpha^2 \eta^2}{32})} \leq \sqrt{1 - \frac{3\alpha \eta}{4}} < 1 - \frac{3 \alpha \eta}{8},
\end{equation}
where \(\xi = \sqrt{\qty(1 + 2 \sqrt{\frac{s^*}{s}} \qty(1 + \sqrt{\frac{s^*}{s}}))\qty(1 - \eta \alpha)}\). Therefore, 
\begin{align*}
\dist\left( \tilde{V}_{(t+1)}, V^* \right) &\leq \xi \dist\left( \tilde{V}_{(t)}, V^* \right) + \qty( 1 - \frac{3 \alpha \eta}{8} + \sqrt{1 + \frac{\alpha \eta}{4} + \frac{\alpha^2 \eta^2}{2}}) \frac{9 \sqrt{r}}{2 \lambda_r^* \sqrt{\lambda}} W^{(k-1)}\\
    &\leq \xi \dist\left( \tilde{V}_{(t)}, V^* \right) + \frac{9 \sqrt{r}}{\lambda_r^* \sqrt{\lambda}} W^{(k-1)}.
\end{align*}
From \eqref{eq: induction for distance Vbart and V} at the $t$ step,
\begin{align*}
\dist\left( \tilde{V}_{(t+1)}, V^* \right) &\leq \xi \qty(\xi^{t-1} \dist\left( \tilde{V}_{(1)}, V^* \right) + \frac{9 \sqrt{r}}{(1 - \xi) \lambda_r^* \sqrt{\lambda}} W^{(k-1)}) + \frac{9 \sqrt{r}}{\lambda_r^* \sqrt{\lambda}} W^{(k-1)}\\
    &= \xi^{t} \dist\left( \tilde{V}_{(1)}, V^* \right) + \frac{9 \sqrt{r}}{(1 - \xi) \lambda_r^* \sqrt{\lambda}} W^{(k-1)}.
\end{align*}
This proves that \eqref{eq: induction for distance Vbart and V} holds for $t + 1$ step. 

Furthermore, by \eqref{eq: bound for xi}, we have, 
\begin{align*}
\dist\left( \tilde{V}_{(t+1)}, \hat{V}(S_{(t+1)}) \right) &\leq \dist\left( \tilde{V}_{(t+1)}, V^* \right) + \dist\left( \hat{V}(S_{(t+1)}), V^* \right) \\
    &\leq \xi^t \dist\left( V_{(1)}, V^* \right) + \qty(\frac{9}{1 - \xi} + \frac{9}{2})\frac{\sqrt{r}}{\lambda_r^* \sqrt{\lambda}} W^{(k-1)}\\
    &\leq \frac{\lambda_r^{*2}}{24 \sqrt{\lambda (\lambda_1^{*2} + W^{(k-1)})}} + \qty(\frac{24}{\alpha \eta} + 2) \frac{\sqrt{r}}{\lambda_r^* \sqrt{\lambda}} W^{(k-1)}.
\end{align*}
By \(W^{(k-1)} \leq \frac{c_3\lambda_r^{*3}}{\lambda_1^*\sqrt{r}}\), we have
\begin{equation*}
\dist\left( \tilde{V}_{(t+1)}, \hat{V}(S_{(t+1)}) \right) \leq \frac{\lambda_r^{*2}}{12 \sqrt{\lambda (\lambda_1^{*2} + W^{(k-1)})}}.
\end{equation*}
By induction, we have that, 
\begin{align*}
\dist\left( \tilde{V}_{(t+1)}, V^* \right) & \leq \xi^t \dist\left( \tilde{V}_{(1)}, V^* \right) + \frac{24\sqrt{r}}{\alpha \eta\lambda_r^* \sqrt{\lambda}} W^{(k-1)} \\
& \leq \xi^t \dist\left( \tilde{V}_{(1)}, V^* \right) + \frac{3}{\lambda_r^*}\sqrt{\frac{sr}{s^*\lambda}} W^{(k-1)}.
\end{align*}

This completes the proof of Theorem \ref{prop: distance V and Vtk}.
\end{proof}

\subsection{Proof of Theorem \ref{thm: edge embeddings sparse}}

\begin{proof}
Suppose $s \geq s^*$. we define the event $\Omega_1$ as
\begin{multline*}
    \Omega_1 = \Bigg\{
    \norm{\qty(M - Q^* Q^{*^\top} - \Sigma)_{\gI\gI}}_{\text{sp}} 
    \leq c_1' (\lambda_1^{*2} + \sigma_{(1)}^{*2}) \qty(\sqrt{\frac{s \log d}{n}} + \frac{s^2 \log d}{n}),\\
    \text{for all } \gI \subset [d] \text{ with } |\gI| = 2s + s^*
    \Bigg\}.
\end{multline*}
The following lemma ensure $\Omega_1$ occurs with a high probability.

\begin{lem}
\label{lem: Omega1}
Suppose $2s + s^* < n$. There exist a constant $c_1' > 0$, such that $\P(\Omega_1) \geq 1 - ce^{-c s \log d}$.
\end{lem}

Under $\Omega_1$ happens, we consider the loss $\mL(V)$ in \eqref{loss: cl conti triplet transpose expectation simplified}, and perform Algorithm \ref{alg: edge embeddings conti sparse}.

Taking the masking parameter as \eqref{eq: masking p at step k} at the $k$th iteration, we have
\begin{equation*}
    \qty(P^{(k)^2} D(M) )_{e,e} = 
    \begin{cases}
        M_{e,e} & \text{when } N^{(k-1)}_{e,e} \geq M_{e,e} \\
        N^{(k-1)}_{e,e} & \text{when } N^{(k-1)}_{e,e} < M_{e,e}
    \end{cases},
\end{equation*}
where $P^{(k)} = \diag ( p^{(k)}_1, \cdots, p^{(k)}_d )$, and $\tilde N^{(k-1)} = \hat{Q}^{(k-1)} \hat{Q}^{(k-1)^\top}$. Denote $\tilde N^{(k-1)} = \Delta (M) + P^{(k)^2} D(M)$, and $\tilde N^{(0)} = N^{(0)} = \Delta (M)$.

After establishing Theorem \ref{prop: distance V and Vtk} for the $k$th iteration, we set \(\lambda = 1\) and analyze \( W^{(k)} = \norm{\qty(\tilde{N}^{(k)} - Q^* Q^{*^\top})_{\gI \gI}}_{\text{sp}}\) as \(k\) increases, leveraging Theorem \ref{prop: distance V and Vtk}. The goal is to derive a recursive relationship between \(W^{(k-1)}\) and \(W^{(k)}\), and subsequently express \(W^{(k)}\) in terms of \(T_0\). We claim that
\begin{equation}
\label{eq: W^{(k)}}
    W^{(k)} \leq 6 T_0 + \frac{c_2's^*\lambda_1^{*2}}{2^k sr},
\end{equation}
where \(T_0 = 2 c_1' (\lambda_1^{*2} + \sigma_{(1)}^{*2}) \qty(\sqrt{\frac{s \log d}{n}} + \frac{s^2 \log d}{n})\) and \(c_2'\) is a sufficiently small constant.

We start with the base case of \(k = 1\). From Lemma \ref{lem: zhang off diagonal bound}, we have
\begin{align*}
    W^{(0)} & = \|(\tilde{N}^{(0)} - Q^* Q^{*^\top})_{\gI \gI}\|_{\text{sp}} \leq \|\qty(\Delta (M - Q^* Q^{*^\top} - \Sigma))_{\gI \gI}\|_{\text{sp}} + D\qty(Q^* Q^{*^\top}) \|_{\text{sp}}\\
    & \leq T_0 + \lambda_1^{*2} I(U^*) \leq T_0 + \frac{c_2\lambda_1^{*2}}{r} \leq 6 T_0 + \frac{c_2' s^* \lambda_1^{*2}}{s r},
\end{align*}
by \(c_2 \leq \frac{c_2's^*}{s}\). Thus, \(W^{(0)}\) satisfies \eqref{eq: W^{(k)}}.

Next we verify that \(W^{(0)}\) meets the condition in Theorem \ref{prop: distance V and Vtk}. Since \(W^{(0)} \leq 6 T_0 + \frac{c_2\lambda_1^{*2}}{r}\), and by that \(s \log d \ll n \) and \(c_2\) is sufficiently small, it follows that \(W^{(0)} \leq \frac{c_3\lambda_r^{*3}}{\lambda_1^*\sqrt{r}}\).

For our initializer $\hat Q^{(0)}$, the assumption $\big\| \hat{Q}^{(0)}\hat{Q}^{(0)^\top} - Q^*Q^{*^\top} \big\|_F \leq {\lambda_r^{*3}}/{(20 \lambda_1^*)}$ required by Theorem \ref{prop: distance V and Vtk} is automatically satisfied. Given that $W^{(0)} \leq \frac{c_3\lambda_r^{*3}}{\lambda_1^*\sqrt{r}}$, the base case is verified. Therefore, both $W^{(0)}$ and $\hat Q^{(0)}$ satisfy the initialization conditions in Theorem \ref{prop: distance V and Vtk}.

We then continue with the induction step. Suppose that 
\[
W^{(k-1)} \leq 6 T_0 + \frac{c_2's^*\lambda_1^{*2}}{2^{k - 1} s r} \leq \frac{c_3\lambda_r^{*3}}{\lambda_1^*\sqrt{r}}, \quad
\big\| \hat Q^{(k - 1)}\hat Q^{(k - 1)^\top} - Q^*Q^{*^\top} \big\|_F \leq {\lambda_r^{*3}}/{(20 \lambda_1^*)}.
\]
At the \(k\)th iteration, we run
\[
T^{(k)} = \Omega\qty(\frac{\log \qty(\frac{1}{\lambda_r^*}\sqrt{\frac{sr}{s^*}} W^{(k-1)})}{\log \qty( 1 - \frac{\lambda_r^{*2} \eta}{2})})
\]
By Assumptions \ref{asm: Sigma} and \ref{asm: Lambda}, we have
\begin{equation*}
    T^{(k)} = \Omega\qty(\log n \wedge \log \qty(2^k r)).
\end{equation*}
Taking the output \(\tilde{V}^{(k)}_{(T^{(k)})}\) as \(\hat Q^{(k)}\). By applying Theorem \ref{prop: distance V and Vtk}, we obtain that 
\[
\dist\left( \hat Q^{(k)}, Q^* \right) \leq \frac{49}{16\lambda_r^*}\sqrt{\frac{sr}{s^*}} W^{(k-1)}.
\]
This implies that there exists an orthogonal matrix \( O \in \sO_{r, r}\), such that
\begin{equation}
\label{eq: distance between hatQkO and Q}
\dist\left( \hat Q^{(k)}, Q^* \right) = \norm{\hat {Q}^{(k)} O - Q^*}_{F} \leq \frac{49}{16\lambda_r^*}\sqrt{\frac{sr}{s^*}} W^{(k-1)}
\end{equation}
Since \(c_2'\) is sufficiently small,
\begin{equation}
\begin{aligned}
\label{eq: distance between hatQk and Q}
    \norm{\hat{Q}^{(k)} \hat{Q}^{(k)^\top} - Q^* Q^{*^\top}}_{F} 
    &= \norm{\hat{Q}^{(k)} O O^\top \hat{Q}^{(k)^\top} - U^* U^{*^\top}}_{F} \\
    &= \norm{\qty(\hat{Q}^{(k)} O - Q^*) O^\top \hat{Q}^{(k)^\top} + Q^* \qty(O^\top \hat{Q}^{(k)^\top} - Q^{*^\top})}_{F} \\
    &\leq \qty(\|\hat{Q}^{(k)}\|_{\text{sp}} + \|Q\|_{\text{sp}}) \|\hat{Q}^{(k)} O - Q\|_{F}\\
    &\leq \frac{25\lambda_1^*}{4\lambda_r^*} \sqrt{\frac{sr}{s^*}} W^{(k-1)}
\end{aligned}
\end{equation}
Then by Lemma \ref{lem: yu}, we have, 
\begin{equation}
\label{eq: distance between projectionhatQk and Q}
    \norm{\hat{Q}^{(k)} \qty(\hat{Q}^{(k)^\top} \hat{Q}^{(k)})^{-1} \hat{Q}^{(k)^\top} - U^* U^{*^\top}}_{F} \leq \frac{\sqrt{2}}{\lambda_r^{*2}}\|\hat{Q}^{(k)} \hat{Q}^{(k)^\top} - Q^* Q^{*^\top} \|_F \leq \frac{25\sqrt{2}\lambda_1^*}{4\lambda_r^{*3}} \sqrt{\frac{sr}{s^*}} W^{(k-1)}
\end{equation}
Since
\begin{multline}
\label{eq: Rk T0 and Ntildek QQtop gIgI}
    W^{(k)} = \norm{\qty(\tilde{N}^{(k)} - Q^* Q^{*^\top})_{\gI \gI}}_{\text{sp}} \leq \norm{\qty(\Delta (M - Q^* Q^{*^\top} - \Sigma))_{\gI \gI}}_{\text{sp}} \\+  \norm{\qty(D (\tilde{N}^{(k)} - Q^* Q^{*^\top}))_{\gI \gI} }_{\text{sp}} 
    = T_0 + \norm{\qty(D (\tilde{N}^{(k)} - Q^* Q^{*^\top}))_{\gI \gI} }_{\text{sp}},
\end{multline}
we only need to analyze the diagonal term.

First, we bound the difference between 
\[
\norm{\qty(D (\tilde{N}^{(k)} - Q^* Q^{*^\top}))_{\gI \gI} }_{\text{sp}} \;\; \text{ and } \;\; \norm{\qty(D (N^{(k)} - Q^* Q^{*^\top}))_{\gI \gI} }_{\text{sp}}.
\]
Since $D (\tilde{N}^{(k)})$, $D (Q^* Q^{*^\top})$ and $D(N^{(k)})$ are all diagonal and positive semi-definite, we analyze the difference by computing each entry.

If $N^{(k)}_{e,e} \geq (Q^* Q^{*^\top})_{e,e}$, we have
\begin{equation*}
    \left| \min \{ M_{e,e}, N^{(k)}_{e,e} \} - (Q^* Q^{*^\top})_{e,e} \right| \leq \left| N^{(k)}_{e,e} - (Q^* Q^{*^\top})_{e,e} \right|, \text{i.e.}
\end{equation*}
\begin{equation*}
    \left| \tilde{N}^{(k)}_{e,e} - (Q^* Q^{*^\top})_{e,e} \right| \leq \left| N^{(k)}_{e,e} - (Q^* Q^{*^\top})_{e,e} \right|.
\end{equation*}
If $D(N^{(k)})_{e,e} < D(Q^* Q^{*^\top})_{e,e}$, only when $D(M)_{e,e} < D(N^{(k)})_{e,e} < D(Q^* Q^{*^\top})_{e,e}$, we have
\begin{equation*}
    D(Q^* Q^{*^\top})_{e,e} - D(M)_{e,e} > D(Q^* Q^{*^\top})_{e,e} - D(N^{(k)})_{e,e}.
\end{equation*}
In this situation,
\begin{align*}
    D(Q^* Q^{*^\top})_{e,e} - D(M)_{e,e} &\leq D(Q^* Q^{*^\top} + \Sigma)_{e,e} - D(M)_{e,e} \leq \norm{\qty(D(Q^* Q^{*^\top} + \Sigma - M))_{\gI \gI}}_{\text{sp}}\\
    &\leq \norm{\qty(Q^* Q^{*^\top} + \Sigma - M)_{\gI \gI}}_{\text{sp}} \leq \frac{T_0}{2}.
\end{align*}
Combining the two situations, we have
\begin{equation}
\label{eq: Nktilde and Nk gIgI}
    \norm{\qty(D (\tilde{N}^{(k)} - Q^* Q^{*^\top}))_{\gI \gI} }_{\text{sp}} \leq \norm{\qty(D (N^{(k)} - Q^* Q^{*^\top}))_{\gI \gI} }_{\text{sp}} + \frac{T_0}{2}.
\end{equation}

Second, we bound \(\norm{\qty(D (N^{(k)} - Q^* Q^{*^\top}))_{\gI \gI} }_{\text{sp}}\). Note that
\begin{equation}
\begin{aligned}
\label{eq: Nk gIgI and hatQk}
    \norm{\qty(D \qty(N^{(k)} - Q^* Q^{*^\top}))_{\gI \gI} }_{\text{sp}} &= \norm{\qty(D\qty(\hat{Q}^{(k)} \hat{Q}^{(k)^\top} - Q^* Q^{*^\top}))_{\gI \gI}}_{\text{sp}}\\ &\leq \norm{D\qty(\hat{Q}^{(k)} \hat{Q}^{(k)^\top} - Q^* Q^{*^\top})}_{\text{sp}}.    
\end{aligned}
\end{equation}

We focus on \(\norm{D\qty(\hat{Q}^{(k)} \hat{Q}^{(k)^\top} - Q^* Q^{*^\top})}_{\text{sp}}\). Define the projection matrix \(P_Q = U^* U^{*^\top}\) and \(P_{Q^{(k)}} = \hat{Q}^{(k)} \qty(\hat{Q}^{(k)^\top} \hat{Q}^{(k)})^{-1} \hat{Q}^{(k)^\top}\), we have \(P_{Q^{(k)}} \hat{Q}^{(k)} \hat{Q}^{(k)^\top} = \hat{Q}^{(k)} \hat{Q}^{(k)^\top}\). Note the following decomposition,
\begin{align*}
    D\qty(\hat{Q}^{(k)} \hat{Q}^{(k)^\top} - Q^* Q^{*^\top}) =& D\qty(P_Q \qty(\hat{Q}^{(k)} \hat{Q}^{(k)^\top} - Q^* Q^{*^\top})) - D\qty(P_{\hat{Q}_\perp^{(k)}} Q^* Q^{*^\top})\\
    &+ D\qty((P_{\hat{Q}^{(k)}} - P_Q) \qty(\hat{Q}^{(k)} \hat{Q}^{(k)^\top} - Q^* Q^{*^\top})).
\end{align*}
By Lemma \ref{lem: zhang projection contraction} and \eqref{eq: distance between hatQk and Q}, we have
\begin{equation}
\begin{aligned}
\label{eq: PQ hatQkhatQktop and QQtop}
\norm{D\qty(P_Q \qty(\hat{Q}^{(k)} \hat{Q}^{(k)^\top} - Q^* Q^{*^\top}))}_{\text{sp}} 
&\leq \sqrt{I(U^*)} \norm{\hat{Q}^{(k)} \hat{Q}^{(k)^\top} - Q^* Q^{*^\top}}_{\text{sp}} \\
&\leq \frac{25\sqrt{2}\lambda_1^*}{4\lambda_r^*} \sqrt{\frac{sr}{s^*}I(U^*)} W^{(k-1)}.
\end{aligned}
\end{equation}
By Lemmas \ref{lem: zhang projection contraction}, \ref{lem: zhang noise porjection bound} and \eqref{eq: distance between hatQk and Q}, we have
\begin{equation}
\label{eq: PhatQk QQtop}
    \norm{D\qty(P_{\hat{Q}_\perp^{(k)}} Q^* Q^{*^\top})}_{\text{sp}} 
    \leq 2 \sqrt{I(U^*)} \norm{\hat{Q}^{(k)} \hat{Q}^{(k)^\top} - Q^* Q^{*^\top}}_{\text{sp}}
    \leq \frac{25\sqrt{2}\lambda_1^*}{2\lambda_r^*} \sqrt{\frac{sr}{s^*}I(U^*)} W^{(k-1)}.
\end{equation}
By \eqref{eq: distance between hatQk and Q} and \eqref{eq: distance between projectionhatQk and Q}, we have
\begin{equation}
\begin{aligned}
\label{eq: PhatQk and PQ hatQkhatQktop QQtop}
    & \norm{D\qty(P_{\hat{Q}^{(k)}} - P_Q) \qty(\hat{Q}^{(k)} \hat{Q}^{(k)^\top} - Q^* Q^{*^\top})}_{\text{sp}}
    \\ \leq &
     \norm{P_{\hat{Q}^{(k)}} - P_Q}_{\text{sp}} \norm{\hat{Q}^{(k)} \hat{Q}^{(k)^\top} - Q^* Q^{*^\top}}_{\text{sp}}\\
    \leq & \frac{25\sqrt{2}\lambda_1^*}{4\lambda_r^*} \sqrt{\frac{sr}{s^*}} W^{(k-1)} \norm{\hat{Q}^{(k)} \qty(\hat{Q}^{(k)^\top} \hat{Q}^{(k)})^{-1} \hat{Q}^{(k)^\top} - U^* U^{*^\top}}_{\text{sp}}\\
    \leq& \frac{625 \lambda_1^{*2} sr}{8 \lambda_r^{*4} s^*} W^{(k-1)^2}.
\end{aligned}
\end{equation}
Combining \eqref{eq: Nk gIgI and hatQk}, \eqref{eq: PQ hatQkhatQktop and QQtop}, \eqref{eq: PhatQk QQtop} and \eqref{eq: PhatQk and PQ hatQkhatQktop QQtop}, we have
\begin{equation}
\label{eq: D Nk QQtop}
    \norm{\qty(D (N^{(k)} - Q^* Q^{*^\top}))_{\gI \gI} }_{\text{sp}} \leq  \frac{75\lambda_1^*}{4\lambda_r^*} \sqrt{\frac{sr}{s^*}I(U^*)} W^{(k-1)} + \frac{625 \lambda_1^{*2} sr}{16 \lambda_r^{*4} s^*} W^{(k-1)^2}.
\end{equation}
Moreover, by combining \eqref{eq: Rk T0 and Ntildek QQtop gIgI}, \eqref{eq: Nktilde and Nk gIgI} and \eqref{eq: D Nk QQtop}, we obtain that
\begin{equation}
\label{eq: bound for Rk gIgI}
     W^{(k)} \leq \frac{3}{2} T_0 +  \frac{75\sqrt{2}\lambda_1^*}{4\lambda_r^*} \sqrt{\frac{sr}{s^*}I(U^*)} W^{(k-1)} + \frac{625 \lambda_1^{*2} sr}{8 \lambda_r^{*4} s^*} W^{(k-1)^2}.
\end{equation}
Finally, since \eqref{eq: W^{(k)}} holds for \(k - 1\), we have 
\begin{align*}
    W^{(k)} &\leq \frac{3}{2} T_0 + \frac{75\sqrt{2}\lambda_1^*}{4\lambda_r^*} \sqrt{\frac{sr}{s^*}I(U^*)} \qty(6 T_0 + \frac{c_2's^*\lambda_1^{*2}}{2^{k - 1} s r}) 
    + \frac{625 \lambda_1^{*2} sr}{8 \lambda_r^{*4} s^*} \qty(6 T_0 + \frac{c_2's^*\lambda_1^{*2}}{2^{k - 1} s r})^2 \\
    &\leq \qty(\frac{3}{2} + \frac{225\sqrt{2}\lambda_1^*}{2\lambda_r^*} \sqrt{\frac{sr}{s^*}I(U^*)}
    + \frac{1875 \lambda_1^{*4} c_2'}{2^{k+1} \lambda_r^{*4}} + \frac{5625 \lambda_1^{*2} sr}{2 \lambda_r^{*4} s^*}T_0) T_0 \\
    &\quad + \qty(\frac{75\sqrt{2}\lambda_1^*}{2\lambda_r^*} \sqrt{\frac{sr}{s^*}I(U^*)}
    + \frac{625 \lambda_1^{*4} c_2'}{2^{k+1} \lambda_r^{*4}}) \frac{c_2' s^* \lambda_1^{*2}}{2^k s r}.
\end{align*}
Using the bounds \(I(U^*) \leq \frac{c_1}{r} \leq \frac{c_2' s}{s' r}\) and \(c_2'\) is sufficiently small, along with the assumption \(s \log d \ll n \), we obtain that, for \(k \geq 1\), 
\[
\begin{cases}
    \frac{3}{2} + \frac{225\sqrt{2}\lambda_1^*}{2\lambda_r^*} \sqrt{\frac{sr}{s^*}I(U^*)}
    + \frac{1875 \lambda_1^{*4} c_2'}{2^{k+1} \lambda_r^{*4}} + \frac{5625 \lambda_1^{*2} sr}{2 \lambda_r^{*4} s^*}T_0 \leq 6 \\
    \frac{75\sqrt{2}\lambda_1^*}{2\lambda_r^*} \sqrt{\frac{sr}{s^*}I(U^*)}
    + \frac{625 \lambda_1^{*4} c_2'}{2^{k+1} \lambda_r^{*4}} \leq 1.
\end{cases}
\]
Therefore, 
\[
W^{(k)} \leq 6 T_0 + \frac{c_2' s^* \lambda_1^{*2}}{2^k s r}.
\]

Furthermore, we verify that $W^{(k)}$ and $\hat Q^{(k)}$ satisfy the conditions required in Theorem \ref{prop: distance V and Vtk}. First, since $W^{(k)} \leq 6 T_0 + \frac{c_2' s^* \lambda_1^{*2}}{2^k s r}$, and by \(s \log d \ll n \) and \(c_2'\) is sufficiently small, it follows that \(W^{(k)} \leq \frac{c_3\lambda_r^{*3}}{\lambda_1^*\sqrt{r}}\). Second, from \eqref{eq: distance between hatQk and Q}, we have,
\begin{equation*}
    \norm{\hat{Q}^{(k)} \hat{Q}^{(k)^\top} - Q^* Q^{*^\top}}_{F} \leq  \frac{25\lambda_1^*}{4\lambda_r^*} \sqrt{\frac{sr}{s^*}} W^{(k-1)} \leq \frac{\lambda_r^{*3}}{20 \lambda_1^*}.
\end{equation*}
Thus, both conditions are satisfied, and the induction is complete.

Substituting \(T_0 = 2 c_1' (\lambda_1^{*2} + \sigma_{(1)}^{*2}) \qty(\sqrt{\frac{s \log d}{n}} + \frac{s^2 \log d}{n})\), we peroform \(K = \Omega\qty(\log n)\) iterations. By Assumptions \ref{asm: Sigma} and \ref{asm: Lambda}, and \eqref{eq: distance between hatQk and Q}, we have,
\begin{align*}
\norm{\hat{Q}^{(K)} \hat{Q}^{(K)^\top} - Q^* Q^{*^\top}}_F \lesssim \frac{\lambda_1^*}{\lambda_r^*}\sqrt{\frac{sr}{s^*}}(\lambda_1^{*2} + \sigma_{(1)}^{*2}) \qty(\sqrt{\frac{s \log d}{n}} + \frac{s^2 \log d}{n})
\lesssim \frac{s}{s^*}\sqrt{\frac{s^* r \log d}{n}}.
\end{align*}
This completes the proof of Theorem \ref{thm: edge embeddings sparse}. 
\end{proof}

\subsection{Proof of Corollary \ref{cor: edge embeddings}}

\begin{proof}
Our proof focuses on the estimator from the outer iteration. This is sufficient, because the inner loop's gradient descent converges to the global minimizer with a high probability. As established in Lemma \ref{lem: solution of loss}, this convergence is guaranteed because the loss function \eqref{loss: cl conti triplet transpose expectation} is convex with a high probability. We still consider the loss $\mL(V)$ in \eqref{loss: cl conti triplet transpose expectation simplified}, and perform Algorithm \ref{alg: edge embeddings conti sparse} without the hard thresholding step.

Similar as before, we define the event $\Omega_2$ as
\begin{equation*}
\Omega_2 = \qty { \norm{M - Q^* Q^{*^\top} - \Sigma}_{\text{sp}} \leq c_3' (\lambda_1^{*2} + \sigma_{(1)}^{*2})\sqrt{\frac{d + c_4'}{n}} },
\end{equation*}
where $c_4' = o\qty(n)$ is a positive constant. The following lemma ensures that $\Omega_2$ occurs with a high probability.

\begin{lem}
\label{lem: Omega2}
Suppose $d < n$. There exist constants $c_3', c_4' > 0$ and $c_4' = o\qty(n)$, such that $\P(\Omega_2) \geq 1 - ce^{-c_4'\sqrt{d + c_4'}(\frac{\sqrt{n}}{d} \wedge \sqrt{d + c_4'})}$.
\end{lem}

Under $\Omega_2$, the next lemma characterizes the solution for the inner iteration.

\begin{lem} \label{lem: solution of loss}
Suppose Assumptions \ref{asm: Lambda} and \ref{asm: Q incoherent} hold, and that $r \ll n$ and $d \ll n$. Let $\hat V$ be the minimizer of the loss \eqref{loss: cl conti triplet transpose expectation simplified}. Then, with probability at least $1 - ce^{-c_4'\sqrt{d + c_4'}(\frac{\sqrt{n}}{d} \wedge \sqrt{d + c_4'})}$, we have
\begin{align}
        \hat V \hat V^\top = \frac{1}{\lambda} \rank_r\qty(\Delta \qty(M) + P^2 D\qty(M)).
\end{align}
\end{lem}

Accordingly, when we set $\lambda = 1$ and choose the masking parameter as \eqref{eq: masking p at step k} during the $k$th outer iteration, denoting $\hat{Q}^{(k)}$ as the output of the $k$th outer iteration, we have
\begin{equation*}
    \qty(P^{(k)^2} D(M) )_{e,e} = 
    \begin{cases}
        M_{e,e} & \text{when } N^{(k-1)}_{e,e} \geq M_{e,e} \\
        N^{(k-1)}_{e,e} & \text{when } N^{(k-1)}_{e,e} < M_{e,e}
    \end{cases}
\end{equation*}
where $P^{(k)} = \text{diag} ( p^{(k)}_1, \cdots, p^{(k)}_d )$, and $N^{(k-1)} = \hat{Q}^{(k-1)} \hat{Q}^{(k-1)^\top}$. 

Moreover, denote $\tilde{N}^{(k-1)} = \Delta (M) + P^{(k)^2} D(M)$, and $\tilde{N}^{(0)} = N^{(0)} = \Delta (M)$. So we take $\lambda = 1$, the solution of loss \ref{loss: cl conti triplet transpose expectation simplified} at the $k$th outer iteration is $N^{(k)} = \rank_r(\tilde{N}^{(k-1)})$.

Next, we analyze the initial error $W^{(0)} = \norm{\tilde{N}^{(0)} - Q^* Q^{*^\top}}_{\text{sp}}$, and the evolution of iterations' error $W^{(k)} = \norm{\tilde{N}^{(k)} - Q^* Q^{*^\top}}_{\text{sp}}$.

For the initial error $W^{(0)}$, we have
\begin{equation*}
    W^{(0)} = \norm{\Delta (M) - \Delta ( Q^* Q^{*^\top} )}_{\text{sp}} + \norm{D(Q^* Q^{*^\top})}_{\text{sp}} \leq T_0 + \lambda_1^{*2} I(U^*),
\end{equation*}
where $T_0 = 2c_1 (\lambda_1^{*2} + \sigma_{(1)}^{*2})\sqrt{\frac{d + c_4'}{n}}$ by \eqref{eq: offM bound}.

Next, we establish an upper bound for $W^{(k)}$ based on $W^{(k-1)}$. Since
\begin{multline}
\label{eq: Rk T0 and Ntildek QQtop}
    W^{(k)} = \norm{\qty(\tilde{N}^{(k)} - Q^* Q^{*^\top})_{\gI \gI}}_{\text{sp}} \leq \norm{\qty(\Delta (M - Q^* Q^{*^\top} - \Sigma))_{\gI \gI}}_{\text{sp}} + \\ \norm{\qty(D (\tilde{N}^{(k)} - Q^* Q^{*^\top}))_{\gI \gI} }_{\text{sp}}
    = T_0 + \norm{\qty(D (\tilde{N}^{(k)} - Q^* Q^{*^\top}))_{\gI \gI} }_{\text{sp}},    
\end{multline}
we only need to analyze the diagonal term.

First, we bound the difference between 
\[
\norm{D (\tilde{N}^{(k)}) - D (Q^* Q^{*^\top})}_{\text{sp}} \;\; \text{ and } \;\; \norm{D(N^{(k)}) - D(Q^* Q^{*^\top})}_{\text{sp}}.
\]
Since $D (\tilde{N}^{(k)})$, $D (Q^* Q^{*^\top})$ and $D(N^{(k)})$ are all diagonal and positive semi-definite, we can analyze the difference by computing each entry.

If $N^{(k)}_{e,e} \geq (Q^* Q^{*^\top})_{e,e}$, we have
\begin{equation*}
    \left| \min \{ M_{e,e}, N^{(k)}_{e,e} \} - (Q^* Q^{*^\top})_{e,e} \right| \leq \left| N^{(k)}_{e,e} - (Q^* Q^{*^\top})_{e,e} \right|, \text{i.e.}
\end{equation*}
\begin{equation*}
    \left| \tilde{N}^{(k)}_{e,e} - (Q^* Q^{*^\top})_{e,e} \right| \leq \left| N^{(k)}_{e,e} - (Q^* Q^{*^\top})_{e,e} \right|.
\end{equation*}
If $D(N^{(k)})_{e,e} < D(Q^* Q^{*^\top})_{e,e}$, only when $D(M)_{e,e} < D(N^{(k)})_{e,e} < D(Q^* Q^{*^\top})_{e,e}$, we have
\begin{equation*}
    D(Q^* Q^{*^\top})_{e,e} - D(M)_{e,e} > D(Q^* Q^{*^\top})_{e,e} - D(N^{(k)})_{e,e}.
\end{equation*}
In this situation,
\begin{align*}
    D(Q^* Q^{*^\top})_{e,e} - D(M)_{e,e} &\leq D(Q^* Q^{*^\top} + \Sigma)_{e,e} - D(M)_{e,e} \leq \norm{D(Q^* Q^{*^\top} + \Sigma) - D(M)}_{\text{sp}}\\
    &\leq \norm{Q^* Q^{*^\top} + \Sigma - M}_{\text{sp}} \leq \frac{T_0}{2}.
\end{align*}
Combining the two situations, we have
\begin{equation}
\label{eq: Nktilde and Nk}
    \norm{D(\tilde{N}^{(k)}) - D(Q^* Q^{*^\top})}_{\text{sp}} \leq \norm{D(N^{(k)}) - D(Q^* Q^{*^\top})}_{\text{sp}} + \frac{T_0}{2}.
\end{equation}

Second, we bound $\norm{D(N^{(k)}) - D(Q^* Q^{*^\top})}_{\text{sp}}$. Denote $\tilde{U}^{(k)}$ as the first $r$th eigenvector of $\tilde{N}^{(k)}$.Then by definition,
\begin{equation*}
    D(N^{(k)} - Q^* Q^{*^\top}) = D \qty( P_{\tilde{U}^{(k-1)}} \tilde{N}^{(k-1)} - Q^* Q^{*^\top} ).
\end{equation*}
We decompose
\begin{equation*}
    Q^* Q^{*^\top} = P_{\tilde{U}^{(k-1)}} Q^* Q^{*^\top} + P_{\tilde{U}^{(k-1)}_\perp} Q^* Q^{*^\top}.
\end{equation*}
Then
\begin{equation*}
    D(N^{(k)} - Q^* Q^{*^\top}) = D \qty( P_{\tilde{U}^{(k-1)}} \qty( \tilde{N}^{(k-1)} - Q^* Q^{*^\top} ) ) - D \qty( P_{\tilde{U}^{(k-1)}_\perp} Q^* Q^{*^\top} ).
\end{equation*}

Next, we analyze $D \qty( P_{\tilde{U}^{(k-1)}} (\tilde{N}^{(k-1)} - Q^* Q^{*^\top}))$. Let $P_{\tilde{U}^{(k-1)}} = P_{U^*} + (P_{\tilde{U}^{(k-1)}} - P_{U^*})$. Then
\begin{align*}
    D(N^{(k)} - Q^* Q^{*^\top}) &= D \qty( P_{U^*} (\tilde{N}^{(k-1)} - Q^* Q^{*^\top}) ) - D \qty( P_{\tilde{U}^{(k-1)}_\perp} Q^* Q^{*^\top} ) \\
    &\quad + D \qty( (P_{\tilde{U}^{(k-1)}} - P_{U^*}) (\tilde{N}^{(k-1)} - Q^* Q^{*^\top}) ).
\end{align*}
By Lemma \ref{lem: zhang projection contraction},
\begin{equation}
\label{eq: Nk-1tilde and QQtop}
    \norm{D(P_{U^*} (\tilde{N}^{(k-1)} - Q^* Q^{*^\top}))}_{\text{sp}} \leq \sqrt{I(U^*)} \norm{\tilde{N}^{(k-1)} - Q^* Q^{*^\top}}_{\text{sp}} = \sqrt{I(U^*)} W^{(k-1)}.
\end{equation}
By Lemmas \ref{lem: zhang projection contraction} and \ref{lem: zhang noise porjection bound},
\begin{equation}
    \begin{aligned}
    \label{eq: PQtildek-1QQtop}
        \norm{D(P_{\tilde{U}^{(k-1)}_\perp} Q^* Q^{*^\top})}_{\text{sp}} &= \norm{D(P_{\tilde{U}^{(k-1)}_\perp} Q^* Q^{*^\top} P_Q)}_{\text{sp}} \leq \sqrt{I(U^*)} \norm{P_{\tilde{U}^{(k-1)}_\perp} Q^* Q^{*^\top}}_{\text{sp}}\\
        &\leq 2 \sqrt{I(U^*)} \norm{\tilde{N}^{(k-1)} - Q^* Q^{*^\top}}_{\text{sp}} = 2 \sqrt{I(U^*)} W^{(k-1)}.
    \end{aligned}
\end{equation}
Note that
\begin{align*}
    \norm{\tilde{U}^{(k-1)} \tilde{U}^{(k-1)^\top} -  U^{*^\top}}_{\text{sp}}
    & \leq 2 \norm{\sin \Theta (\tilde{U}^{(k-1)}, U^*)}_{\text{sp}} = 2 \norm{\tilde{U}^{(k-1)^\top}_\perp U^*}_{\text{sp}}\\
    & \leq 2 \norm{\tilde{U}^{(k-1)^\top}_\perp  U^* U^{*^\top} Q^* Q^{*^\top}} \cdot \sigma_r^{-1}(U^{*^\top} Q^* Q^{*^\top})\\
    & \leq 2 \norm{\tilde{U}^{(k-1)^\top}_\perp Q^* Q^{*^\top}} \cdot \sigma_r^{-1}(Q^* Q^{*^\top})\\
    & \leq \frac{4}{\lambda_r^{*2}} \norm{\tilde{N}^{(k-1)} - Q^* Q^{*^\top}}_{\text{sp}}
\end{align*}
Then
\begin{equation}
\label{eq: P and Nk-1tilde and QQtop}
    \| D (P_{\tilde{U}^{(k-1)}} - P_{U^*}) (\tilde{N}^{(k-1)} - Q^* Q^{*^\top}) \|_{\text{sp}}
    \leq \| P_{\tilde{U}^{(k-1)}} - P_{U^*} \|_{\text{sp}} W^{(k-1)} \leq \frac{4}{\lambda_r^{*2}} W^{(k-1)^2}.
\end{equation}
Combining \eqref{eq: Nk-1tilde and QQtop}, \eqref{eq: PQtildek-1QQtop} and \eqref{eq: P and Nk-1tilde and QQtop}, we have
\begin{equation}
\label{eq: bound for Nk and QQtop}
    \norm{D(N^{(k)}) - D(Q^* Q^{*^\top})}_{\text{sp}} \leq 3 \sqrt{I(U^*)} W^{(k-1)} + \frac{4}{\lambda_r^{*2}} W^{(k-1)^2}
\end{equation}
Moreover, combining \eqref{eq: Rk T0 and Ntildek QQtop}, \eqref{eq: Nktilde and Nk} and \eqref{eq: bound for Nk and QQtop}, we obtain that
\begin{equation}
\label{eq: bound for Rk}
    W^{(k)} \leq \frac{3}{2} T_0 + 3 \sqrt{I(U^*)} W^{(k-1)} + \frac{4}{\lambda_r^{*2}} W^{(k-1)^2}.
\end{equation}

Finally, we use induction to show $W^{(k)} \leq 3 T_0 + 2^{-k - 7}\lambda_r^{*2}$. 

We start with the base case of $k=0$. By Assumption \ref{asm: Q incoherent}, $W^{(0)} \leq T_0 + \lambda_1^{*2} I(U^*) \leq 3 T_0 + 2^{-7}\lambda_r^{*2}$.

We then continue with the induction step. Suppose $W^{(k-1)} \leq 3 T_0 + 2^{-k - 6}\lambda_r^{*2}$. We have
\begin{align*}
    W^{(k)} &\leq \frac{3}{2} T_0 + 3 \sqrt{I(U^*)} \qty( 3 T_0 + 2^{-k - 6}\lambda_r^{*2} ) + \frac{4}{\lambda_r^{*2}} \qty( 3 T_0 + 2^{-k - 6}\lambda_r^{*2} )^2 \\
    &= \frac{3}{2} T_0 + 9 \sqrt{I(U^*)} T_0 + 6 \sqrt{I(U^*)} 2^{-k - 7}\lambda_r^{*2} + 3 \cdot 2^{-k - 4} T_0 + \frac{36}{\lambda_r^{*2}} T_0^2 + 2^{-k - 3}2^{-k - 7}\lambda_r^{*2} \\
    &= \qty( \frac{3}{2} + 9 \sqrt{I(U^*)} + 3 \cdot 2^{-k - 4} + \frac{36}{\lambda_r^{*2}} T_0 ) T_0 + \qty( 6 \sqrt{I(U^*)} + 2^{-k - 3}) 2^{-k - 7}\lambda_r^{*2}
\end{align*}
By Assumption \ref{asm: Q incoherent}, $r \ll n$ and $d \ll n$, we have $\sqrt{I(U^*)} \leq \frac{1}{12}$ and $T_0 \leq \frac{\lambda_r^{*2}}{72}$.
Therefore, for any $k \geq 1$,
\[
\begin{cases}
    \frac{3}{2} + 9 \sqrt{I(U^*)} + 3 \cdot 2^{-k - 4} + \frac{36}{\lambda_r^{*2}} T_0 \leq 3 \\
    6 \sqrt{I(U^*)} + 2^{-k - 3} \leq 1.
\end{cases}
\]
We have $W^{(k)} \leq 3 T_0 + 2^{-k - 7}\lambda_r^{*2}$. Furthermore, by applying \eqref{eq: bound for Nk and QQtop} again, we obtain that
\[
\norm{N^{(k+1)} - Q^* Q^{*^\top}}_{\text{sp}} \leq T_0 + 3 \sqrt{I(U^*)} W^{(k)} + \frac{4}{\lambda_r^{*2}} R^{(k)^2} \leq 3 T_0 + 2^{-k - 8}\lambda_r^{*2},
\] which implies that
\[
\norm{\hat Q^{(k+1)} \hat Q^{(k+1)^\top} - Q^* Q^{*^\top}}_{\text{sp}} \leq 3 T_0 + 2^{-k - 8}\lambda_r^{*2}.
\]
Since $T_0 = 2c_3' (\lambda_1^{*2} + \sigma_{(1)}^{*2})\sqrt{\frac{d + c_4'}{n}}$, we perform $K = O\qty(\log n)$ iterations. Then
\begin{equation*}
    \norm{\hat Q^{(K)} \hat Q^{(K)^\top} - Q^* Q^{*^\top}}_{\text{sp}} \lesssim (\lambda_1^{*2} + \sigma_{(1)}^{*2})\sqrt{\frac{d + c_4'}{n}}.
\end{equation*}
By taking $c_4' = d$, with probability at least $1 - ce^{-c\sqrt{\frac{n}{d}} \wedge d}$, we obtain that, 
\begin{equation*}
    \norm{\hat Q^{(K)} \hat Q^{(K)^\top}  - Q^* Q^{*^\top}}_{\text{sp}} \lesssim \sqrt{\frac{d}{n}}.
\end{equation*}
This completes the proof of Corollary \ref{cor: edge embeddings}. 
\end{proof}

\subsection{Proofs of supporting lemmas}
\label{sec:app proof of lemma in app}

\subsubsection{Proof of Lemma \ref{lem: simplified loss of cl conti triplet transpose expectation}}
\begin{proof}
Note that we can rewrite the expected loss $\mL(V)$ in \eqref{loss: cl conti triplet transpose expectation}  as
\begin{align*}
    \mL(V) &= -\frac{1}{n} \E_A \qty[ \langle V^\top A X, V^\top (I-A) X \rangle ] 
    + \frac{1}{n^2} \E_A \qty[ \langle V^\top A X 1_n, V^\top (I-A) X 1_n \rangle ] + \Pi(V) \\
    & = \E_A \qty[ -\frac{1}{n} \operatorname{tr} \qty( X^\top A^\top V V^\top (I-A) X )
    + \frac{1}{n^2} \operatorname{tr} \qty( 1_n^\top X^\top A^\top V V^\top (1-A) X 1_n ) ] + \Pi(V) \\
    & = \frac{\lambda}{8} \| V V^\top \|_F^2 - \E_A \qty[ \frac{1}{n} \operatorname{tr} \qty( G^\top G (I-A) X X^\top A^\top )
    - \frac{1}{n^2} \operatorname{tr} \qty( V V^\top (I-A) X 1_n 1_n^\top X^\top A^\top ) ] \\
    & = \frac{\lambda}{8} \E_A \qty[ \left \| V V^\top - \frac{4}{\lambda} (I-A) \qty( \frac{1}{n} X X^\top - \frac{1}{n^2} X 1_n 1_n^\top X^\top ) A \right \|_F^2 + C(X, A)].
\end{align*}
Consider the matrix $(I - A) M A$. For the off-diagonal entry $(e, e')$, $e \neq e'$, we have
\begin{align*}
    \E \qty[ (1 - a_e) (a_{e'}) M_{e, e'} ] = \E \qty[(1 - a_e)] \E \qty[a_{e'}] M_{e, e'} = \frac{1}{4} M_{e, e'}.
\end{align*}
For the diagonal entry $(e, e)$, we have
\begin{align*}
    \E \qty[ (1 - a_e) a_e M_{e, e} ] = \E \qty[ (1 - a_e) a_e ] M_{e, e} = \frac{p_e^2}{4} M_{e, e}.
\end{align*}
Then, we can further rewrite the loss as
\begin{align*}
    \mL(V) = \frac{\lambda}{8} \norm{V V^\top - \frac{1}{\lambda} \qty(\Delta \qty(M) + P^2 D\qty(M))}_F^2 + C(X).
\end{align*}
This completes the proof of Lemma \ref{lem: simplified loss of cl conti triplet transpose expectation}. 
\end{proof}

\subsubsection{Proof of Lemma \ref{lem: gradient descent}}

\begin{proof}
Due to the sparsity of \(\hat{V}_{(t)}, \tilde{V}_{(t)}\), at step \(t\), we can consider the restricted loss function, 
\[
f_t(L) = \frac{\lambda}{2} \norm{L L^\top - \frac{1}{\lambda} N_{(t)}}_F^2,
\]
where we denote \(N_{S_{(t)} S_{(t)}}\) by \(N_{(t)}\) for simplicity in the proof.

We begin with the gradient expression,
\begin{equation}
\label{eq: f_tL gradient}
    \frac{1}{2} \nabla f_t(L) = -N_{(t)} L + \lambda L L^\top L.
\end{equation}
Let \(L_{(t)} = \tilde{V}_{t, S_{(t)}*}\). During the gradient descent, the update rule becomes \(L_{(t+1)} = L_{(t)} - \eta \nabla f_t(L_{(t)})\). We have that \(L_{(t+1)} = V_{(t+1), S_{(t)}*}\), which holds because \(\supp(\tilde{V}_{(t)}) \subseteq S_{(t)}\). So we work on the distance involving \(L_{(t)}\) and \(L_{(t+1)}\).

We leverage Lemma \ref{lem: gao gradient descent}. First, we define the minimizer of \(f_t(L)\) as \(L_{(t)}^*\).
Under the condition \(\|N_{(t)} - (Q^* Q^{*^\top})_{S_{(t)} S_{(t)}}\|_{\text{sp}} \leq W^{(k-1)} < \frac{\lambda_r^{*2}}{4} < \frac{\lambda_r^{*2}}{2}\), applying the Weyl's inequality in Lemma \ref{lem: Weyl}, we have the following eigenvalue bounds for \(\tilde N_{(t)}\)
\[
\lambda_i(N_{(t)}) > \frac{\lambda_r^{*2}}{2}, \quad i \leq r, \quad 
\lambda_i(N_{(t)}) > -\frac{\lambda_r^{*2}}{2}, \quad i > r,
\]
where $\lambda_j(A)$ is the $j$th largest eigenvalue value of $A$. This implies that the largest $r$ eigenvalues of $N_{(t)}$ are also the largest $r$ singular values. By Lemma \ref{lem: eckart},
\[
L_{(t)}^* L_{(t)}^{*\top} = \hat{Q}_{S_{(t)}*}(S_{(t)}) \frac{\hat{\Lambda}_r(S_{(t)})}{\lambda} \hat{Q}_{S_{(t)}*}(S_{(t)})^\top.
\]
Then we can take \(L_{(t)}^* = \hat{V}_{S_{(t)}*}(S_{(t)})\), up to an orthogonal transformation.

Next, we check the conditions in Lemma \ref{lem: gao gradient descent} one by one. It is straightforward to verify the condition \(\nabla f_t(L) P = \nabla f_t(LP)\) by \eqref{eq: f_tL gradient}. Next we derive the expression for \(\operatorname{vec}(Z)^\top \nabla^2 f_t(L) \operatorname{vec}(Z)\). The gradient expression in the vectorized form is
\[
\frac{1}{2} \operatorname{vec} \nabla f_t(L) = -\qty(I_r \otimes N_{(t)}) \operatorname{vec}(L) 
+ \lambda \qty(I_r \otimes L L^\top) \operatorname{vec}(L).
\]
For the first term,
\[
\frac{\partial -\qty(I_r \otimes N_{(t)}) \operatorname{vec}(L)}{\partial \operatorname{vec}(L)} 
= - I_r \otimes N_{(t)}.
\]
Henceforth, 
\[
\begin{pmatrix}
z_1^\top & \cdots & z_r^\top
\end{pmatrix}
\begin{pmatrix}
-N_{(t)} & 0 & \cdots & 0 \\
0 & -N_{(t)} & \cdots & 0 \\
\vdots & \vdots & \ddots & \vdots \\
0 & 0 & \cdots & -N_{(t)}
\end{pmatrix}
\begin{pmatrix}
z_1 \\
\vdots \\
z_r
\end{pmatrix}
= \langle Z Z^\top, -N_{(t)} \rangle.
\]
For the second term, recalling that \(L \in \mathbb{R}^{d \times r}\), its vectorized form can be written as,
\[
\operatorname{vec}(L) = \qty(l_1^\top, l_2^\top, \dots, l_r^\top)^\top,
\]
where \(l_i \in \mathbb{R}^d\) represents the \(i\)th column of \(L\). Note that \(L L^\top = \sum_{i=1}^r l_i l_i^\top\), and
\[
\qty(I_r \otimes L L^\top) \operatorname{vec}(L) = 
\begin{pmatrix}
\sum_{i=1}^r l_i l_i^\top l_1 \\
\sum_{i=1}^r l_i l_i^\top l_2 \\
\vdots \\
\sum_{i=1}^r l_i l_i^\top l_r
\end{pmatrix}.
\]
We then compute the gradient of $\sum_{i=1}^r l_i l_i^\top l_j$ with respect to $l_j$. When \(j \neq k\),
\[
\frac{\partial \sum_{i=1}^r l_i l_i^\top l_j}{\partial l_k} = \frac{\partial l_k l_k^\top l_j}{\partial l_k} = l_j^\top l_k I_d + l_j l_k^\top.
\]
When \(j = k\),
\[
\frac{\partial \sum_{i=1}^r l_i l_i^\top l_j}{\partial l_k} = \frac{\partial \sum_{i=1}^r l_i l_i^\top l_j}{\partial l_j} = \frac{\partial l_j l_j^\top l_j}{\partial l_j} + \sum_{i \neq j} \frac{\partial l_i l_i^\top l_j}{\partial l_j} 
= L L^\top + l_j^\top l_jI_d + l_j l_j^\top.
\]
Henceforth, 
\[
\begin{pmatrix}
z_1^\top & \cdots & z_r^\top
\end{pmatrix}
\begin{pmatrix}
L L^\top & 0 & \cdots & 0 \\
0 & L L^\top & \cdots & 0 \\
\vdots & \vdots & \ddots & \vdots \\
0 & 0 & \cdots & L L^\top
\end{pmatrix}
\begin{pmatrix}
z_1 \\
\vdots \\
z_r
\end{pmatrix}
= \langle Z Z^\top, L L^\top \rangle,
\]
\[
\begin{pmatrix}
z_1^\top & \cdots & z_r^\top
\end{pmatrix}
\begin{pmatrix}
l_1^\top l_1 I_d & l_1^\top l_2 I_d & \cdots & l_1^\top l_r I_d \\
l_2^\top l_1 I_d & l_2^\top l_2 I_d & \cdots & l_2^\top l_r I_d \\
\vdots & \vdots & \ddots & \vdots \\
l_r^\top l_1 I_d & l_r^\top l_2 I_d & \cdots & l_r^\top l_r I_d
\end{pmatrix}
\begin{pmatrix}
z_1 \\
\vdots \\
z_r
\end{pmatrix}
= \langle L^\top L, Z^\top Z \rangle,
\]

\[
\begin{pmatrix}
z_1^\top & \cdots & z_r^\top
\end{pmatrix}
\begin{pmatrix}
l_1 l_1^\top & l_1 l_2^\top & \cdots & l_1 l_r^\top \\
l_2 l_1^\top & l_2 l_2^\top & \cdots & l_2 l_r^\top \\
\vdots & \vdots & \ddots & \vdots \\
l_r l_1^\top & l_r l_2^\top & \cdots & l_r l_r^\top
\end{pmatrix}
\begin{pmatrix}
z_1 \\
\vdots \\
z_r
\end{pmatrix}
= \langle Z^\top L, L^\top Z \rangle.
\]

Combining the above results together, we define
\begin{multline}
\label{eq: hessian for L and Z}
g_t(L, Z) = \frac{1}{2} \operatorname{vec}(Z)^\top \nabla^2 f_t(L) \operatorname{vec}(Z) = \\
-\langle N_{(t)}, ZZ^\top \rangle 
+ \lambda \langle ZZ^\top, LL^\top \rangle
+ \lambda \langle Z^\top Z, L^\top L \rangle 
+ \lambda \langle Z^\top L, L^\top Z \rangle.
\end{multline}

Next, we verify the smoothness condition and the strong convexity condition under a proper radius condition. 

For the smoothness condition, suppose \(\|L - L_{(t)}^*\|_F \leq \delta\). By the condition \(\|N_{(t)} - (Q^* Q^{*^\top})_{S_{(t)} S_{(t)}}\|_{\text{sp}} \leq W^{(k-1)}\), we have
\begin{equation*}
\begin{aligned}
\| L_{(t)}^* \|_{\text{sp}} = \| \hat{V}_{S_{(t)}*}(S_{(t)}) \|_{\text{sp}} = \| \hat{Q}_{S_{(t)}*}(S_{(t)}) \qty(\frac{\hat{\Lambda}_r(S_{(t)})}{\lambda})^{\frac{1}{2}} \|_{\text{sp}} \leq& \sqrt{\frac{\lambda_1^{*2} + W^{(k-1)}}{\lambda}},\\
\lambda_{\max}(-N_{(t)}) \leq& W^{(k-1)},
\end{aligned}
\end{equation*}
where $\lambda_{\max}(A)$ is the maximum eigenvalue of $A$. We analyze each term. 
\[
\langle -N_{(t)}, ZZ^\top \rangle \leq W^{(k-1)} \|Z\|_F^2,
\]
\[
\langle LL^\top, ZZ^\top \rangle \leq \|LL^\top\|_{\text{sp}} \|Z\|_F^2 
\leq \qty( \delta + \sqrt{\frac{\lambda_1^{*2} + W^{(k-1)}}{\lambda}} )^2 \|Z\|_F^2,
\]
\[
\langle L^\top L, ZZ^\top \rangle \leq \|L^\top L\|_{\text{sp}} \|Z\|_F^2 
\leq \qty( \delta + \sqrt{\frac{\lambda_1^{*2} + W^{(k-1)}}{\lambda}} )^2 \|Z\|_F^2,
\]
\[
\langle Z^\top L, L^\top Z \rangle \leq \|Z^\top L\|_F \|L^\top Z\|_F 
\leq \|L\|_{\text{sp}} \|L^\top\|_{\text{sp}} \|Z\|_F^2 
\leq \qty(\delta + \sqrt{\frac{\lambda_1^{*2} + W^{(k-1)}}{\lambda}} )^2 \|Z\|_F^2.
\]
Combining the above results, we have,
\[
g_t(L, Z) \leq \qty(W^{(k-1)} + 3 \lambda \qty( \delta + \sqrt{\frac{\lambda_1^{*2} + W^{(k-1)}}{\lambda}} )^2 ) \|Z\|_F^2.
\]
We then choose 
\[
\beta = 2 W^{(k-1)} + 6 \lambda \qty( \delta + \sqrt{\frac{\lambda_1^{*2} + W^{(k-1)}}{\lambda}} )^2.
\]

For the strong convexity condition, denoting \(\tilde{Z} = Z H_Z - L_{(t)}^*\), we have
\begin{align*}
    g_t(L_{(t)}^*, \tilde{Z}) &= -\langle N_{(t)}, \tilde{Z} \tilde{Z}^\top \rangle 
    + \lambda \langle L_{(t)}^* L_{(t)}^{*\top}, \tilde{Z} \tilde{Z}^\top \rangle 
    + \lambda \langle L_{(t)}^{*\top} L_{(t)}^*, \tilde{Z}^\top \tilde{Z} \rangle 
    + \lambda \langle \tilde{Z}^\top L_{(t)}^*, L_{(t)}^{*\top} \tilde{Z} \rangle \\
    &= \underbrace{\langle \tilde{Z} \hat{\Lambda}_r(S_{(t)}) \tilde{Z}^\top, I \rangle + \langle \lambda L_{(t)}^*L_{(t)}^{* \top} - N_{(t)}, \tilde{Z} \tilde{Z}^\top \rangle}_{D_1}
    + \underbrace{\lambda \langle \tilde{Z}^\top L_{(t)}^*, L_{(t)}^{*\top} \tilde{Z} \rangle}_{D_2}
\end{align*}
For $D_1$, \(L_{(t)}^{*} L_{(t)}^{*\top} = \hat{Q}_{S_{(t)}*}(S_{(t)}) \frac{\hat{\Lambda}_r(S_{(t)})}{\lambda} \hat{Q}_{S_{(t)}*}(S_{(t)})^\top\).
Denote 
\[
N_{(t)} = \hat{Q}_{S_{(t)}*}(S_{(t)}) \hat{\Lambda}_r(S_{(t)}) \hat{Q}_{S_{(t)}*}(S_{(t)})^\top + \hat{Q}_{S_{(t)}*}(S_{(t)})_\perp \hat{\tilde \Lambda}_r(S_{(t)}) \hat{Q}_{S_{(t)}*}(S_{(t)})_\perp^\top,
\]
where we define the remaining eigenvectors by \(\hat{Q}_{S_{(t)}*}(S_{(t)})_\perp\) and the rest of eigenvalues by the diagonal entries of \(\hat{\tilde \Lambda}_r(S_{(t)})\). Then, 
\begin{align*}
    D_1 &\geq (\lambda_r^{*2} - W^{(k-1)}) \langle \tilde{Z} \tilde{Z}^\top, I \rangle 
    + \langle -\hat{Q}_{S_{(t)}^*}(S_{(t)})_\perp \hat{\tilde \Lambda}_r(S_{(t)}) \hat{Q}_{S_{(t)}^*}(S_{(t)})_\perp^\top, \tilde{Z} \tilde{Z}^\top \rangle \\
    &= \langle \tilde{Z} \tilde{Z}^\top, \hat{Q}_{S_{(t)}^*}(S_{(t)}) \qty((\lambda_r^* - W^{(k-1)}) I - \hat{\tilde \Lambda}_r(S_{(t)})) \hat{Q}_{S_{(t)}^*}(S_{(t)})^\top \rangle \\
    &\geq (\lambda_r^{*2} - 2R^{(k-1)}) \|\tilde{Z}\|_F^2
\end{align*}
\begin{align*}
    D_2 &= \langle (Z H_Z - L_{(t)}^*)^\top L_{(t)}^{*}, L_{(t)}^{*\top} (Z H_Z - L_{(t)}^*) \rangle \nonumber \\
    &= \langle (Z H_Z)^\top L_{(t)}^{*}, L_{(t)}^{*\top} (Z H_Z) \rangle 
    - 2 \langle L_{(t)}^{*\top} L_{(t)}^*, L_{(t)}^{*\top} Z H_Z \rangle 
    + \langle L_{(t)}^{*\top} L_{(t)}^*, L_{(t)}^{*\top} L_{(t)}^* \rangle
\end{align*}
By Lemma \ref{lem: ten symmetric of XH^TL}, there exists $Z_1$, such that $Z_1 Z_1 = L_{(t)}^{*\top} Z H_Z$. Then, 
\[
D_2 = \langle Z_1^\top Z_1, Z_1^\top Z_1 \rangle + \langle L_{(t)}^{*\top} L_{(t)}^*, L_{(t)}^{*\top} L_{(t)}^* \rangle - 2 \langle L_{(t)}^{*\top} L_{(t)}^*, Z_1 Z_1 \rangle = \|Z_1^\top Z_1 - L_{(t)}^{*\top} L_{(t)}^*\|_F^2 \geq 0
\]
Now for a general \(L\), we have, 
\begin{multline*}
    \left| g_t(L, \tilde{Z}) - g_t(L_{(t)}^*, \tilde{Z}) \right| \leq \lambda \left| \langle \tilde{Z} \tilde{Z}^\top, (L L^\top - L_{(t)}^* L_{(t)}^{*\top}) \rangle \right|  + \lambda \left| \langle \tilde{Z}^\top \tilde{Z}, (L^\top L - L_{(t)}^{*\top} L_{(t)}^*) \rangle \right| \\
     + \lambda \left| \langle \tilde{Z}^\top L, L^\top \tilde{Z} \rangle - \langle \tilde{Z}^\top L_{(t)}^*, L_{(t)}^{*\top} \tilde{Z} \rangle \right|.
\end{multline*}
For each of the three terms on the right-hand-side of the above equation, we have
\begin{align*}
& \left| \langle \tilde{Z} \tilde{Z}^\top, L L^\top - L_{(t)}^* L_{(t)}^{*\top} \rangle \right| \leq \|\tilde{Z}\|_F^2 \|L L^\top - L_{(t)}^* L_{(t)}^{*\top}\|_{\text{sp}} 
    \leq \|\tilde{Z}\|_F^2 (\|L\|_{\text{sp}} + \|L_{(t)}^{*\top}\|_{\text{sp}}) \|L - L_{(t)}^*\|_{\text{sp}}, \\
& \left| \langle \tilde{Z}^\top \tilde{Z}, L^\top L - L_{(t)}^{*\top} L_{(t)}^* \rangle \right| \leq \|\tilde{Z}\|_F^2 \|L^\top L - L_{(t)}^{*\top} L_{(t)}^*\|_{\text{sp}} 
    \leq \|\tilde{Z}\|_F^2 (\|L^\top \|_{\text{sp}} + \|L_{(t)}^*\|_{\text{sp}}) \|L - L_{(t)}^*\|_{\text{sp}}, \\
& \left| \langle \tilde{Z}^\top L, L^\top \tilde{Z} \rangle - \langle \tilde{Z}^\top L_{(t)}^*, L_{(t)}^{*\top} \tilde{Z} \rangle \right| = \operatorname{tr}(\tilde{Z}^\top \tilde{Z} L - \tilde{Z}^\top \tilde{Z} L_{(t)}^*) \\
& \leq \|\tilde{Z}\|_F^2 \|L + L_{(t)}^*\|_{\text{sp}} \|L - L_{(t)}^*\|_{\text{sp}} 
\leq \|\tilde{Z}\|_F^2 (\|L \|_{\text{sp}} + \|L_{(t)}^*\|_{\text{sp}}) \|L - L_{(t)}^*\|_{\text{sp}}.
\end{align*}
Therefore, 
\begin{equation*}
    \left| g_t(L, \tilde{Z}) - g_t(L_{(t)}^*, \tilde{Z}) \right| \leq 3 \lambda \left( \delta + 2 \sqrt{\frac{\lambda_1^{*2} + W^{(k-1)}}{\lambda}} \right) \|L - L_{(t)}^*\|_F \|\tilde{Z}\|_F^2.
\end{equation*}
Then, under the radius condition \(\|L - L_{(t)}^*\|_F \leq \delta\), we obtain that
\begin{equation*}
     g_t(L, \tilde{Z}) \geq \left( \lambda_r^{*2} - 2 W^{(k-1)} - 3 \lambda \left( \delta + 2 \sqrt{\frac{\lambda_1^{*2} + W^{(k-1)}}{\lambda}} \right) \delta \right) \|\tilde{Z}\|_F^2.
\end{equation*}
Letting \(\delta < \frac{\lambda_r^{*2}}{12 \sqrt{\lambda (\lambda_1^{*2} +  W^{(k-1)})}}\), we have that,
\begin{align*}
    \lambda \delta^2 < \frac{\lambda_r^{*4}}{144(\lambda_1^{*2} +  W^{(k-1)})} < \frac{\lambda_r^{*2}}{144}, \quad \lambda \delta \sqrt{\frac{\lambda_1^{*2} + W^{(k-1)}}{\lambda}} < \frac{\lambda_r^{*2}}{12}.
\end{align*}
It implies that,
\begin{align*}
\alpha &= 2\lambda_r^{*2} - 4W^{(k-1)} - 6\lambda \delta^2 - 6 \lambda \delta \sqrt{\frac{\lambda_1^{*2} + W^{(k-1)}}{\lambda}} 
\geq \frac{17}{12}\lambda_r^{*2} - 4W^{(k-1)},\\
\beta &= 2 W^{(k-1)} + 6 \lambda \delta^2 + 6(\lambda_1^{*2} +  W^{(k-1)}) + 12 \lambda \delta \sqrt{\frac{\lambda_1^{*2} + W^{(k-1)}}{\lambda}} \\
 &\leq 8 W^{(k-1)} + \qty(\frac{1}{12} + 6 + 1)\lambda_1^{*2} 
 = 8 W^{(k-1)} + \frac{85}{12}\lambda_1^{*2}.
\end{align*}
When \(\eta \leq \frac{1}{8 W^{(k-1)} + \frac{85}{12}\lambda_1^{*2}}\), expanding the matrix of size \(|S_{(t)}| \times r\) to \(d \times r\) by filling in zeros, we obtain that
\[
\dist^2(\hat{V}_{(t+1)}^o, \hat{V}(S_{(t)})) \leq \qty( 1 - \eta \qty(\frac{17}{12}\lambda_r^{*2} - 4W^{(k-1)})) \dist^2(\hat{V}_{(t)}, \hat{V}(S_{(t)})).
\]

This completes the proof of \ref{lem: gradient descent}.
\end{proof}

\subsubsection{Proof of Lemma \ref{lem: Omega1}}
\label{sec: proof of lemma omega1}

\begin{proof}
We control the error on the subset $\gI$ of $M - Q^* Q^{*^\top} - \Sigma$. For a fixed $\gI$, applying the concentration inequality of Wishart-type matrices in Lemma \ref{lem: wainwright concentration bound prob}, and taking $x = \sqrt{\frac{2s + s^* + c}{n}}$, we have 
\begin{equation*}
\P\qty(\norm{\qty(\frac{1}{n} X X^\top - \E \qty[\frac{1}{n} X X^\top] )_{\gI\gI}}_{\text{sp}} 
\gtrsim (\lambda_1^{*2} + \sigma_{(1)}^{*2})\sqrt{\frac{2s + s^* + c}{n}}) 
\lesssim e^{-(2s + s^* + c)}.
\end{equation*}
Taking $c = 2(2s + s^*) \log \frac{d e}{2s + s^*}$, we obtain that
\begin{multline}
\label{eq: XXtop on I}
\P \qty(\norm{\qty(\frac{1}{n} X X^\top - \E \qty[\frac{1}{n} X X^\top] )_{\gI\gI}}_{\text{sp}} 
\gtrsim (\lambda_1^{*2} + \sigma_{(1)}^{*2})\sqrt{\frac{(2s + s^*) \log d}{n}}) \\
\lesssim \qty(\frac{2s + s^*}{de})^{2s + s^*} e^{-(2s + s^*)\log d}.
\end{multline}
Moreover, 
\begin{equation*}
\P\qty(\|\tilde x_{\gI}\|_2^2 \geq (\lambda_1^{*2} + \sigma_{(1)}^{*2}) x^2 ) 
\leq e^{-\frac{x^2 n}{2(2s + s^*)}}.
\end{equation*}
Taking $x^2 = 4\frac{(2s + s^*)^2 \log d}{n}$, we have
\begin{equation}
\label{eq: xbar on I}
\P\qty(\|\tilde x_{\gI}\|_2^2 \gtrsim (\lambda_1^{*2} + \sigma_{(1)}^{*2}) \frac{(2s + s^*)^2 \log d}{n} ) 
\lesssim \qty(\frac{2s + s^*}{de})^{2s + s^*} e^{-(2s + s^*)\log d}.
\end{equation}
Combining \eqref{eq: XXtop on I} and \eqref{eq: xbar on I} and $s' \geq s$, we obtain that
\begin{multline*}
    \P \qty(\norm{\bigl(M - Q^* Q^{*^\top} - \Sigma\bigr)_{\gI\gI}}_{\text{sp}} 
    \gtrsim(\lambda_1^{*2} + \sigma_{(1)}^{*2}) \qty(
    \sqrt{\frac{s \log d}{n}} 
    + \frac{s^2 \log d}{n}))\\
    \lesssim \qty(\frac{2s + s^*}{de})^{2s + s^*} e^{-(2s + s^*)\log d}.
\end{multline*}
Since there are \(\binom{d}{2s + s^*}\) possible subsets, we use the union bound to get
\begin{multline*}
    \P \bigg(
    \norm{\bigg(M - Q^* Q^{*^\top} - \Sigma\bigg)_{\gI\gI}}_{\text{sp}} 
    \gtrsim (\lambda_1^{*2} + \sigma_{(1)}^{*2}) \bigg(
    \sqrt{\frac{s \log d}{n}} 
    + \frac{s^2 \log d}{n}\bigg), \\
    \exists ~ \gI \subset [d] \text{ with } |\gI| = 2s + s^*
    \bigg) \lesssim \binom{d}{2s + s^*}\bigg(\frac{2s + s^*}{de}\bigg)^{2s + s^*} e^{-(2s + s^*)\log d}.
\end{multline*}
Using the binomial coefficient bound 
\[
\binom{d}{2s + s^*} \leq \bigg(\frac{de}{2s + s^*}\bigg)^{2s + s^*}, 
\]
there exist $c_1'>0$, such that $\P (\Omega_1) \geq 1 - ce^{- c s \log d}$. 

This completes the proof of Lemma \ref{lem: Omega1}. 
\end{proof}

\subsubsection{Proof of Lemma \ref{lem: Omega2}}
\begin{proof}
Since $x_i = Q z_i + \xi_i$, we have $\Cov (x_i) =  Q^* Q^{*^\top} + \Sigma$. Moreover, $z_i$ and $\xi_i$ are the sub-Gaussian random vectors with parameters $\lambda_1^{*2}$ and $\sigma_{(1)}^{*2}$. Therefore, $x_i$ is also the sub-Gaussian random vector with the parameter $\lambda_1^{*2} + \sigma_{(1)}^{*2}$. Write 
\begin{align*}
    M - Q^* Q^{*^\top} - \Sigma =& \frac{1}{n} X X^\top - Q^* Q^{*^\top} - \Sigma - \frac{1}{n} \bar X \bar X^\top,
\end{align*}
where $\bar X = \frac{1}{n} X 1_n 1_n^\top$.

First, we note that $\E \qty[\frac{1}{n} X X^\top] = Q^* Q^{*^\top} + \Sigma$. Applying the concentration inequality of Wishart-type matrices in Lemma \ref{lem: wainwright concentration bound prob}, and taking $x = \sqrt{\frac{d + c_4'}{n}}$, we obtain that
\begin{equation}
\label{eq: XXtop}
    \P \qty( \norm{ \frac{1}{n} X X^\top - \E \qty[\frac{1}{n} X X^\top] }_{\text{sp}} \gtrsim (\lambda_1^{*2} + \sigma_{(1)}^{*2})\qty(\sqrt{\frac{d + c_4'}{n}} + \frac{d}{n})) \lesssim e^{-(d + c_4')}.
\end{equation}

Second, we note that each column of $\tilde X$ is the same. So we can rewrite it as
\begin{align*}
    \frac{1}{n} \bar X \bar X^\top = \bar x \bar x^\top,
\end{align*}
where $\bar x = \frac{1}{n} \sum_{i=1}^n x_i$. Since $x_i$'s are independent zero mean sub-Gaussian random vectors, by Lemma \ref{lem: jin norm concentration bound}, we have
\begin{align*}
    \P \qty( \norm{\tilde x}_2^2 \gtrsim (\lambda_1^{*2} + \sigma_{(1)}^{*2})\sqrt{\frac{d + c_4'}{n}} ) \lesssim e^{-\frac{\sqrt{(d + c_4')n}}{d}}.
\end{align*}
Then
\begin{align}
\label{eq: XbarXbartop}
    \P \qty( \norm{ \frac{1}{n} \tilde X \tilde X^\top}_{\text{sp}} \gtrsim (\lambda_1^{*2} + \sigma_{(1)}^{*2})\sqrt{\frac{d + c_4'}{n}} ) \lesssim e^{-\frac{\sqrt{(d + c_4')n}}{d}}.
\end{align}

Under the assumption that $r < d, d < n$, combining \eqref{eq: XXtop} and \eqref{eq: XbarXbartop}, we have
\begin{align*}
    \P \qty( \norm{M - Q^* Q^{*^\top} - \Sigma}_{\text{sp}} \gtrsim (\lambda_1^{*2} + \sigma_{(1)}^{*2})\sqrt{\frac{d + c_4'}{n}}) &\lesssim e^{-\sqrt{d + c_4'}(\frac{\sqrt{n}}{d} \wedge \sqrt{d + c_4'})}.
\end{align*}

Then there exist $c_3', c_4' > 0$, such that $\P \left( \Omega_2\right) \geq 1 - ce^{-c_4'\sqrt{d + c_4'}(\frac{\sqrt{n}}{d} \wedge \sqrt{d + c_4'})}$.

This completes the proof of Lemma \ref{lem: Omega2}. 
\end{proof}

\subsubsection{Proof of Lemma \ref{lem: solution of loss}}
\label{sec: proof oflemma solution of loss}

\begin{proof}
We control the singular value of $\Delta \qty(M) + P^2 D\qty(M)$. Under the event $\Omega_1$, noting that $\Sigma$ is diagonal, we have
\begin{equation*}
    \Delta(M) =  Q^* Q^{*^\top} - D(Q^* Q^{*^\top}) + \Delta \qty( M - Q^* Q^{*^\top} - \Sigma ).
\end{equation*}

Applying Lemma \ref{lem: zhang off diagonal bound}, we obtain that 
\begin{equation}
\label{eq: offM bound}
    \norm{\Delta \qty( M - Q^* Q^{*^\top} - \Sigma )}_{\text{sp}} \leq 2c_3' (\lambda_1^{*2} + \sigma_{(1)}^{*2})\sqrt{\frac{d + c_4'}{n}}.
\end{equation}

Since $d \ll n$ and $c_4' = o\qty(n)$, we have $2 c_3' (\lambda_1^{*2} + \sigma_{(1)}^{*2})\sqrt{\frac{d + c_4'}{n}} < \frac{\lambda_r^{*2}}{4}$. Moreover, by Assumptions \ref{asm: Lambda} and \ref{asm: Q incoherent}, we have $\| D(Q^* Q^{*^\top}) \|_{\text{sp}} = \lambda_1^{*2} I (U^*) \leq c_3 \lambda_1^{*2}< \frac{\lambda_r^{*2}}{4}$. By Weyl's inequality in Lemma \ref{lem: Weyl}, we have that
\begin{equation}
\begin{aligned}
\label{eq: eigenvalue of DeltaM}
        \lambda_i \qty( \Delta(M) ) &\geq \lambda_r^{*2} - \| D(Q^* Q^{*^\top}) - \Delta \qty( M - Q^* Q^{*^\top} - \Sigma ) \|_{\text{sp}} > \frac{\lambda_r^{*2}}{2}, \quad i \leq r, \\
        \lambda_i \qty( \Delta(M) ) &\geq - \| D(Q^* Q^{*^\top}) - \Delta \qty( M - Q^* Q^{*^\top} - \Sigma ) \|_{\text{sp}} > -\frac{\lambda_r^{*2}}{2}, \quad i > r.
\end{aligned}
\end{equation}

Moreover, noting that $M = \frac{1}{n} (X - \tilde{X})(X - \tilde{X})^\top$, we have
\begin{equation}
\label{eq: eigenvalue of DeltaM + DM}
\lambda_i \qty( \Delta(M) + P^2 D(M) ) \geq \lambda_i \qty( \Delta(M) ).
\end{equation}

We have already proved that the largest $r$ eigenvalues of $\Delta(M) + P^2 D(M)$ are also the largest $r$ singular values according to \eqref{eq: eigenvalue of DeltaM} and \eqref{eq: eigenvalue of DeltaM + DM}. By Lemma \ref{lem: eckart}, the minimizer $\hat V$ of loss \ref{loss: cl conti triplet transpose expectation simplified} $\mL(V)$ satisfies $\hat V \hat V^\top = \frac{1}{\lambda} \rank_r\qty(\Delta \qty(M) + P^2 D\qty(M))$.

This completes the proof of Lemma \ref{lem: solution of loss}. 
\end{proof}

\section{Proofs for Section \ref{sec:downstream}}
\label{sec:app proof of section downstream}

\subsection{Proof of Theorem \ref{thm: classification risk}}

\begin{proof}
Note that
\begin{align*}
 \inf_{w \in \mathbb{R}^r} \mathbb{E}_0\qty[l_c\qty(\delta_{\hat{Q}, w})] &= \inf_{w \in \mathbb{R}^r} \mathbb{E}_0\qty[l_c\qty(\mathbbm{1}\left\{F(w^\top (\hat{Q}^\top \hat{Q})^{-1} \hat{Q}^\top x_0) \geq \frac{1}{2}\right\})]\\
 &= \inf_{w \in \mathbb{R}^r} \mathbb{E}_0\qty[l_c\qty(\mathbbm{1}\left\{F(w^\top (\hat{Q}^\top \hat{Q})^{-\frac{1}{2}} \hat{Q}^\top x_0) \geq \frac{1}{2}\right\})].
\end{align*}

Similarly, for $Q$, we have,
\begin{equation*}
    \inf_{w \in \mathbb{R}^r} \mathbb{E}_0\qty[l_c\qty(\delta_{Q, w})] = \inf_{w \in \mathbb{R}^r} \mathbb{E}_0\qty[l_c\qty(\mathbbm{1}\left\{F(w^\top (Q^{*^\top} Q^*)^{-\frac{1}{2}} Q^{*^\top} x_0) \geq \frac{1}{2}\right\})].
\end{equation*}

Also note that $\hat{Q} (\hat{Q}^\top \hat{Q})^{-\frac{1}{2}}, Q^*(Q^{*^\top} Q^*)^{-\frac{1}{2}} \in \sO_{d, r}$, by Lemma \ref{lem: eckart},
\begin{equation*}
    \norm{\sin \Theta(\hat{Q} (\hat{Q}^\top \hat{Q})^{-\frac{1}{2}}, Q^*(Q^{*^\top} Q^*)^{-\frac{1}{2}})}_F \leq E(\hat Q).
\end{equation*}

Since we have $E(\hat Q) \leq \frac{\sigma_{(1)}^{*2}}{\kappa_\Sigma} \wedge \frac{1}{2}$, applying Lemma \ref{lem: ji classification risk}, we have
\begin{multline*}
    \inf_{w \in \mathbb{R}^r} \mathbb{E}_0\qty[l_c\qty(\delta_{\hat{Q}, w})]  - \inf_{w \in \mathbb{R}^r} \mathbb{E}_0\qty[l_c\qty(\delta_{Q^*, w})] \lesssim \\
    \qty(\qty(\kappa_\Sigma \qty(1 + \frac{1}{\sigma_{(1)}^{*2}}))^3 +  \frac{\kappa_\Sigma}{\sigma_{(1)}^{*2}} \qty(1 + \sigma_{(1)}^{*2})^2) \norm{\sin \Theta(\hat{Q} (\hat{Q}^\top \hat{Q})^{-\frac{1}{2}}, Q^*)}_{\text{sp}}. 
\end{multline*}

By the assumption $\sigma_{(1)}^{*2}, \kappa_\Sigma \asymp 1$, we obtain that, 
\begin{equation*}
    \inf_{w \in \mathbb{R}^r} \mathbb{E}_0\qty[l_c\qty(\delta_{\hat{Q}, w})]  - \inf_{w \in \mathbb{R}^r} \mathbb{E}_0\qty[l_c\qty(\delta_{Q^*, w})] \lesssim E(\hat Q).
\end{equation*}

This completes the proof of Theorem \ref{thm: classification risk}. 
\end{proof}

\subsection{Proof of Theorem \ref{thm: detect important edges continuous}}
\begin{proof}
We show that, for any $e \in \mathcal{C}$ and $e' \notin \mathcal{C}$, the following inequality holds,
\[
\| \hat{q}_e\|_2 \geq \|\hat{q}_{e'}\|_2.
\] 

By the triangle inequality, we have 
\begin{align*}
        \| \hat{q}_e\|_2 & \geq \| q_e\|_2 - \| \hat{q}_e - q_e\|_2, \\
        \|\hat{q}_{e'}\|_2 & \leq \| q_{e'}\|_2 + \|\hat{q}_{e'} - q_{e'}\|_2.
\end{align*}
Since $e' \notin \mathcal{C}$, we have $\| q_{e'}\|_2 = 0$, which simplifies the second inequality to:
\[
    \|\hat{q}_{e'}\|_2 \leq \|\hat{q}_{e'} - q_{e'}\|_2.
\]

By Theorem \ref{thm: edge embeddings sparse} and the assumption  
\[
    \| q_e\|_2 \gtrsim \frac{s}{s^*} \sqrt{\frac{s^* r \log d}{n}},
\]  
we have that 
\[
    \| \hat{q}_e\|_2 - \|\hat{q}_{e'}\|_2 \geq 0.
\]  

This completes the proof of Theorem \ref{thm: detect important edges continuous}.
\end{proof}

\subsection{Proof of Theorem \ref{thm: community detection error}}
\label{sec:appendix-community}

\begin{proof}
We first provide a more detailed description of the community detection procedure based on edge embedding. We then analyze its behavior.

\renewcommand{\thealgorithm}{S.\arabic{algorithm}}

\begin{algorithm}[H]
\caption{Node community detection based on edge embedding.}
\label{alg: node clustering}
\begin{algorithmic}[1]
\State \textbf{Input:} Edge embeddings matrix $\hat{Q} \in \mathbb{R}^{d \times r}$; Number of clusters $G$.
\State \textbf{Output:} Partition of nodes into $G$ communities.

\State Initialize the node similarity matrix $\hat{S} \in \mathbb{R}^{v \times v}$ with all entries set to $0$.
\For{each edge $e \in [d]$}
    \State Compute the $l_2$-norm of the edge embeddings and update the similarity matrix:
    \[
    \hat{S}_{u_e u_e'} = \hat{S}_{u_e' u_e} \gets \| \hat{q}_e \|_2.
    \]
\EndFor

\State Compute the normalized Laplacian matrix:
\[
\hat{L} \gets \hat{D}^{-\frac{1}{2}} \hat{S} \hat{D}^{-\frac{1}{2}},
\]
where $\hat{D}$ is the diagonal degree matrix with $\hat{D}_{vv} = \sum_{v'} \hat{S}_{v,v'}$.

\State Compute $\hat{\Gamma} \in \mathbb{R}^{p \times G}$, the matrix of the leading $G$ eigenvectors of $\hat{L}$ (ordered by absolute eigenvalue).

\State Solve the $k$-means clustering problem with $\hat{\Gamma}$ as input. Let $(\hat{\Theta}, \hat{Y})$ be a $(1 + \epsilon)$-approximate solution to the $k$-means problem which means it satisfies $\|\hat{\Theta} \hat{Y} - \hat{\Gamma}\|_F^2 \leq (1 + \epsilon) \min_{\Theta \in \mathbb{M}_{v,G}, Y \in \mathbb{R}^{G \times G}} \|\Theta Y - \hat{\Gamma}\|_F^2$, where $\mathbb{M}_{v,G}$ is the set of membership matrices

\State \textbf{Return:} The membership matrix $\hat{\Theta}$.
\end{algorithmic}
\end{algorithm}

In our analysis, we first bound $\|\hat{L} - L\|_{\text{sp}}$.
\begin{multline*}
\norm{\hat{L} - L}_{\text{sp}} 
= \norm{\hat{D}^{-\frac{1}{2}} \hat{S} \hat{D}^{-\frac{1}{2}} - \hat{D}^{-\frac{1}{2}} S \hat{D}^{-\frac{1}{2}} 
+ \hat{D}^{-\frac{1}{2}} S \hat{D}^{-\frac{1}{2}} - D^{-\frac{1}{2}} S D^{-\frac{1}{2}}}_{\text{sp}} \\
\leq \qty(\norm{D^{- \frac{1}{2}}}_{\text{sp}} + \norm{\hat{D}^{- \frac{1}{2}}}_{\text{sp}})^2 \norm{\hat{S} - S}_{\text{sp}}  + \qty(2 \norm{D^{- \frac{1}{2}}}_{\text{sp}} + \norm{\hat{D}^{- \frac{1}{2}}}_{\text{sp}}) \norm{S}_{\text{sp}} 
\norm{\hat{D}^{- \frac{1}{2}} - D^{- \frac{1}{2}}}_{\text{sp}}.    
\end{multline*}
Note that
\begin{equation*}
E(\hat Q) = \norm{\hat{Q} \hat{Q}^\top - Q^* Q^{*^\top}}_{F}. 
\end{equation*}
Therefore, 
\begin{equation*}
\norm{\hat{S} - S}_{\text{sp}} = \norm{\Delta(\hat{S} - S)}_{\text{sp}} + \rho_G \leq E(\hat Q) + \rho_G,
\end{equation*}
Moreover,
\begin{equation*}
\norm{D^{-\frac{1}{2}}}_{\text{sp}} = \frac{1}{\sqrt{\min_{i \in [v]} D_{ii}}} = \frac{1}{\sqrt{v \tau_v}}.
\end{equation*}
By $\abs{\hat{D}_{ii} - D_{ii}} = \abs{\sum_{j \neq i} (\hat{S}_{ij} - S_{ij})} \leq \sqrt{v} E(\hat Q)$, we have that
\begin{equation*}
\qty| \frac{1}{\sqrt{\hat{D}_{ii}}} - \frac{1}{\sqrt{D_{ii}}} | 
= \qty| \frac{\sqrt{D_{ii}} - \sqrt{\hat{D}_{ii}}}{\sqrt{\hat{D}_{ii}} \sqrt{D_{ii}}} |
= \qty| \frac{D_{ii} - \hat{D}_{ii}}{D_{ii} \sqrt{\hat{D}_{ii}} + \hat{D}_{ii} \sqrt{D_{ii}}} |
\leq \frac{\sqrt{v} E(\hat Q)}{\tau_v^{\frac{3}{2}} v^{\frac{3}{2}} - \tau_v^{\frac{1}{2}} v^{\frac{3}{2}} E(\hat Q)}.
\end{equation*}
With the assumption $E(\hat Q) = o\qty(\tau_v \sqrt{v} \wedge \rho_G)$, we have 
\begin{equation*}
\norm{\hat{D}^{-\frac{1}{2}} - D^{-\frac{1}{2}}}_{\text{sp}} \leq \frac{2 E(\hat Q)}{\tau_v^{\frac{3}{2}} v}.
\end{equation*}
Combining these results yields
\begin{equation*}
\norm{\hat{L} - L}_{\text{sp}} \leq 
\qty(\frac{1}{\sqrt{\tau_v v}} + \frac{2 E(\hat Q)}{\tau_v^{\frac{3}{2}} v})^2 (E(\hat Q) + \rho_G) 
+ \qty(\frac{2}{\sqrt{\tau_v v}} + \frac{2 E(\hat Q)}{\tau_v^{\frac{3}{2}} v}) \frac{2 E(\hat Q)}{\tau_v^{\frac{3}{2}} v} \norm{S}_{\text{sp}}.
\end{equation*}

By Lemma \ref{lem: lei principal subspace perturbation}, there exists an orthogonal matrix $O \in \sO_{G, G}$, such that 
\begin{equation*}
\norm{\hat{\Gamma} - \Gamma O}_F \leq \frac{2 \sqrt{2} G}{\lambda_G} \norm{\hat{L} - L}_{\text{sp}}.
\end{equation*}
where $\hat{\Gamma}, \Gamma$ are the corresponding eigenvector matrices.

Note that $L = D^{-\frac{1}{2}} S D^{-\frac{1}{2}} = D^{-\frac{1}{2}} \Theta B \Theta^\top D^{-\frac{1}{2}} = \Theta B_L \Theta^\top$, where $ B_L = D_B^{-\frac{1}{2}} B D_B^{-\frac{1}{2}}$ and $D_B = \diag(B \Theta^\top 1_v)$. Since \( B \) is symmetric, full rank, and positive definite, it follows that \( B_L \) is also symmetric and full rank. By Lemma \ref{lem: lei basic eigen sbm}, there exists \( Y \), such that \( WO = \Theta Y O = \Theta Y' \), where \( \|Y_{g *} - Y_{l *}\| = \sqrt{v_g^{-1} + v_l^{-1}} \).

Next, we leverage Lemma \ref{lem: lei approximate k means error bound}. We choose
\begin{equation*}
\delta_g = \sqrt{\frac{1}{v_g} + \frac{1}{\max\{v_\ell : \ell \neq g\}}}.
\end{equation*}
Then $v_g \delta_g^2 \geq 1$. We next check the condition \eqref{eq: hatgamma - gamm}. When $E(\hat Q) = o\qty(\tau_v \sqrt{v} \wedge \rho_G)$, we have
\begin{equation*}
    (16 + 8\epsilon)\| \hat{\Gamma} - \Gamma O \|_F^2 \leq \frac{8 G^2}{\lambda_G^2} \norm{\hat{L} - L}_{\text{sp}}^2 \leq 128 (2 + \epsilon ) \frac{\rho_G^2 K ^2}{\tau_v^2 v^2 \lambda_G^2}.
\end{equation*}
We then have $128 (2 + \epsilon ) \frac{\rho_G^2 G^2}{\tau_v^2 v^2 \lambda_G^2} \leq 1 \leq \min_{1 \leq g < G} v_g \delta_g^2$. By \eqref{eq: miserror Sk},
\begin{equation*}
\sum_{g=1}^G |S_g| \qty(\frac{1}{v_g} + \frac{1}{\max \qty{v_l : l \neq k}}) 
\leq 128 (2 + \epsilon ) \frac{\rho_G^2 K ^2}{\tau_v^2 v^2 \lambda_G^2},
\end{equation*}
which implies that
\begin{equation*}
\tilde{L}(\hat{\Theta}, \Theta) \leq \max_{1 \leq g \leq G} \frac{|S_g|}{v_g} 
\leq \sum_{g=1}^G \frac{|S_g|}{v_g} 
\leq 128 (2 + \epsilon ) \frac{\rho_G^2 K ^2}{\tau_v^2 v^2 \lambda_G^2}.
\end{equation*}
Recalling that $v_{\max}'$ is the second largest community size, we obtain that
\begin{equation*}
L(\hat{\Theta}, \Theta) \leq \frac{1}{v} \sum_{g=1}^G |S_g| \leq 128 (2 + \epsilon ) \frac{\rho_G^2 G ^2 v_{\max}'}{\tau_v^2 v^3 \lambda_G^2}.
\end{equation*}

This completest the proof of Theorem \ref{thm: community detection error}.
\end{proof}

\section{Proofs for Section \ref{sec:comparePCA}}
\label{sec:app proof of comparepca}

\subsection{Proof of Theorem \ref{thm: lower bound for sPCA}}

\begin{proof}
Recall that $Q^* Q^{*^\top} = U^* \Lambda^* U^{*^\top}$, where $\Lambda^* = \operatorname{diag}(\lambda_1^{*2}, \dots, \lambda_r^{*2})$ with $\lambda_1^{*2} \geq \cdots \geq \lambda_r^{*2}$, and $U^* \in \mathbb{R}^{d \times r}$. For a matrix $A$, let $\operatorname{supp}(A)$ denote the indices of non-zero rows, i.e., $\operatorname{supp}(A) = \{ i \in [n] : \| A_{i*} \|_2 > 0 \}$ where $A_{i*}$ denotes the $i$-th row of $A$.

{Since to establish lower bounds, it suffices to to consider the least favorable case among the given parameter space, i.e., $(Q^*, \Sigma^*) \in \gM_s$, in the following, we establish the lower bound by focusing on a this least favorable case.} {Specifically, we consider a scenario as follows: suppose $\sigma_i^{*2} = c_\sigma$ for all $i \in \operatorname{supp}\qty(Q^*)$ and $\sigma_{(r)}^{*2} > c_\sigma + \lambda_1^{*2}$. Consider $\operatorname{supp}(Q^*) = [s]$ and $\sigma_{s+i}^{*2} = \sigma_{(i)}^{*2}$ for $i \in [r]$, which means the first $r$th eigenvalues are $(\sigma_{s+1}^{*2},\cdots,\sigma_{s+r}^{*2})$.} Then
\begin{equation*}
Q^* = \begin{pmatrix}
Q^*_{[s]} \\
0
\end{pmatrix},
\end{equation*}
where $Q^*_{[s]}$ represents the first $r$ rows corresponding to the support $\operatorname{supp}\qty(Q^*) = [s]$. Note that 
\begin{equation*}
Q^* {Q^*}^\top = U^* \Lambda^* {U^*}^\top 
= \begin{pmatrix}
U^*_{[s]} \\
0
\end{pmatrix}
\Lambda^*
\begin{pmatrix}
{U^*_{[s]}}^\top & 0
\end{pmatrix}.
\end{equation*}
For $\Sigma$, we have $\sigma_i^{*2} = c_\sigma$ for $i \le s$. Then
\begin{equation*}
\Sigma = 
\begin{pmatrix}
U^*_{[s]} & 0 \\
0 & I_{d-s}
\end{pmatrix}
\begin{pmatrix}
c_\sigma I_s & 0 \\
0 & \mathrm{diag}(\sigma_{s+1}^{*2}, \ldots, \sigma_d^{*2})
\end{pmatrix}
\begin{pmatrix}
{U^*_{[s]}}^\top & 0 \\
0 & I_{d-s}
\end{pmatrix}.
\end{equation*}
Combining the above results, we have,
\begin{equation*}
\mathbb{E}[XX^\top] = 
\begin{pmatrix}
U^*_{[s]} & 0 \\
0 & I_{d-s}
\end{pmatrix}
\begin{pmatrix}
c_\sigma I_s + \Lambda^* & 0 \\
0 & \mathrm{diag}(\sigma_{s+1}^{*2}, \ldots, \sigma_d^{*2})
\end{pmatrix}
\begin{pmatrix}
{U^*_{[s]}}^\top & 0 \\
0 & I_{d-s}
\end{pmatrix}.
\end{equation*}
By the assumption $\sigma_{(r)}^{*2} > \sigma + \lambda_1^{*2}$, we get that the first $r$ leading eigenvectors of $\mathbb{E}[XX^\top]$ are $U_x = (e_{s+1}, \dots, e_{s+r})$, where $\{e_i\}_{i=1}^d$ represents the canonical basis. So the distance between $U_x$ and $U^*$ is
\begin{align*}
\norm{U_x U_x^\top - U^* {U^*}^\top}_F^2 
&= \tr\qty( \qty(U_x U_x^\top - U^* {U^*}^\top)^\top \qty(U_x U_x^\top - U^* {U^*}^\top) ) \\
&= \tr\qty( U_x U_x^\top + U^* {U^*}^\top ) = 2r.
\end{align*}

Moreover, by Assumption~\ref{asm: bound for spca}, with probability at least $1 - c e^{-c s \log d}$, we have
\begin{equation*}
\norm{\hat{U}_x\hat{U}_x^\top - U^*U^{*^\top}}_F \geq \norm{U_x U_x^\top - U^* {U^*}^\top}_F - \norm{\hat{U}_x \hat{U}_x^\top - U_x U_x^\top}_F \gtrsim \sqrt{r}.
\end{equation*}
Therefore, 
\begin{equation*}
\sup_{(Q^*, \Sigma) \in \gM_s} \P \qty(\norm{\hat{U}_x\hat{U}_x^\top - U^*U^{*^\top}}_F \gtrsim \sqrt{r}) \geq 1- c e^{-c s \log d}.
\end{equation*}

This completes the proof of Theorem \ref{thm: lower bound for sPCA}. 
\end{proof}

\subsection{Proof of Theorem \ref{thm: classification risk lower bound}}

\begin{proof}
First, we observe that
\begin{align*}
 \inf_{w \in \mathbb{R}^r} \mathbb{E}_0\qty[l_c\qty(\delta'_{\hat{U}_x, w})] 
 &= \inf_{w \in \mathbb{R}^r} \mathbb{E}_0\qty[l_c\qty(\mathbbm{1}\left\{F\qty(w^\top \hat{\Lambda}_r^{-\frac{1}{2}} \hat{U}_x^\top x_0) \geq \frac{1}{2}\right\})] \\
 &= \inf_{w \in \mathbb{R}^r} \mathbb{E}_0\qty[l_c\qty(\mathbbm{1}\left\{F\qty(w^\top \hat{U}_x^\top x_0) \geq \frac{1}{2}\right\})]\\
 &= \inf_{w \in \mathbb{R}^r} \mathbb{E}_0\qty[l_c\qty(\mathbbm{1}\left\{F\qty(w^\top (\hat{U}_x^\top \hat{U}_x)^{-\frac{1}{2}}\hat{U}_x^\top x_0) \geq \frac{1}{2}\right\})].
\end{align*}
Moreover,
\begin{align*}
    \inf_{w \in \mathbb{R}^r} \mathbb{E}_0\qty[l_c\qty(\delta_{Q^*, w})] & = \inf_{w \in \mathbb{R}^r} \mathbb{E}_0\qty[l_c\qty(\mathbbm{1}\left\{F(w^\top (Q^{*^\top} Q^*)^{-\frac{1}{2}} Q^{*^\top} x_0) \geq \frac{1}{2}\right\})]\\
    &= \inf_{w \in \mathbb{R}^r} \mathbb{E}_0\qty[l_c\qty(\mathbbm{1}\left\{F\qty(w^\top U^{*^\top} x_0) \geq \frac{1}{2}\right\})].
\end{align*}
By Lemma \ref{lem: ji classification risk lower bound}, and under the condition
\[
E_U^2(\hat U) \geq r - \frac{\sigma_{(1)}^{*2}}{\kappa_\Sigma \qty(1 + \sigma_{(1)}^{*2})} + \frac{c_6}{\kappa_\Sigma},
\]
we have
\begin{equation*}
    \inf_{w \in \mathbb{R}^r} \mathbb{E}_0\qty[l_c\qty(\delta_{\hat{U}_x, w})]  - \inf_{w \in \mathbb{R}^r} \mathbb{E}_0\qty[l_c\qty(\delta_{Q^*, w})] \gtrsim \frac{\qty(\sigma_{(1)}^{*2} + 1)^{\frac{3}{2}}}{\qty(\sigma_{(1)}^{*2} + \kappa_\Sigma)^{\frac{3}{2}}\sigma_{(1)}^{*2}} c_6.
\end{equation*}
Therefore, there exists a constant $c_P$, which depends only on $c_6$, $\sigma_{(1)}^{*2}$, and $\kappa_\Sigma$, such that:
\begin{equation*}
    \inf_{w \in \mathbb{R}^r} \mathbb{E}_0\qty[l_c\qty(\delta'_{\hat{U}_x, w})]  - \inf_{w \in \mathbb{R}^r} \mathbb{E}_0\qty[l_c\qty(\delta_{Q^*, w})] \geq c_P.
\end{equation*}

This completes the proof of Theorem \ref{thm: classification risk lower bound}.
\end{proof}

\end{document}

%% file: math_commands.tex
\usepackage{amsmath,amsfonts,bm}

\def\Eqref#1{Equation~\eqref{#1}}

\def\1{\bm{1}}

\def\mL{{\bm{L}}}

\DeclareMathAlphabet{\mathsfit}{\encodingdefault}{\sfdefault}{m}{sl}
\SetMathAlphabet{\mathsfit}{bold}{\encodingdefault}{\sfdefault}{bx}{n}

\def\gB{{\mathcal{B}}}

\def\gD{{\mathcal{D}}}
\def\gE{{\mathcal{E}}}

\def\gG{{\mathcal{G}}}

\def\gI{{\mathcal{I}}}
\def\gJ{{\mathcal{J}}}

\def\gL{{\mathcal{L}}}
\def\gM{{\mathcal{M}}}

\def\gV{{\mathcal{V}}}

\def\sO{{\mathbb{O}}}

\def\sR{{\mathbb{R}}}

\newcommand{\E}{\mathbb{E}}

\newcommand{\R}{\mathbb{R}}

\newcommand{\Var}{\mathrm{Var}}
\newcommand{\diag}{\mathrm{diag}}

\newcommand{\Cov}{\mathrm{Cov}}

\DeclareMathOperator*{\argmin}{arg\,min}

\DeclareMathOperator\supp{supp}

\DeclareMathOperator{\dist}{dist}

\def\mL{\mathcal{L}}

\def\P {\mathbb{P}}
\def\E {\mathbb{E}}